\documentclass[acmsmall, screen, nonacm]{acmart}
\usepackage{algorithm}
\usepackage{algpseudocode}
\usepackage{xcolor}
\usepackage{textcomp} 
\usepackage{multirow} 
\usepackage{mathtools} 
\usepackage{array} 
\usepackage{makecell}
\usepackage{subcaption}
\usepackage{wrapfig}

\definecolor{blue(ncs)}{rgb}{0.0, 0.53, 0.74}
\usepackage[most]{tcolorbox} 
\tcbset{enhanced, breakable} 

\newtcolorbox{researchgap}[1][]{
  colback=blue(ncs)!4,      
  colframe=black!30,    
  boxrule=0.4pt,
  arc=0pt,              
  left=6pt,right=6pt,top=6pt,bottom=6pt,
  borderline west={4pt}{0pt}{blue(ncs)!70},
  breakable,
  #1
}

\newtcolorbox{takeaway}[1][]{
  enhanced,
  colback=white,   
  colframe=white,  
  boxrule=0pt,         
  top=6pt,bottom=6pt,left=0pt,right=0pt,
  borderline north={1pt}{0pt}{blue(ncs)}, 
  borderline south={1pt}{0pt}{blue(ncs)},
  breakable,
  #1
}

\AtBeginDocument{%
  }

\setcopyright{none}
\acmDOI{} \acmISBN{} \acmYear{} \copyrightyear{} \acmPrice{}

\usepackage{fancyhdr}
\makeatletter
\AtBeginDocument{%

  \fancypagestyle{firstpagestyle}{%
    \fancyhf{}%
    \fancyfoot[LO,LE]{\footnotesize \textit{Preprint (September 2025). 
    }}%
  }%

  \fancypagestyle{standardpagestyle}{%
    \fancyfoot[LO,LE]{\footnotesize \textit{Preprint (September 2025). 
    }}%
  }%
}
\makeatother

\begin{document}

\title{Mind Meets Space: Rethinking Agentic Spatial Intelligence from a Neuroscience-inspired Perspective}

\author{Bui Duc Manh}
\email{ducmanh.bui@ntu.edu.sg}
\authornote{Equal contribution.}
\affiliation{%
  \institution{School of EEE, Nanyang Technological University}
  \country{Singapore}
}

\author{Soumyaratna Debnath}
\email{soumyara004@e.ntu.edu.sg}
\authornotemark[1]
\affiliation{%
  \institution{School of EEE, Nanyang Technological University}
  \country{Singapore}
}

\author{Zetong Zhang}
\email{zhangzt22@mails.tsinghua.edu.cn}
\authornote{Co-second author. Work done during internship at Nanyang Technological University.}
\affiliation{%
  \institution{Department of Civil Engineering, Tsinghua University}
  \country{China}
}

\author{Shriram Damodaran}
\email{ds.ic.21@nitj.ac.in}
\authornotemark[2]
\affiliation{%
  \institution{Dr. B. R. Ambedkar National Institute of Technology, Jalandhar}
  \country{India}
}

\author{Arvind Kumar}
\email{arvkumar@kth.se}
\affiliation{%
  \institution{KTH Royal Institute of Technology}
  \country{Sweden}
}

\author{Yueyi Zhang}
\email{yueyi.zhang@miromind.ai}
\affiliation{%
  \institution{MiroMind}
  \country{Singapore}
}

\author{Lu Mi}
\email{milu@mail.tsinghua.edu.cn}
\affiliation{%
  \institution{College of AI, Tsinghua University}
  \country{China}
}

\author{Erik Cambria}
\email{cambria@ntu.edu.sg}
\affiliation{%
  \institution{CCDS, Nanyang Technological University}
  \country{Singapore}
}

\author{Lin Wang}
\email{linwang@ntu.edu.sg}
\authornote{Corresponding author.}
\affiliation{%
  \institution{School of EEE, Nanyang Technological University}
  \country{Singapore}
}
 
\renewcommand{\shortauthors}{Manh, et al.}

\begin{abstract}
Recent advances in agentic AI have led to systems capable of autonomous task execution and language-based reasoning, yet their spatial reasoning abilities remain limited and underexplored, largely constrained to symbolic and sequential processing. In contrast, human spatial intelligence, rooted in integrated multisensory perception, spatial memory, and cognitive maps, enables flexible, context-aware decision-making in unstructured environments. Therefore, bridging this gap is critical for advancing Agentic Spatial Intelligence toward better interaction with the physical 3D world.
To this end, we first start from scrutinizing the spatial neural models as studied in computational neuroscience, and accordingly introduce a novel computational framework grounded in neuroscience principles. This framework maps core biological functions to six essential computation modules: bio-inspired multimodal sensing, multi-sensory integration, egocentric–allocentric conversion, an artificial cognitive map, spatial memory, and spatial reasoning. Together, these modules form a perspective landscape for agentic spatial reasoning capability across both virtual and physical environments.
On top, we conduct a framework-guided analysis of recent methods, evaluating their relevance to each module and identifying critical gaps that hinder the development of more neuroscience-grounded spatial reasoning modules. We further examine emerging benchmarks and datasets and explore potential application domains ranging from virtual to embodied systems, such as robotics.
Finally, we outline potential research directions, emphasizing the promising roadmap that can generalize spatial reasoning across dynamic or unstructured environments. We hope this work will benefit the research community with a neuroscience-grounded perspective and a structured pathway.
Our project page can be found at \href{https://github.com/BioRAILab/Awesome-Neuroscience-Agentic-Spatial-Reasoning}{Github}.
\end{abstract}




\maketitle
\pagestyle{standardpagestyle}

\section{Introduction}

Artificial Intelligence (AI) has recently achieved remarkable advances, evolving from passive and data-driven models to interactive autonomous systems with complex decision-making abilities~\cite{7pillars}. This transformation marks the emergence of agentic AI, which is capable of reasoning and adaptively interacting in dynamic environments. 
For instance, OpenAI introduced ChatGPT Agent~\cite{openaiagent}, an agentic system that leverages the GPT-based large language model (LLM) with reasoning abilities (e.g., OpenAI o3) to autonomously plan, execute, and refine multi-step tasks such as browsing, coding, and interacting with real-world tools. Also, Google presented Gemini CLI~\cite{geminicli}, an open-source agent that is capable of integrating with the command line interface to locally perform autonomous tasks, including content generation, problem solving, deep research, and task management. By employing LLM with language-based generalization, agentic systems have shown rapid advancement across both industry and academia. While these systems demonstrate robust capabilities in language-based reasoning and virtual task automation, their intelligence remains primarily symbolic and sequential~\cite{maogpt}. Hence, although current agents excel at manipulating structured inputs, they often lack the ability to represent spatial structures, reason about spatial relationships, or decision-planning in spatial contexts, such as physical layouts, navigational environments, or abstract relational spaces~\cite{chen2025spatial}. 

This limitation contrasts sharply with the spatial intelligence in humans. Based on Dual Coding Theory (DCT)~\cite{paivio2013imagery}, alongside the verbal system (e.g., language, text), human intelligence is also grounded in a spatial system that encodes visual and spatial information, enabling individuals to perceive their surroundings, construct internal representations, and make context-aware decisions in complex environments~\cite{xuuare}. Therefore, enhancing agents with spatial reasoning capabilities is essential for achieving better interaction with the physical 3D environments for both virtual and physical embodied agentic systems, such as robotics. However, recent approaches often explore spatial reasoning as static perception with task-specific planning, hindering the construction of internal spatial models that support generalizable, context-aware decision-making in space. 
\textbf{Bridging this gap requires rethinking agentic spatial reasoning not just as a computational problem but as a cognitive process}~\cite{zinan2025nature, maobri}.

\begin{figure*}[t!]
    \centering
    \vspace{-0.5em}
    \includegraphics[width=\linewidth]{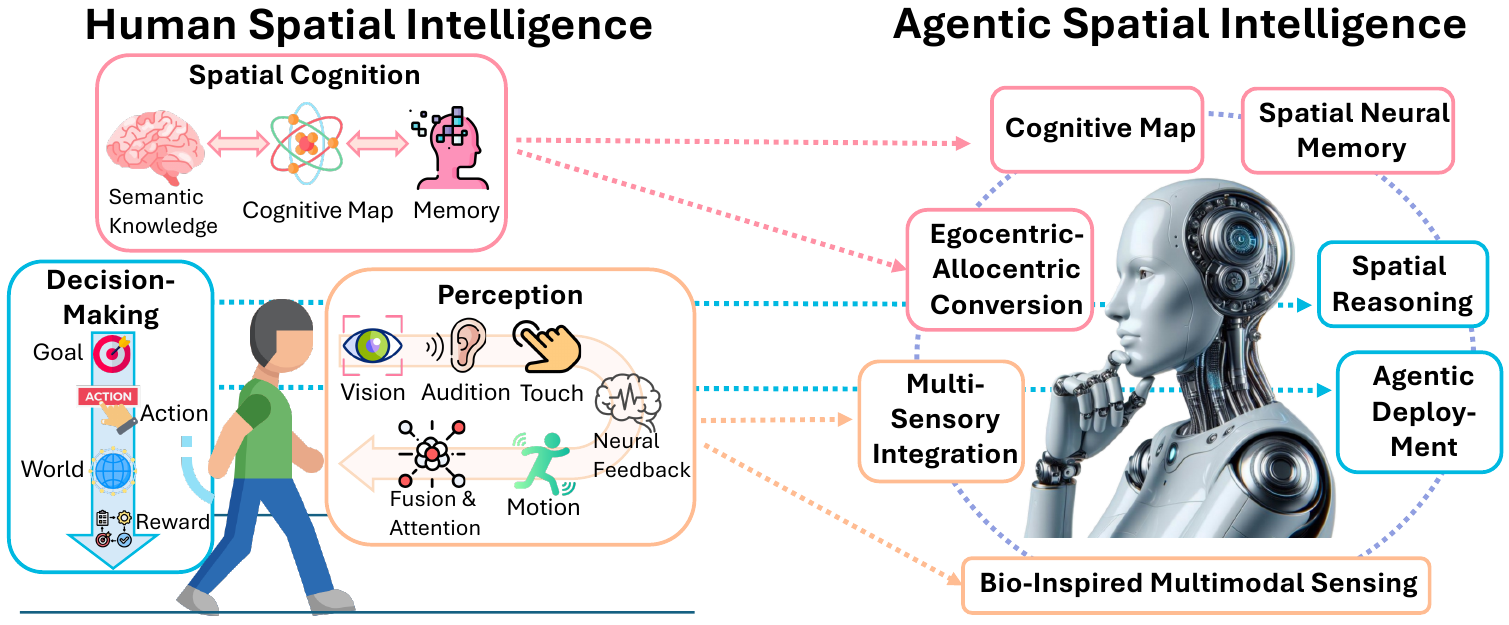}
    \Description{A framework diagram illustrating human-inspired agentic spatial intelligence.}
    \vspace{-18pt}
    \caption{Illustration of neuroscience-inspired Agentic Spatial Intelligence. As the core functions of human spatial reasoning lie in multimodal perception, cognitive mapping, memory systems, and spatial reasoning for decision-making, agents can be abstracted into corresponding AI modules that form a bio-inspired framework for enabling spatial reasoning and adaptive behavior. In our framework, these functions are mimicked through components, including bio-inspired multimodal sensing, multi-sensory integration, egocentric–allocentric conversion, cognitive map, spatial memory, and spatial reasoning for adaptive deployment.}
    \vspace{-1.2em}
    \label{fig:framework}
\end{figure*}

In this work, we cross discipline boundaries by exploring the studies in neuroscience regarding how humans navigate and reason about space.
Previous research reveals that the brain processes spatial information through a structured hierarchy, commencing from multisensory perception and culminating in an internal representation of the world, commonly referred to as a cognitive map, which supports goal-directed spatial reasoning~\cite{byrne2007remembering}. These processes are distributed across specialized brain regions including the posterior parietal cortex (PPC), hippocampus (HPC), and prefrontal cortex (PFC), forming a network that underlies flexible navigation and planning in real-world environments~\cite{keefe1971hippocampus}. \textbf{Therefore, neuroscience presents a robust biologically-grounded baseline for spatial reasoning, yielding better interaction with physical 3D environments}.

\textit{\textbf{Related Works}:} Following the emerging agentic AI, several surveys have extensively explored its capabilities, architectures, and emerging applications~\cite{zinan2025nature, wang2024survey, durante2024agent, li2025system, liu2025neural, qureshi2025thinking}. Among them,~\cite{wang2024survey} provides a comprehensive overview of agents powered by LLM, focusing on their architectural design, enhancement modules (such as memory, planning, and tool use), and diverse application domains. Following that,~\cite{durante2024agent} surveys the broader landscape of agentic systems, with an emphasis on multimodal interaction capabilities that integrate vision, language, and action. It highlights recent architectural trends, training paradigms, and discusses the deployment of the agents across embodied settings. Additionally,~\cite{li2025system} explores reasoning ability in LLM, focusing on examination and evaluation of core reasoning methods in representative models (e.g., OpenAI o1/o3, DeepSeek R1). Standing on neuroscience viewpoints,~\cite{liu2025neural} provides a biological-inspired design for embodied agents, focusing on both definition and implementation that mimic human brain. The authors from~\cite{qureshi2025thinking} also survey brain-inspired approaches toward artificial general intelligence, bridging AI with cognitive neuroscience. It critiques token-centric LLM and highlights reasoning, memory, and multi-agent coordination as key components for embodied intelligence. Focusing more on reasoning,~\cite{zinan2025nature} explores general reasoning approaches in agentic AI by drawing on neuroscience foundations to examine multiple perspectives, including perceptual, dimensional, logical, and interactive reasoning. It further highlights the key gaps that prevent current AI methods from achieving these human-like reasoning capabilities. 
In summary, these works remain largely LLM-centric and language-based reasoning in symbolic and logical domains.

\textit{\textbf{Challenges}:} Current spatial reasoning models still face fundamental challenges: \textbf{(1)} They remain heavily vision-centric, often relying on pre-trained vision-language models (VLM), such as GPT-4V and LLaVA, as their backbone. While this enables strong performance on perception-driven tasks, it largely overlooks the integration of multisensory inputs such as auditory, tactile, or motion cues, which are essential for embodied spatial intelligence in AI agents~\cite{liu2025neural}. \textbf{(2)} The reliance on pre-trained models limits the development of a genuine internal world representation, a structured of the environment that can shift between egocentric and allocentric perspectives~\cite{xing2025critiques}, akin to the cognitive maps formed in the human brain. \textbf{(3)} The memory systems of current AI agents face limitations as they depend heavily on the scale of large models rather than on mechanisms for retrieval, contextualization, and long-term adaptation, which are fundamental to how the human brain organizes and reuses past experience~\cite{liu2025neural}.

\textit{\textbf{Contributions}}: Given the aforementioned challenges and the lack of representative works that provide a basic framework or theoretical foundation with detailed analysis, \textbf{the research and design of the AI spatial reasoning capable of mimicking human spatial cognition remain largely unexplored}. To address this problem, we introduce a \textbf{conceptual computational framework grounded in the neuroscience principle of human spatial cognition}, paving the way for the development of human-like spatial intelligence for AI agents. 
As described in Fig.~\ref{fig:framework}, the design of human-like spatial intelligence can be viewed as a biological extension of human spatial cognition, emerging at the intersection of neuroscience, cognitive science, and machine learning.
Just as humans rely on multimodal perception, integration, and memory to form cognitive maps that guide decision-making, agents require analogous components to operate effectively in complex environments. These include \textbf{bio-inspired multimodal sensing} for capturing diverse inputs, \textbf{multi-sensory integration} for fusing signals into coherent representations, and mechanisms for \textbf{egocentric–allocentric conversion} that enable flexible spatial reasoning. At the core, an artificial \textbf{cognitive map} supported by \textbf{spatial memory} provides the internal model upon which \textbf{ spatial reasoning and planning} can unfold, thereby forming the core components of our framework to facilitate spatial reasoning in dynamic environments. 
Together, these components form a conceptual roadmap for designing AI agents capable of human-like spatial reasoning and adaptive behavior.

By drawing inspiration from how the human brain reasons in space, our bio-inspired framework aims to provide a \textbf{perspective landscape} for the research community, highlighting the essential components required for agents to achieve better interaction with the physical 3D world. To reflect the biological view, we first establish the theoretical basis from neuroscience that can effectively function as the backbone for bio-inspired spatial reasoning (Sec.~\ref{sec:2.A}). Building on this foundation, we introduce a computational framework that outlines the key modules, highlights potential techniques for their implementation, and conceptualizes them as an integrated pipeline for spatial reasoning in agents (Sec.~\ref{sec:2.B}). On top of our  framework, we comprehensively analyze existing works relevant to each module, highlight their potential contributions to the module’s function, and identify key gaps that must be addressed toward better spatial reasoning module design (Sec.~\ref{sec:3.A}). Then, we examine emerging benchmarks and datasets for spatial reasoning evaluation, followed by a discussion of potential applications across both virtual and physical environments (Sec.~\ref{sec:3.B}). Moreover, we provide the future direction based on the limitations of recent works (Sec.~\ref{sec:4}). Finally, we conclude this perspective work in Sec.~\ref{sec:5}. In summary, our major contributions are as follows:

\vspace{-0.1em}
\begin{itemize}
    \item \textbf{We propose a neuroscience-inspired spatial reasoning framework for agentic AI. } To the best of our knowledge, this is the first work to explore the neuroscience-based framework design for agentic spatial reasoning. Our proposed framework lays the foundation to achieve the human-like intelligence in AI spatial reasoning. 
    
    \item \textbf{We conduct a framework-guided systematic analysis of recent spatial reasoning methods.} Unlike prior works, we conduct a comprehensive review of the representative research on AI spatial intelligence across our designed framework, enabling a clear understanding of the gap between current research and the ideal of human-like spatial reasoning. 
    \item \textbf{We highlight the key limitations of agentic spatial reasoning and advocate some future directions based on our proposed framework. } Following the analysis, we systematically highlight key limitations in current spatial reasoning models, thereby providing the future research direction towards the human intelligence in agents. 
\end{itemize}

\vspace{-12pt}
\section{Neuroscience-Inspired Agentic Spatial Intelligence}
\label{sec:2}

In this section, we revisit the neuroscience foundations of spatial cognition, covering (1) the neural mechanisms of key brain regions involved in spatial reasoning, (2) the cognitive map theory, (3) spatial memory systems, and (4) emerging computational neuroscience models relevant to spatial reasoning (Sec.~\ref{sec:2.A}). Building on this strong foundation, we propose a generic computational framework alongside the analysis of the types of spatial reasoning behaviors essential for agents and the pathways for its implementation (Sec.~\ref{sec:2.B}).
\subsection{Spatial Cognition From Neuroscience}
\label{sec:2.A}

\subsubsection{Neural mechanisms for spatial processing}
\label{sec:2.A.1}

In the human brain, spatial reasoning has been conceptualized as a sequence of neural computations that transform sensory input into abstract spatial representations used for decision-making~\cite{byrne2007remembering, maocom}. In this context, neuroscience research follows a general framework that captures the spatial processing as a hierarchical flow, commencing from early sensory integration to egocentric and allocentric map, then finally high-order reasoning~\cite{byrne2007remembering}. 
This framework reflects how spatial cognition is functionally organized across brain regions, forming a coherent basis to analyze its neural mechanisms.

\textit{Sensory Integration.} 
The brain's ability to perceive and navigate the world relies on the process of combining information from different sensory modalities to form a unified representation of space. Classical neuroscience defines this integration as a fundamental function of higher cortical areas, allowing the brain to resolve ambiguities and enhance perceptual accuracy by leveraging complementary inputs~\cite{Ghazanfar2006Multisensory}. 

\textbf{Visual Sensing} plays a dominant role in humans, guiding both perception and action. Visual information is initially processed in the primary visual cortex (V1) of the occipital lobe, where neurons respond to basic features like orientation, spatial frequency, and motion direction. Among them, the dorsal stream, projecting to the posterior parietal cortex (PPC), is involved in motion detection, spatial localization, and visually guided action~\cite{ungerleider1982two}. The ventral stream projects to the inferior temporal cortex (ITC) and is specialized for identifying and making sense of visual information, allowing us to perceive the shape, color, and texture of objects~\cite{ungerleider1982two}.

\textbf{Auditory sensing} relies on cortical and subcortical pathways centered in the temporal lobe. Sound features (e.g., pitch, loudness) are first decoded in the primary auditory cortex (A1) and refined in secondary areas, while subcortical structures like the inferior colliculus (IC) analyze binaural cues for spatial hearing~\cite{bizley2013what,middlebrooks1991sound}. Specifically, interaural time differences (ITD) localize low-frequency sounds, whereas interaural level differences (ILD) guide high-frequency localization.

\textbf{Tactile sensing} is processed primarily in the parietal lobe, where haptic inputs are first handled by the primary (S1) and then the secondary somatosensory cortex (S2) for higher-level interpretation and multisensory integration~\cite{Ghazanfar2006Multisensory}.
Object features are processed along partially segregated pathways: orientation information follows a dorsal stream toward parietal areas for spatial planning, while texture information engages a ventral stream that converges with visual regions, enabling early cross-modal integration of tactile and visual cues.

Overall, human spatial perception arises from integrating visual, auditory, and haptic inputs, which are initially encoded in egocentric coordinates and then converge in the PPC to support spatial awareness, motor planning, and egocentric mapping.

\setlength{\textfloatsep}{8pt}
\begin{table}[t]
\centering
\vspace{-1em}
\caption{Biological functions of corresponding regions of the brain}
\vspace{-1em}
\label{table:neural_mechanism}
\resizebox{\textwidth}{!}{%
\begin{tabular}{@{}l|l|l|l@{}}
\toprule
\textbf{Category} & \textbf{Type} & \textbf{\makecell[l]{Brain\\ Regions}} & \textbf{\makecell[c]{Biological Mechanisms}} \\
\midrule

\multirow{7}{*}{\makecell[l]{Multi-\\Sensory\\Integration}} 
& \multirow{2}{*}{Visual}   & Visual Cortex (V1)     &  $\bullet$ Detection of spatial frequency, orientation and motion. \\
&                           & Inferior Temporal Cortex (ITC)    & $\bullet$ Perceiving shape, color and texture. \\
\noalign{\vskip 2pt}\cline{2-4}\noalign{\vskip 2pt}
& \multirow{1}{*}{Tactile}  & Somatosensory Cortex (S1/S2)  &  $\bullet$ Process haptic signals and integrate with other senses. \\
\noalign{\vskip 2pt}\cline{2-4}\noalign{\vskip 2pt}
& \multirow{4}{*}{Auditory} & Auditory Cortex (A1)     & $\bullet$ Process pitch, loudness, and direction for sound localization. \\
&                           & Interaural Time Difference (ITD)    & $\bullet$ Locate low-frequency sounds \\
&                           & Interaural Level Difference (ILD)    & $\bullet$ Locate high-frequency sounds \\
&                           & Inferior Colliculus (IC)     & $\bullet$ Analyze binaural cues \\
\midrule

\multirow{2}{*}{\makecell[l]{Egocentric\\Map}} 
& \multirow{2}{*}{\makecell[l]{Spatial\\Encode}} & Posterior Parietal Cortex (PPC) & $\bullet$ Egocentric stream integration \\
&  & Retrosplenial Cortex (RSC) &  $\bullet$ Directional orientation via head-direction cells. \\
\midrule

\multirow{4}{*}{\makecell[l]{Allocentric\\Map}} 
& \multirow{3}{*}{\makecell[l]{Place\\Cells}} & Hippocampus (HPC) & $\bullet$ Align with place cells under contextual and location signals. \\
&                                             & CA1 & $\bullet$ Pattern completion \\
&                                             & CA3 & $\bullet$ Pattern separation \\
\noalign{\vskip 2pt}\cline{2-4}\noalign{\vskip 2pt}
& \makecell[l]{Grid\\Cells}                   & Medial Entorhinal Cortex (MEC) & $\bullet$ Enable grid cells with firing patterns and self-motion encoding. \\
\midrule

\multirow{6}{*}{\makecell[l]{Spatial\\Reasoning\\\&\\Decision\\Making}} 
& \multirow{6}{*}{Inference} & Prefrontal Cortex (PFC)   & $\bullet$ Maintain reasoning process \\
& & Medial Prefrontal Cortex (mPFC)  & $\bullet$ Recall the location of targets \\
& & Dorsomedial Prefrontal Cortex (dmPFC) & $\bullet$ Route planning \\
& & Dorsal Anterior Cingulate Cortex (dACC)  & $\bullet$ Alternative route searching \\
& & Lateral Prefrontal Cortex (lPFC)  & $\bullet$ Replanning \\
& & Orbitofrontal Cortex (OFC)   & $\bullet$ Managing inference process \\
\bottomrule
\end{tabular}%
}
\end{table}

\textit{Egocentric Map.} Humans predominantly perceive and interact with their environment from a first-person perspective, relying on egocentric frames of reference to encode and recall spatial information. Egocentric maps represent space relative to the observer’s current position and orientation, emphasizing subject-to-object relationships that give rise to body-centered spatial reasoning tasks such as path integration, obstacle avoidance, and goal-directed action~\cite{burgess2008spatial}. 
In the human brain, egocentric spatial representations are primarily supported by the Caudate Nucleus and, more generally, by the Medial Parietal Lobe regions \cite{burgess2008spatial}. Among these, the Posterior Parietal Cortex (PPC) integrates egocentric information to construct a first-person perspective of space, supported by regions included in the parietal lobe~\cite{burgess2008spatial}. Additionally, directional orientation is supported by head-direction cells located in the Retrosplenial Cortex (RSC), which help encode heading information relative to the environment~\cite{shine2016human}.

\textit{Allocentric Map.} 
The allocentric map refers to a world-centered spatial representation in the brain that encodes the positions of objects relative to each other, independent of the observer’s current location or orientation~\cite{o1978hippocampus}. From a neuroscience perspective, its neural functions are primarily mediated by the Hippocampal-Entorhinal system~\cite{keefe1971hippocampus,hafting2005microstructure}. This neuron system encodes the spatial relationships between external cues in a manner that is independent of the animal’s current location or orientation. Anatomically, it involves reciprocal connectivity between the Hippocampus (HPC) and the Medial Entorhinal Cortex (MEC), where distinct classes of spatially modulated neurons contribute to the encoding and integration of a world-centered spatial representation.

The \textbf{Medial Entorhinal Cortex} (MEC) is critical for the allocentric map by containing grid cells, a type of neuron that exhibits a distinctive firing pattern when humans navigate across physical space. Each grid cell becomes active at multiple locations that form a regular, hexagonal grid, thereby creating a structured internal map of the surrounding environment~\cite{hafting2005microstructure,doeller2010evidence}. This periodic firing supports the brain in measuring distance and position by encoding movement through self-motion signals~\cite{hafting2005microstructure}. Hence, grid cells within MEC provides a continuous metric foundation for the brain, which is subsequently integrate with HPC for building the allocentric spatial map. 

Within the \textbf{Hippocampus} (HPC), the CA1 and CA3 subfields are key regions involved in allocentric spatial representation~\cite{o1978hippocampus}. These areas contain place cells, which are pyramidal neurons that fire selectively when an animal or human occupies a specific location in an environment, regardless of its heading direction~\cite{moser2008place}. In CA3, place cell activity arise from recurrent connectivity, supporting pattern completion and stable encoding of familiar spatial contexts~\cite{moser2008place, rolls2013mechanisms}. In contrast, CA1 place cells integrate input from both CA3 and the MEC, enabling the pattern seperation capability in distinguishing the overlapped spatial inputs~\cite{rolls2013mechanisms}, while maintaining contextual modulation~\cite{moser2008place}. Therefore, hippocampal place cells operates in conjunct with grid cells, supporting the context-aware and location-specific in the allocentric map.

While grid cells within MEC offer a continuous coordinate system for estimating position and distance, place cells in HPC bind this metric to discrete context-aware spatial experiences. This integration forms the spatial representation widely known in neuroscience as the cognitive map, a foundational theory in spatial cognition that will be comprehensively examined in Sec.~\ref{sec:2.1.2}. The resulting map serves as the key component for higher-order spatial reasoning tasks, bridging low-level encoding with complex cognition behaviours.

\textit{Spatial Reasoning and Decision-Making}
Spatial reasoning and decision-making arise from hippocampal–cortical circuits that integrate sensory inputs into internal representations, relay them via the RSC to the PPC, PFC, and then to cortical regions for action, and encode goal-related rewards to guide adaptive behavior~\cite{2017187}.
The prefrontal cortex (PFC), as the brain’s executive control center, supports spatial reasoning by managing working memory and engaging distinct regions for different stages of spatial behaviors~\cite{patai2021versatile}. Taking navigation as an example, the medial prefrontal cortex (mPFC) recalls the location of the target, which is then passed to the dorsomedial prefrontal cortex (dmPFC) for route planning. Subsequently, if a detour is necessary, the dorsal anterior cingulate cortex (dACC) searches for alternative routes. Following this, the lateral prefrontal cortex (lPFC) replans the route and avoids incorrect shortcuts. During this process, the Orbitofrontal Cortex (OFC) selects specific paths and infers their potential outcomes, while the mPFC represents the proximity or distance of the path. When the navigator goes beyond the target range, the dACC assists them in returning along the original path.
In summary, as a higher-order brain region, the main role of the PFC is the operator of the cognitive map. It receives spatial information from the HPC and then flexibly applies it for planning, inference, and prediction.

\subsubsection{Cognitive map theory}
\label{sec:2.1.2}
\begin{figure*}[t]
    \centering
    \vspace{-1em}
    \includegraphics[width=0.8\linewidth]
    {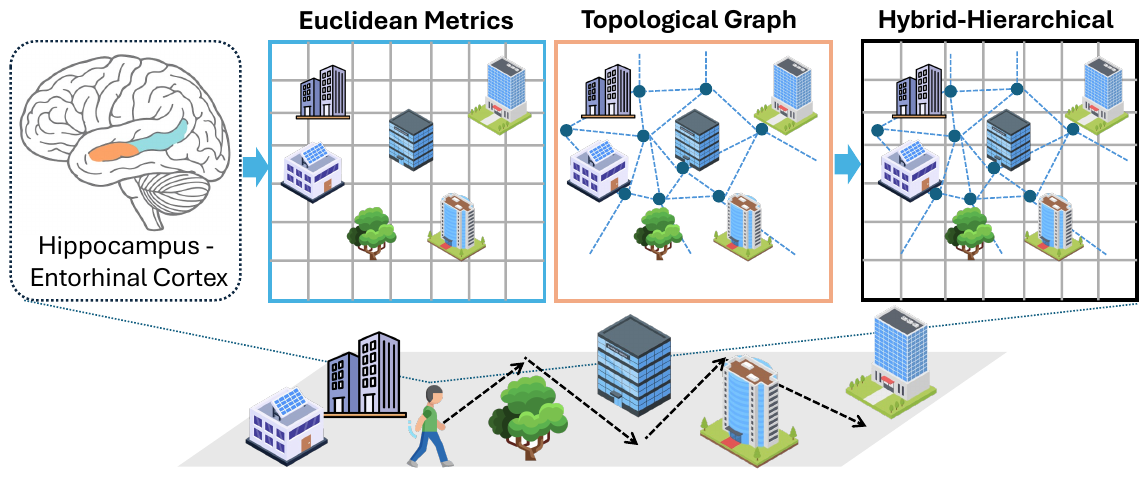}
    \vspace{-1em}
    \caption{The neuroscience-based cognitive map. It is rooted in the hippocampus (orange) and entorhinal cortex (blue), where the entorhinal cortex encodes Euclidean metrics and the hippocampus encodes relational topological graphs, together forming integrated hybrid and hierarchical maps.}
    \label{fig:cognitivemap}
    \Description{cognitive}
\end{figure*}

As a foundation of spatial cognition, Tolman \textit{et al.}~\cite{tolman1948cognitive} first proposed the concept of cognitive map based on the spatial navigation behavior of rats in maze tasks. 
In both humans and animals, this representation is a subjective internal form of external information and objective knowledge, which is functionally analogous to geographically created maps.
As illustrated in Fig.~\ref{fig:cognitivemap}, current theoretical debates primarily involve two hypotheses: the Euclidean map and the topological graph, representing opposite extremes regarding the precision of representation~\cite{tolman1948cognitive}. The Euclidean map demands the accurate representation of spatial coordinates, while the topological graph contains no metric information. As both have limitations in explaining spatial behavior, researchers have developed integrated metric–topological cognitive maps and more recently explored hierarchical models.

\textit{Euclidean Map.}
Building on Tolman's concept of the cognitive map, O'Keefe and Nadel further proposed that the cognitive map possesses a unified Euclidean metric structure~\cite{o1978hippocampus}. The primary physiological evidence supporting this proposal comes from the discovery of place cells in the HPC and grid cells in the MEC (described in Sec.~\ref{sec:2.A.1}). These two cell types provide a metric background and landmark anchoring for the spatial environment.
The Euclidean cognitive map is founded on an allocentric  reference frame, meaning it is not affected by the navigator's own embodied experience~\cite{o1978hippocampus}. It functions like a cartographic map that preserves geometric relations between locations, enabling flexible navigation through novel paths and shortcuts, with pointing performance further indicating that spatial updating operates on Euclidean geometry~\cite{o1971hippocampus, widdowson2022human}.
However, rather than conforming to strict Euclidean rules, human spatial representations appear to follow more adaptive, context-driven geometries~\cite{tversky1981distortions}. 

\textit{Topological Graph.}
Inspired by the memory space theory~\cite{eichenbaum1997declarative}, the topological graph hypothesis has been proposed as an alternative form for the cognitive map~\cite{kuipers1978modeling}. Unlike the precise metrics of a Euclidean map, the topological graph hypothesis suggests that the cognitive map only encodes a rough topological structure. An environment's topological structure is a network composed of nodes and the connecting edges between them. Generally, nodes represent locations, while edges represent the path relationships between these locations. This ``place graph'' captures the connectivity between locations but does not require embedding them into a globally consistent coordinate system (coordinate-free). Additionally, nodes can also represent specific views or local areas; correspondingly, edges represent the actions required to transition between different views or areas~\cite{jacobs2003unpacking}. A topological graph balances between route memory and Euclidean maps by flexibly representing multiple paths in a compressed, non-metric structure that reduces memory demands.
However, its lack of metric detail makes it insufficient to fully explain human spatial navigation.

\textit{Hierarchical Map.} 
While Euclidean and topological map theories treat locations as uniformly represented, real-world environments often have nested structures, \textit{e.g.,} rooms within buildings or buildings within a campus~\cite{meilinger2016qualitative}. To build a coherent cognitive map, the brain must integrate spatial information from these discrete regions. This suggests that cognitive maps may have a hierarchical structure, where fine-grained local details are stored at lower levels and abstract, generalized layouts at higher levels~\cite{hirtle1985evidence}. In such a structure, different regions are represented as separate branches, allowing the brain to manage complexity by organizing space into meaningful subunits. This helps reduce cognitive load while supporting flexible navigation across different spatial scales. 

\textit{Hybrid Map.} 
Hybrid cognitive maps combine Euclidean, topological, and hierarchical representations, providing a flexible, biologically plausible framework that better reflects the variability and adaptability of human spatial behavior than traditional single-paradigm models.
They embed locally metric topological graphs within a global metric space while enabling multi-scale abstraction without enforcing strict hierarchical boundaries~\cite{baumann2023metric}. This fusion allows for context-sensitive reasoning, balancing efficiency in planning with biological plausibility in representation. 
A typical representative of a hybrid cognitive map is the hybrid graph model, where edges encode path lengths and nodes capture angular relations, enabling flexible navigation, such as shortcuts and detours, even without globally consistent coordinates. By integrating hybrid aspects of graph-based structures and cognitive maps, the theory provides a unified framework that overcomes the limitations of relying solely on either representation or treating them independently.

\subsubsection{Spatial Memory Systems}
\label{sec:2.A.3}

\begin{wrapfigure}[14]{r}{0.42\columnwidth}
  \centering
  \includegraphics[width=\linewidth]{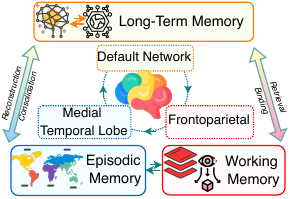}
  \caption{Memory systems in human cognition: working, episodic, and long-term.}
  \Description{Schematic with three blocks labelled working, episodic, and long-term memory.}
  \label{fig:neuro_mem}
\end{wrapfigure}

Spatial memory remains a foundational role in human spatial reasoning, enabling individuals to represent, manipulate, and navigate within their environments. From a neuroscience perspective, spatial reasoning emerges not solely from perception or motor planning, but also from the integration and coordination of memory systems that operate across the brain in multiple timescales. As described in Fig.~\ref{fig:neuro_mem}, neuroscientists categorize these into working memory, episodic memory, and long-term semantic memory, in which each corresponds to different functional networks and cognitive roles. 


\textit{Working Memory.}
In the human brain, working memory is a short-term cognitive mechanism that temporarily maintains and manipulates task-relevant information. Specifically, it enables the brain to keep the observed information accessible over short periods, supporting mental operations such as planning and decision-making when the sensory input is no longer perceptually available. Functionally, this memory system relies on persistent neural activity rather than long-term synaptic changes~\cite{goldman1995memory}, making it highly flexible but also sensitive to cognitive load and interference. 
It operates over a timescale of seconds and acts as a dynamic buffer for integrating real-time sensory input with internal task goals. 

Neuroscientific research has localized spatial working memory primarily to the \textbf{frontoparietal network}, with key contributions from the dlPFC and PPC~\cite{d2015fronto}. These regions coordinate attentional control, goal maintenance, and sensorimotor integration, supported by sustained delay-period activity in prefrontal and parietal neurons. The PPC encodes spatial attention and transient location cues, while the dlPFC manipulates and updates this information in line with behavioral goals.
Given the neural basis in the frontoparietal network, working memory effectively supports spatial reasoning by maintaining egocentric spatial representations (i.e., current position, heading direction, and surrounding cues) through its reliance on the dlPFC and PPC, whose roles in egocentric processing were outlined in Sec.~\ref{sec:2.A.1}. 
It functions as an online buffer for short-term spatial inference and decision-making, allowing individuals to simulate movement, retain temporary goals, and update self-location during navigation. These abilities are crucial when external spatial cues are lacking, relying on internally maintained egocentric information. 

\textit{Episodic Memory.}
Tulving~\cite{tulving1983elements} defined episodic memory as the system that encodes, stores, and retrieves autobiographical experiences within specific spatiotemporal contexts. It supports episodic recollection by enabling the mental recall of the experienced events with both spatial and temporal contexts (e.g., ``where and when they occurred'').

Biologically, episodic memory is primarily supported by the \textbf{medial temporal lobe} (MTL), which includes the HPC, MEC, parahippocampal cortex, and perirhinal cortex~\cite{eichenbaum2007medial}. Building upon the HPC and the MEC, as described in Sec.~\ref{sec:2.A.1}, this network enables the encoding and retrieval of experiences embedded in space and time. The HPC via place cells plays a central role in binding spatial and temporal elements with contextual-awareness into unified memory traces, while the MEC not only provides spatial metrics through grid cells but also channels multimodal sensory information toward the hippocampus~\cite{moser2008place}. Complementing this core loop, the parahippocampal cortex encodes scene-based contextual features (i.e., environmental layouts and spatial geometry) and the perirhinal cortex contributes object-related content and familiarity signals~\cite{eichenbaum2007medial}. 
Moreover, the MTL consolidates episodic traces from the HPC into neocortical regions, forming stable semantic knowledge that preserves spatial experiences, with episodic memory enabling recall and recombination of past events to support spatial reasoning~\cite{hassabis2007deconstructing}.

\textit{Long-Term Memory.}
From a neuroscience standpoint, long-term memory refers to the brain’s capacity to store information over extended periods, ranging from hours to a lifetime. Unlike working or episodic memory, which deal with short-term access or specific experiences, long-term memory supports the accumulation of abstracted, semantic knowledge about the world~\cite{tulving1983elements}. This includes spatial schemas, landmark familiarity, navigational rules, and environmental regularities, all of which contribute to efficient spatial reasoning across diverse contexts~\cite{byrne2007remembering}.

A central neural substrate of long-term semantic memory is the \textbf{default network}, including the medial prefrontal cortex (mPFC), posterior cingulate cortex (PCC), angular gyrus, and medial temporal lobe (MTL) sub-systems, which is consistently engaged during internally directed cognition (e.g., mental simulation, and prospective thinking)~\cite{buckner2008brain}.
To support the corresponding neural functions, the default network coordinates with the hippocampus and frontoparietal systems, enabling the consolidation and abstraction of past spatial experiences into task-relevant knowledge through interactions with working and episodic memory~\cite{buckner2008brain}. 
Within this network, the mPFC encodes high-level schemas and behavioral rules, while the angular gyrus integrates multimodal concepts and retrieves context-rich memories. The PCC serves as a hub linking episodic traces in the medial temporal lobe with semantic representations in the neocortex. Through repeated reactivation, such as during rest or sleep, episodic content is transformed into generalized priors that guide perception, planning, and goal-directed action~\cite{dudai2015consolidation}. This semantic foundation allows the brain to support spatial reasoning not by simulating space directly, but by reusing generalized priors to guide perception, planning, and goal-directed action.

\begin{figure*}[t]
    \centering
    \vspace{-1em}
    \includegraphics[width=\linewidth]{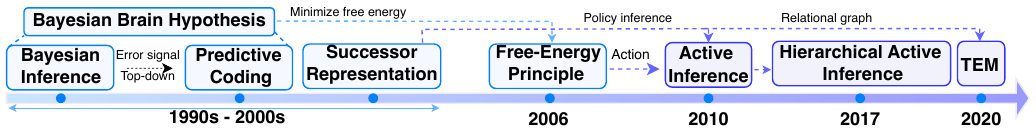}
    \vspace{-2em}
    \caption{Development of the backbone neuroscience models for spatial reasoning.}
    \label{fig:neuro_timeline}
    \Description{time}
\end{figure*}

\subsubsection{Computational Models from Neuroscience}
\label{sec:2.A.4}

To understand how the biological brain performs spatial reasoning, neuroscience provides a rich set of computational models that describe how the brain perceives, represents, and acts upon spatial information. These models have long served as a conceptual bridge to AI research, where both foundational and emerging neuroscience models have inspired the development of AI algorithms~\cite{zinan2025nature}. Following that, the neuroscience models potentially provide the lens for analyzing the spatial intelligence in agents, offering structured ways to understand perception, internal representations, embodied cognition, and planning in dynamic environments. 

Serving as a theoretical foundation for many brain models, Bayesian inference provides the foundation for viewing the brain as an inference engine, leading to the Bayesian Brain Hypothesis~\cite{knill2004bayesian} and its operationalization through Predictive Coding~\cite{rao1999predictive}. This framework was generalized by the Free Energy Principle (FEP)~\cite{friston2006free} and extended by Active Inference~\cite{friston2017graphical} to include action, culminating in Hierarchical Active Inference (HAI)~\cite{friston2017graphical}, which overlaps with Successor Representation in predictive planning. Recent models such as the Tolman–Eichenbaum Machine (TEM)~\cite{whittington2020tolman} build on these ideas to link neural inference with AI, particularly for spatial cognition. A general timeline of these theoretical developments is illustrated in Fig.~\ref{fig:neuro_timeline}.

\textit{Bayesian Inference.}
As the core of emerging neuroscience models, Bayesian Inference offers a foundational theory for understanding the information processing within the brain in uncertainty conditions~\cite{knill2004bayesian}, which can be formulated as:
\begin{equation}
    P(\mathcal{H} \mid \mathcal{E}) = \frac{P(\mathcal{E} \mid \mathcal{H}) \cdot P(\mathcal{H})}{P(\mathcal{E})}
\end{equation}
where $P(\mathcal{H} \mid \mathcal{E})$ is the posterior probability, demonstrating the updated belief about the hypothesis $\mathcal{H}$ after observing the evidence $\mathcal{E}$. Meanwhile, $P(\mathcal{E} \mid \mathcal{H})$ is the likelihood, indicating the probability of the observed evidence $\mathcal{E}$ with the given hypothesis $\mathcal{H}$. The $P(\mathcal{H})$ is the prior probability, illustrating the internal belief of the brain regarding the hypothesis $\mathcal{H}$ before considering new evidence. Finally, $P(\mathcal{E})$ is the evidence probability, a normalization factor that ensures all possible hypotheses' probabilities sum to one.

\textit{Predictive Coding.} In the context of spatial reasoning, predictive coding facilitates forward inference \cite{penny2013forward}, enabling the brain to anticipate sensory consequences of self-motion, integrate multimodal spatial cues, and resolve ambiguity in localization and goal-directed planning by refining its spatial belief map in real time.

Predictive coding posits that the brain maintains internal models of spatial structure and continuously minimizes the prediction error $\epsilon_{t}=x_{t} - \hat{x}_{t}$ between the incoming sensor input $x_{t}$ and top-down spatial predictions $\hat{x}_{t}$. This error drives the update of latent spatial beliefs $H$
through gradient descent on the variational free energy $F$: $$\frac{dH}{dt} \propto -\frac{\partial F}{\partial H}$$


\textit{Successor Representation.} 
The Successor Representation (SR) theory proposes that Euclidean and topological maps are two forms of the same neural encoding, with the HPC representing current states along with future-reachable states and their probabilities~\cite{stachenfeld2017hippocampus}. Hence, topological graphs encode states as nodes with weighted transitions, while Euclidean maps anchor states to spatial positions connected by paths.

The SR matrix \( M(s, s') \) is defined as: the expected discounted cumulative number of future visits to state \( s' \) starting from the current state \( s \) (following a specific policy). In other words, \( M(s, s') \) measures the ``reachability'' or ``predictiveness'' of \( s' \) from \( s \).
 \[ M(s, s') = \mathbb{E}\left[ \sum_{t=0}^{\infty} \gamma^t \cdot \mathbb{I}(s_t = s') \mid s_0 = s \right] \]

\noindent where \( \gamma \) is the discount factor, \( \mathbb{I} \) is the indicator function (1 if \( s_t = s' \), 0 otherwise), and \( \mathbb{E} \) denotes expectation.

The value of a state \( V(s) \) (i.e., the cumulative reward obtainable from state \( s \)) can be decomposed into the inner product of the SR and the reward function \( R(s') \) (the immediate reward in state \( s' \)):
\[ V(s) = \sum_{s'} M(s, s') \cdot R(s') \]

\textit{Free Energy Principle.} 
The Free Energy Principle (FEP), as formulated by Friston et al.~\cite{friston2006free} , states that the brain maintains its internal states and structure by minimizing a quantity known as \textbf{variational free energy} $F$, which serves as a tractable bound on sensory surprise or negative log evidence $-\log P(D)$. Given sensory data $D$ and hidden environmental causes $H$, the brain constructs a generative model 
\begin{equation}
\label{eq:genmodel_FEP}
    P(D, H) = P(D \mid H) P(H),
\end{equation}
and seeks to approximate the true posterior $P(H \mid D)$ using a recognition density $Q(H)$. Since direct computation of $P(H \mid D)$ is generally intractable due to the marginal likelihood 
\begin{equation}
    P(D) = \int P(D, H) \, dH,
\end{equation}
variational inference is used to minimize the Kullback–Leibler (KL) divergence between $Q(H)$ and $P(H \mid D)$:
\begin{equation}
    F[Q] = D_{\mathrm{KL}}(Q(H) \parallel P(H \mid D)) - \log P(D) \geq -\log P(D).
\end{equation}
This variational free energy can be equivalently expressed as:
\begin{equation}
\label{eq:vfe}
    F[Q] = \mathbb{E}_{Q}[\log Q(H)] - \mathbb{E}_{Q}[\log P(H, D)],
\end{equation}
highlighting that free energy is minimized when the approximate posterior $Q(H)$ matches the true posterior and when the generative model accurately predicts sensory input. 
These dynamics explain how the brain updates beliefs and selects actions to reduce free energy and thereby resist entropy. Furthermore, Friston~\cite{friston2010free} extends the theory to hierarchical generative models, aligning with empirical architectures in cortical hierarchies, where top-down predictions and bottom-up prediction errors are exchanged between layers. 

\textit{Hierarchical Active Inference.}
Building upon the Free-Energy Principle (FEP), Friston \textit{et al.}~\cite{friston2010free} introduces Active Inference theory by treating both perception and action as processes that minimize variational free energy. While FEP describes how the brain maintains homeostasis by suppressing sensory surprise, active inference operationalizes this through continual updating of internal beliefs and selection of actions that fulfill prior expectations.

\begin{figure}[t]
    \centering
    \vspace{-1em}
    \includegraphics[width=0.8\linewidth]{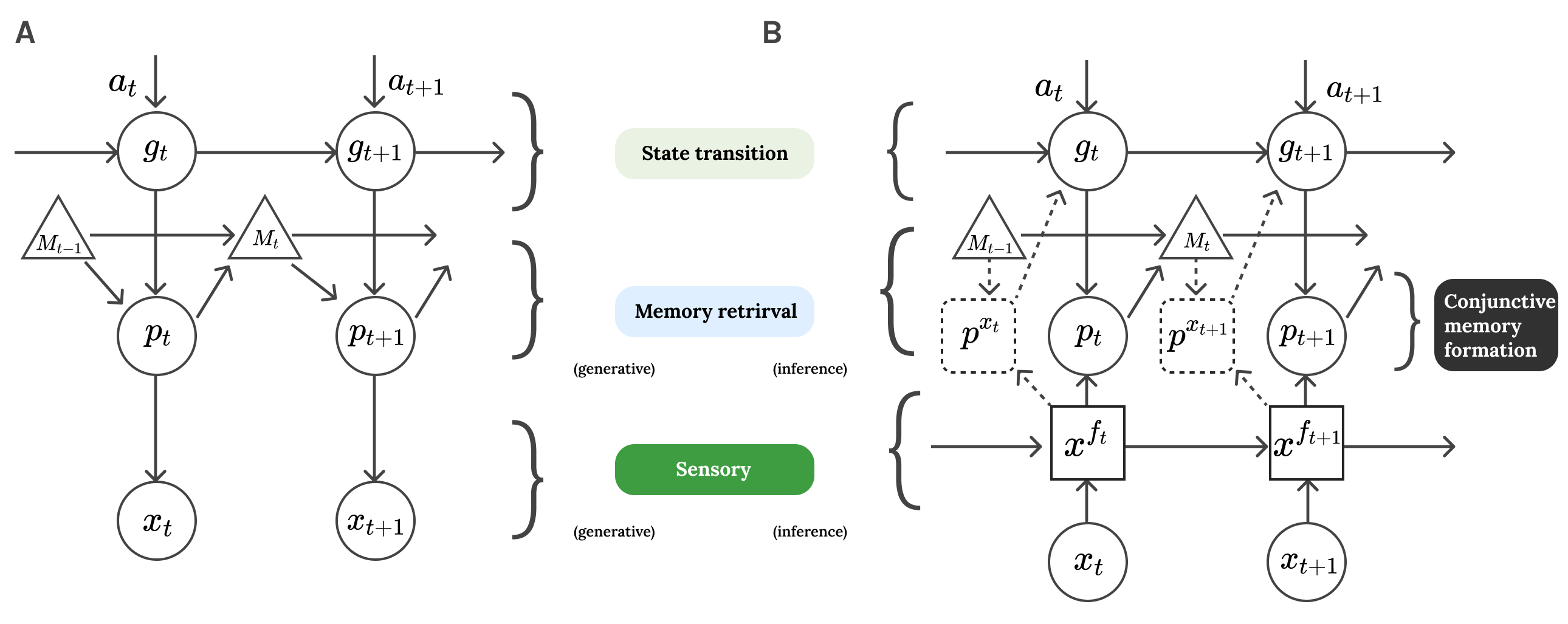}
    \vspace{-1em}
    \caption{The architecture of TEM. (A) Generative model showing the top-down process from actions ($a$) and latent states ($g$) through state transitions, memory retrieval, and temporal filtering to generate sensory predictions ($p$) and observations ($x$). (B) Inference model depicting the bottom-up inference from observations ($x$) to latent states via memory retrieval, conjunctive memory formation, and updates to memory ($M$).}
    \label{fig:TEM}
    \Description{TEM}
\end{figure}

Active Inference has frequently been analyzed by neuroscientists through the lens of hierarchical generative models~\cite{friston2017graphical}, as a means to characterize how the brain constructs and uses internal models of the world to perform inference and reasoning across multiple cortical regions. This hierarchical formulation allows for the modeling of distinct layers of processing, ranging from low-level sensory predictions to higher-order cognitive functions, within a unified probabilistic framework~\cite{pezzulo2018hierarchical}. Extending from Eq.~(\ref{eq:genmodel_FEP}) of FEP, Hierarchical Active Inference (HAI) demonstrates the internal world belief of the brain via a structured generative model, illustrated as:
\begin{equation}
\label{eq:generative_HAI}
    P(o_{1:T}, s_{1:T}, \pi) = P(\pi) \cdot P(s_1) \cdot \prod_{t=1}^{T} P(o_t \mid s_t) \cdot P(s_t \mid s_{t-1}, \pi)
\end{equation}
where \( o_{1:T} \) denotes sequences of sensory observations, \( s_{1:T} \) are the corresponding hidden neural states inferred by the brain, and \( \pi \) represents a policy encoding internal preferences or behavioral tendencies. \( P(\pi) \) is the prior over such policies; \( P(s_1) \) encodes prior beliefs about initial causes; \( P(o_t \mid s_t) \) defines the likelihood mapping from latent states to sensory input; and \( P(s_t \mid s_{t-1}, \pi) \) models state transitions modulated by internal policy dynamics.

Based on this hierarchical generative model, the brain constantly strives to minimize variational free energy, as mentioned in Eq.~(\ref{eq:vfe}), acting as an upper bound on sensory surprise. This optimization principle drives the system to either update internal beliefs or initiate actions in response to mismatches between predicted and actual sensory input.
Despite this unified principle, Friston's framework functionally layers this generative model into three hierarchical cognitive processes: Perception, Belief Updating, and Policy Selection~\cite{friston2010free}.
\begin{itemize}
    \item \textbf{Perception} focuses on inferring low-level hidden states $s_t$ that best causes sensory observations $o_t$, primarily through the likelihood mapping $P(o_t \mid s_t)$.
    \item \textbf{Belief Updating} involves continually revising internal representations of latent states $s_{1:T}$, guided by prior knowledge $P(s_1)$ and the dynamic evolution of states $P(s_t \mid s_{t-1}, \pi)$. This refinement allows the brain to move beyond immediate sensory explanations and infer deeper structural, contextual, and abstract properties of the environment over time.
    \item \textbf{Policy Selection} is the brain's prospective decision-making process, where it chooses the policy $\pi$ expected to minimize future uncertainty and achieve preferred outcomes, actively shaping future state transitions $P(s_t \mid s_{t-1}, \pi)$ and subsequent sensory experiences.
\end{itemize}

\textit{TEM.} 
Tolman-Eichenbaum Machine (TEM) was proposed by Whittington \textit{et al.}~\cite{whittington2020tolman}, a framework that unifies Euclidean and topological spatial representations to generalize across environments with different sensory details.
As illustrated in Fig.~\ref{fig:TEM}, the core idea behind TEM is structural generalization with the ability to abstract latent relational structures and bind them to specific sensory inputs. This involves factorizing knowledge into separable components, enabling flexible recombination in new environments. In the model, structural abstraction occurs in the entorhinal cortex, while the hippocampus binds specific sensory details. After training, TEM exhibits neural properties analogous to grid cells, place cells, and object-vector cells, mirroring biological data in both physical and abstract spaces. These findings suggest that Euclidean and topological maps may reflect different manifestations of a common representational mechanism. TEM was further enhanced by integrating self-attention, aligning it with Transformer architectures~\cite{whittington2021relating}.

\subsection{The Proposed Generic Framework}
\label{sec:2.B}
\begin{figure*}[t]
    \centering
    \vspace{-1em}
    \includegraphics[width=\linewidth]{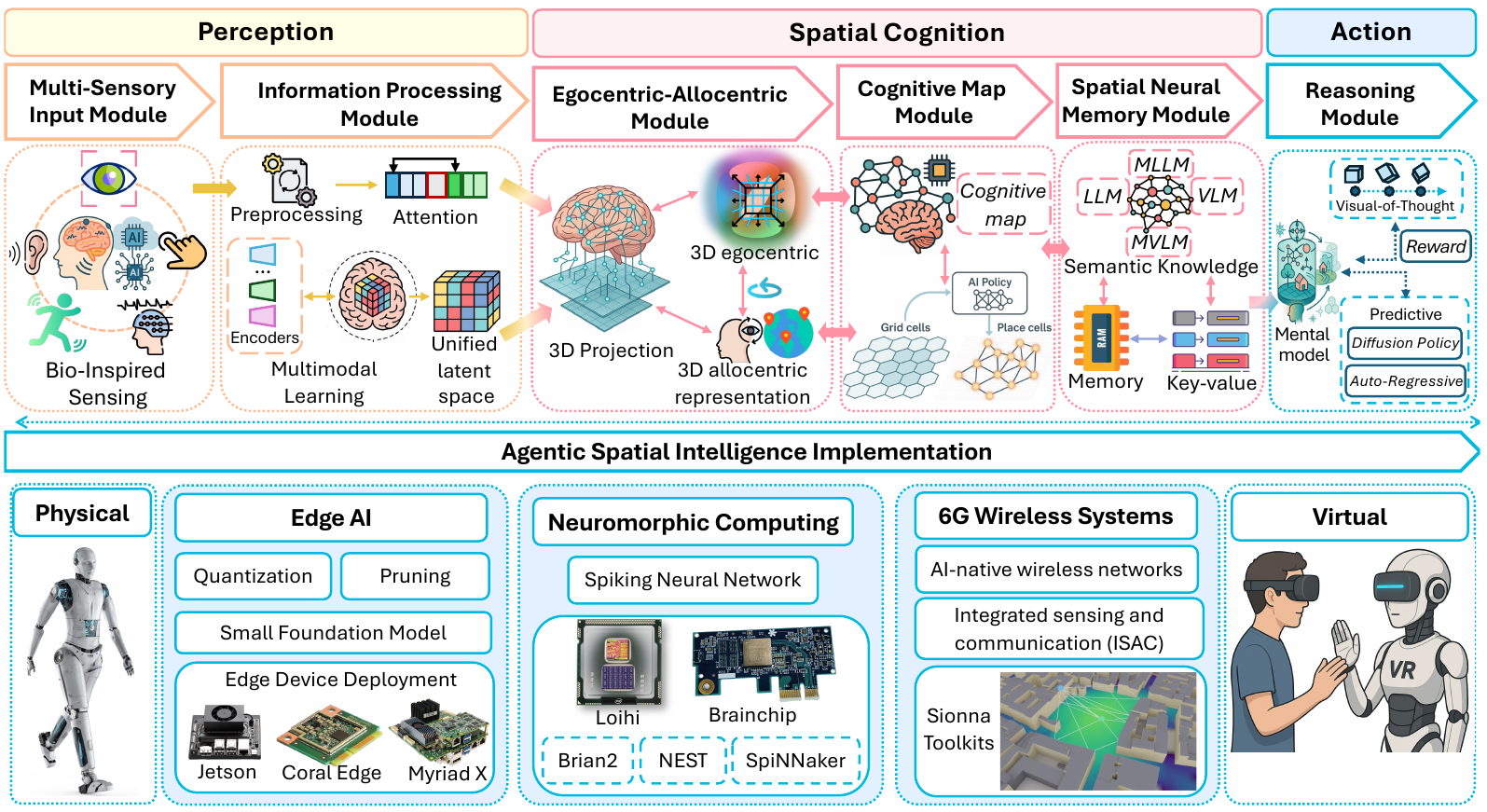}
    \caption{The proposed framework of Agentic Spatial Intelligence. Following human cognition from perception, cognition, to action, the framework contains six modules: (1) multi-sensory input, (2) information processing, (3) egocentric-allocentric, (4) cognitive map, (5) spatial neural memory, and (6) spatial reasoning.}
    \label{fig:computation}
    \Description{frame}
\end{figure*}

Inspiring from the neuroscience foundations, we propose a conceptual computation framework, serving as a perspective landscape for achieving agents with human-like spatial intelligence, as illustrated in Fig.~\ref{fig:computation}. Alongside the required modules, we emphasize the spatial reasoning behaviors based on the neuroscience models discussed in Sec.~\ref{sec:2.A.4} and subsequently discuss the system-level implementation of agents' spatial reasoning, including both emerging software and hardware. 

\subsubsection{Low-level Multimodal Sensing}
\label{sec:2.B.1}
\begin{figure*}[t]
    \centering
    \vspace{-1em}
    \includegraphics[width=0.8\linewidth]{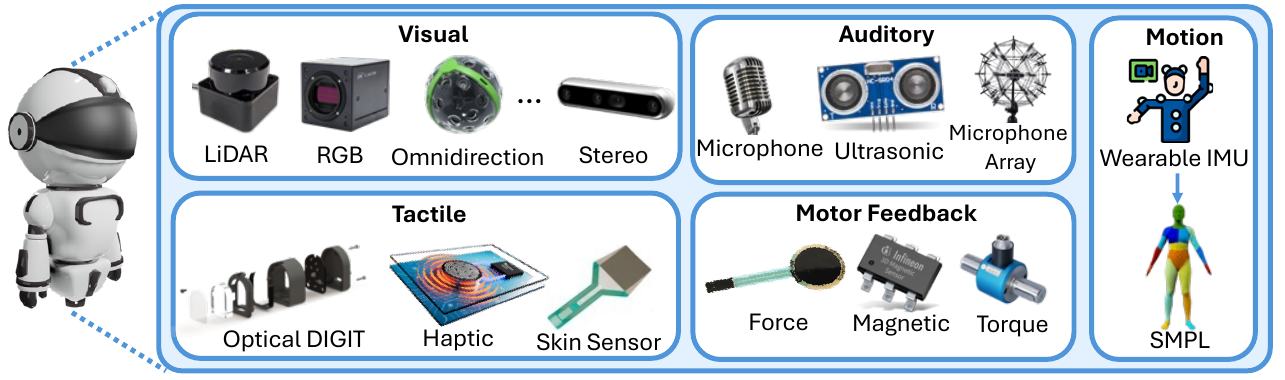}
    \vspace{-1em}
    \caption{The multi-sensory input required for agents. It consists of bio-inspired modalities: visual, auditory, tactile, bio-signals, motion, and motor feedback.}
    \label{fig:sensory}
    \Description{sensor}
\end{figure*}

The human brain is capable of processing information across multiple sensory pathways via PPC and early sensory cortices (discussed in Sec.~\ref{sec:2.A.1}). 
Therefore, to support spatial reasoning in AI agents, the low-level sensing contains multimodal sensory inputs (shown in Fig.~\ref{fig:sensory}) and an information processing module.

\textbf{(a) Multi-Sensory Input Module.} 
introduces core modalities that provide diverse input signals for spatial reasoning, including vision, auditory, tactile, motion, and motor feedback.

\textit{Vision.}
Vision-based sensing is central to spatial intelligence in embodied agents, with diverse modalities offering complementary strengths. As the primary means of perceiving scene structure and semantics, cameras provide the geometric and contextual cues that support spatial mapping, localization, and interaction. \textbf{Monocular} and \textbf{Stereo cameras} capture 2D images for mapping, navigation, and object interaction, while \textbf{Depth cameras} such as structured light and ToF sensors directly acquire dense 3D geometry for motion-robust manipulation and low-light operation. \textbf{Event-based cameras} provides microsecond latency and high dynamic range for agile motion perception. \textbf{Omnidirectional} and \textbf{Multispectral} systems expand situational awareness and robustness under varying illumination, and specialized devices such as light field and polarization cameras enable refocusing, occlusion reasoning, and material property inference. 
\textbf{LiDAR} excels at generating dense, high-resolution 3D point clouds, while \textbf{Radar} provides reliable long-range, all-weather detection that is crucial for safety-critical applications.

\textit{Auditory.} 
Inspired by the biological auditory cortex (Sec.~\ref{sec:2.A.1}), auditory sensing equips agents with temporal and spectral information for localizing sound sources, estimating distance, and interpreting acoustic cues, thereby enhancing spatial awareness and intelligence for navigation and decision-making.
\textbf{Microphones} \cite{elko2008microphone} enhances perception by extracting binaural cues (Interaural Time Difference and Interaural Level Difference) from phase and amplitude disparities. These cues, analogous to processing in the mammalian superior olivary complex and inferior colliculus, enable Direction of Arrival (DoA) estimation from time delays and intensity gradients, inspired by tonotopic mapping in the auditory cortex. \textbf{Ultrasonic sensors}~\cite{shung1996ultrasonic} on the other hand operate beyond human hearing, emitting high-frequency pulses and measuring echo time-of-flight to compute distances, similar to bat and dolphin echolocation. 

\textit{Tactile.}
Tactile sensing, inspired by the human somatosensory cortex, enables embodied agents to perceive surface geometry, texture, and contact forces, supporting spatial reasoning through touch and physical interaction.
\textbf{Haptic sensors}, analogous to skin thermoreceptors and mechanoreceptors \cite{johansson2009coding}, capture tactile properties like texture, vibration, and temperature, allowing agents to differentiate materials such as soft cloth and cold metal.
\textbf{Tactile skins}, inspired by the somatosensory homunculus, embeds dense taxel arrays on flexible surfaces to provide high-resolution contact sensing~\cite{hammock2013evolution}.
\textbf{Whisker sensors} \cite{pearson2007whisker}, inspired by the rodents' vibrissal, are flexible filaments embedded with strain gauges or bending sensors that detect contact direction, angle, and force. \textbf{Optical tactile sensors} detect deformation of a soft medium (e.g., elastomer) via internal reflections or direct imaging, with LEDs illuminating the interface and cameras tracking changes in light patterns or geometry to estimate pressure, contours, and shear forces. The \textbf{DIGIT sensor}~\cite{lambeta2020digit} integrates a deformable gel, internal LEDs, and a high-resolution camera to provide rich 2D tactile information for precise manipulation. 

\textit{Motion.} 
Humans perceive movement from an egocentric viewpoint through inertial and proprioceptive cues for continuous self-localization and navigation without external references, a capacity that motion sensing seeks to replicate in agents.
\textbf{Inertial Measurement Units (IMUs)} provide continuous estimates of linear acceleration and angular velocity~\cite{huang2018deep}, enabling dead reckoning and short-term trajectory estimation. Analogous to the vestibular system, they combine tri-axial \textit{accelerometers} for translational motion, tilt, and vibration sensing with \textit{gyroscopes} for rotational motion and orientation. Motion can also be derived visually via optical flow, feature tracking, and visual–inertial odometry (VIO), using egocentric cameras to extract self-motion cues relative to landmarks. This complements IMUs by correcting drift and adding spatial context. Low-level motion data can be transformed into structured forms such as skeletal sequences or the \textit{Skinned Multi-Person Linear} (SMPL) parametric body model~\cite{loper2015smpl}, enabling physically grounded tracking.


\textit{Motor Feedback.} Reliable physical motor feedback plays a critical role in spatial intelligence, enabling embodied agents to reason about contact dynamics, manipulate objects in unstructured environments, and adapt their spatial strategies through physical interaction. 
\textbf{Optical force sensors} \cite{park2018optical} detect light transmission or reflection changes in deformable media, captured by cameras or photodiodes, offering resistance for soft robotics and tactile fingertips.
\textbf{Capacitive and piezoelectric sensors} \cite{kemp2007challenges} measure force through capacitance or voltage changes, often embedded in skins or grippers.
\textbf{Magnetic/Hall effect sensors} \cite{kim2016soft} track magnet displacement in elastomeric substrates, providing compact, robust designs with vector readouts for normal and shear forces. \textbf{Proprioceptive sensors}~\cite{shi2020electronic} use flexible conductors, liquid metals, or optical fibers to monitor deformation, curvature, and load in soft robotics and wearables. 
In robotics, motor feedback is fused with vision and tactile inputs using filtering or learning methods to enhance contact estimation, compliant planning, and manipulation.

\begin{takeaway}
    \textcolor{blue(ncs)}{\textbf{Key Insight 1.}} Apart from vision, agents should be equipped with human-inspired sensing modalities, such as auditory and tactile, thereby enabling spatial reasoning across diverse modalities.
\end{takeaway}

\begin{figure}[t]
    \centering
    \vspace{-1em}
    \includegraphics[width=0.6\linewidth]{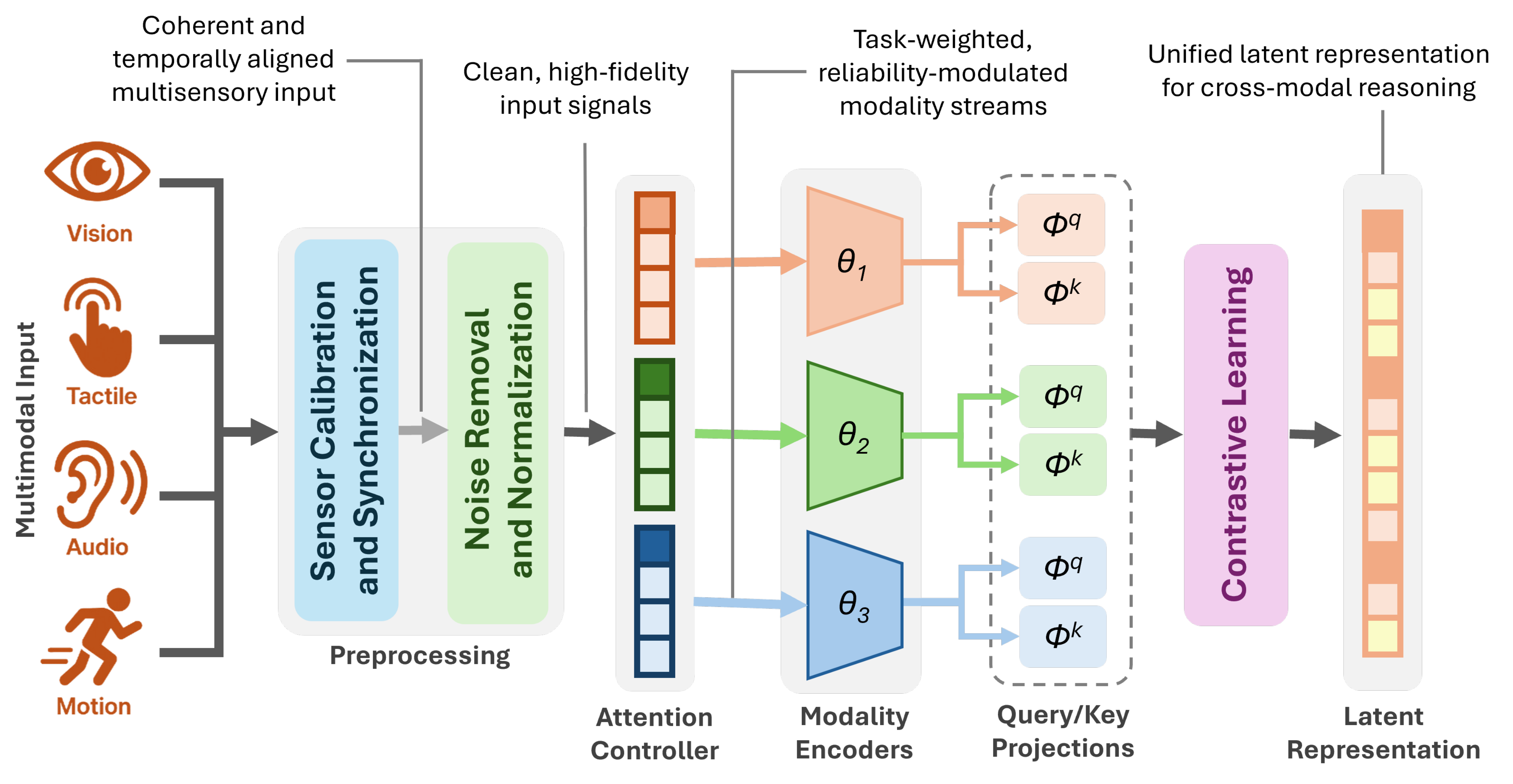}
    \caption{Information Processing Module (IPM) for Spatial Reasoning. Multisensory inputs are preprocessed and attention-weighted before passing through modality-specific encoders. Query/key projections and contrastive learning fuse them into a unified cross-modal latent representation.}
    \label{fig:information_processing_module}
    \Description{ipm}
\end{figure}

\textbf{(b) Information Processing Module.}
\label{sec:2.2.2}
The human brain naturally integrates multi-modal inputs into a unified perceptual experience via specialized sensory organs and convergent neural circuits (Sec.~\ref{sec:2.A.1}). Similarly, the Information Processing Module (IPM) is designed to transduce raw data from heterogeneous sensors into a structured representation for spatial reasoning, as illustrated in Fig.~\ref{fig:information_processing_module}. Inspired by neurobiology, it preconditions raw multimodal signals such as vision, audio, and proprioception through four key computational stages for downstream cognitive operations.

\textit{Sensor Calibration and Synchronization.}  In the brain, primary sensory cortices encode modality-specific signals that are aligned in time and space by multisensory integration regions such as the PPC, and Parietal Operculum~\cite{beauchamp2004integration}. These regions play a critical role in reconciling discrepancies across sensory modalities, enabling coherent perceptual experiences despite variations in timing, spatial origin, or intensity. Analogously, the IPM first performs spatiotemporal calibration across sensors. This includes timestamp synchronization of different cameras with IMUs and LiDAR; alignment of audio streams via interaural difference analysis; and integration of proprioceptive and tactile data through reference frame transformation. This ensures that all incoming signals are temporally coherent and spatially consistent, enabling robust multimodal perception and downstream reasoning.

\textit{Noise Removal and Normalization.} In the early stages of sensory processing, mechanisms like lateral inhibition, adaptation, and spectral filtering enhance signal salience and suppress redundancy \cite{del2025lateral}. IPM replicates these through a modality-specific signal conditioning pipeline. For example, image denoising and histogram equalization for vision; spectral noise reduction for audio; baseline correction and spatial smoothing for tactile skins; and drift correction for IMUs. These operations optimize the fidelity and relevance of each sensory stream, ensuring that subsequent processing stages operate on clean, informative representations.

\begin{figure}[t]
\centering
\vspace{-2em}
\begin{algorithm}[H]
\caption{Low-Level Information Processing}
\label{algo:information-processing}
\begin{algorithmic}[1]
\Require $sensory\_data$
\State $calibrated\_data \gets SensorCalibrationAndSync(sensory\_data)$
\State $clean\_data \gets NoiseRemovalAndNormalization(calibrated\_data)$
\State $attended\_modalities \gets MultisensoryAttentionController(clean\_data)$
\State $unified\_latent\_space \gets CrossModalFeatureIntegration(attended\_modalities)$
\State \Return $unified\_latent\_space$
\end{algorithmic}
\end{algorithm}
\vspace{-1em}
\end{figure}

\textit{Multisensory Attention Controller.} Just as the biological agents modulate sensory gain via attention, like saccadic suppression during gaze shifts or auditory inhibition during vocalization \cite{ khalilian2024corollary}; the IPM includes a multisensory attention controller. Depending on task demands and sensory reliability, it prioritizes the most informative modalities. For example, under low light, LiDAR and radar take precedence; when hands are in contact, tactile-visual attention is invoked; and during human-agent interaction, visual and speech cues may be prioritized to infer user intention. This dynamic attention mechanism ensures computational efficiency and resilience in noisy or occluded environments. Once the most salient modalities are selectively attended, the sensory signals are routed into a cross-modal integration pipeline for deeper representational synthesis.

\textit{Cross-Modal Feature Integration.} In biological systems, multisensory representations emerge in heteromodal areas such as the Superior Colliculus and PPC, where visual, auditory, and somatosensory inputs converge into common egocentric formats \cite{stein1992representation}. The IPM captures this multisensory convergence through foundation models, employing modality-specific backbones, such as vision-language models (VLMs)~\cite{zhang2024vision}, audio-visual transformers \cite{akbari2021vatt}, and tactile-image encoders~\cite{liu2025vitamin}. These modality-specific streams are aligned and integrated to produce a unified multimodal latent space. This latent space constitutes a compact representation, wherein semantically and spatially aligned features from vision, audition, touch, and proprioception are jointly encoded (e.g., 3D semantic segmentation~\cite{lan20242d}). It serves as a rich and continuous perceptual scaffold, enabling context-aware spatial reasoning, predictive modeling, and the subsequent transformation into 3D allocentric reference frames.


\begin{takeaway}
    \textcolor{blue(ncs)}{\textbf{Key Insight 2.}} The IPM transforms multimodal sensory data into a unified representation for spatial reasoning, mimicking the brain’s sensory integration via calibration, noise reduction, dynamic attention, and cross-modal integration. This process is summarized in Algorithm~\ref{algo:information-processing}, which provides a high-level pseudo-code sketch of the IPM's algorithmic flow.
\end{takeaway}


\subsubsection{Egocentric-Allocentric Module}
\label{sec:2.2.3}
\begin{figure}[t]
    \centering
    \vspace{-1em}
    \includegraphics[width=0.65\linewidth]{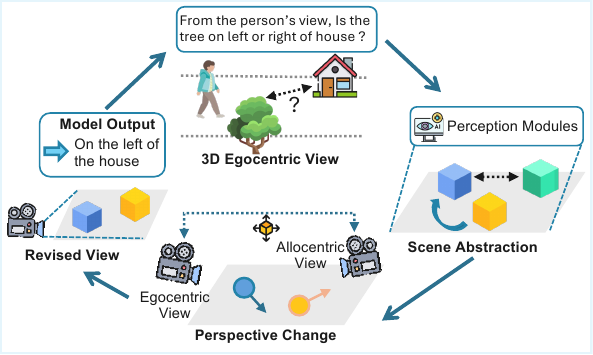}
    \vspace{-1em}
    \caption{Egocentric-to-Allocentric transformation. It commence by projecting previsous cross-modal latent to 3D representation, which are then passed through the scence abstraction and persspective change for allocentric construction.}
    \label{fig:Steps involved in Egocentric-Allocentric Change}
    \Description{egoallo}
\end{figure}

Following low-level sensing (Sec.~\ref{sec:2.A.1}), the human brain converts egocentric views into stable allocentric maps via the hippocampus and entorhinal cortex, with grid and place cells supporting mental simulation. Inspired by this mechanism, our egocentric–allocentric module performs analogous scene abstraction and perspective transformation, producing a persistent, geometry-consistent allocentric 3D map from the unified latent space previously generated by the IPM. As illustrated in Fig.~\ref{fig:Steps involved in Egocentric-Allocentric Change}, building upon the 3D generative modeling, it learns a direct mapping from the abstract latent space of an agent's perception to the parameter space of an explicit 3D scene. 

\textit{3D Egocentric Construction.}
After low-level sensory preprocessing by the IPM, the 3D Egocentric Construction module builds spatially grounded geometric maps from the agent’s viewpoint. These maps are tokenized via emerging 3D layers, such as point cloud transformer~\cite{zhao2021point} and Gaussian splatting feature extractor~\cite{qin2024langsplat}, which convert the low-level latent space into structured 3D tokens~\cite{li2024dreamscene}. Through the constructed 3D maps, it provides a degree-of-freedom (DoF) pose-anchored egocentric space~\cite{liu2025diff9d} that encodes the object extents, distances, orientations, and free-space geometry~\cite{ravi2024outsight}. 

\textit{Scene Abstraction.}
The scene abstraction module derives higher-level, semantic representations from sensory input. It detects and localizes objects in 3D, estimates poses, and infers symbolic spatial relationships (e.g., ``chair left of table'', ``object behind obstacle'')\cite{lan20242d}. It leverages geometric priors from the 3D construction to ensure that semantic labels and spatial relationships are anchored to accurate physical structure while remaining pose-aligned and egocentric~\cite{TII2023, liu2024survey}. The module maintains bidirectional connectivity with the cognitive map module (Sec.~\ref{sec:2.B.4}), serving as the symbolic substrate for contextual inference, memory formation, and mental simulation~\cite{yang2025thinking, yin2025spatialmental}. This loop enables the estimation and refinement of symbolic spatial relationships in light of evolving knowledge, attention, or goals.

\textit{Perspective Shift.} 
Building on egocentric scene abstractions, this component retrieves past 3D snapshots with their poses from working memory and transforms them into a common allocentric frame~\cite{goral2024seeing}. The produced task-specific, short-term allocentric map supports mental reorientation by enabling transformations back into any desired egocentric viewpoint~\cite{lee2025perspective}. When a new viewpoint is needed, it retrieves relevant memories, transforms object positions from each egocentric view into a common allocentric frame, and then reprojects this scene into a new egocentric frame defined by the target viewpoint. This process parallels human spatial reorientation, where the brain builds a viewpoint-independent allocentric model and reprojects it into a new perspective.

\begin{takeaway}
    \textcolor{blue(ncs)}{\textbf{Key Insight 3.}} This module enables agents to build and manipulate spatial representations like the human brain, constructing 3D egocentric maps from sensory data, abstracting semantic relationships, and flexibly shifting perspectives between egocentric and allocentric frames. The sequential logic of this module is demonstrated in Algorithm~\ref{algo:egocentric-allocentric}.
\end{takeaway}

\begin{figure}[t]
\centering
\begin{algorithm}[H]
\caption{Egocentric-Allocentric Conversion}
\label{algo:egocentric-allocentric}
\begin{algorithmic}[1]
\Require $unified\_latent\_space$
\State $WorkingMemory \gets InitializeWorkingMemory()$
\State $egocentric\_3D\_map \gets Construct3DEgocentric(unified\_latent\_space)$
\State $scene\_abstraction \gets SemanticSceneAbstraction(egocentric\_3D\_map)$
\State $allocentric\_map \gets PerspectiveShift(scene\_abstraction, WorkingMemory)$
\State \Return $allocentric\_map$
\end{algorithmic}
\end{algorithm}
\vspace{-1em}
\end{figure}

\subsubsection{Internal Mental Model}
\label{sec:2.B.4}
In humans, the spatial mental model is crucial as it involves both cognitive maps (Sec.~\ref{sec:2.1.2}) and long-term knowledge (Sec.~\ref{sec:2.A.3}) that support navigation and reasoning. To mimic this capability, we introduce two modules: a Cognitive Map Module and a Spatial Neural Memory Module, as outlined in Algorithm~\ref{algo:internal-memory}. The pseudo-code abstracts the algorithmic flow of these modules in forming the internal spatial mental model.

\begin{figure}[t]
    \centering
    \vspace{-1em}
    \includegraphics[width=0.7\linewidth]{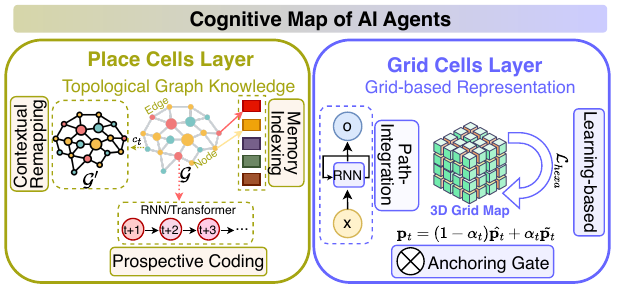}
    \vspace{-1em}
    \caption{Cognitive map for spatial reasoning. This module encodes the agent’s internal representation via a grid-based map and a topological graph.}
    \label{fig:cogniton_module}
    \Description{c}
\end{figure}

\textbf{(a) Cognitive Map Module.}
\label{sec:2.B.4}
In the human brain, the cognitive map is the core mechanism supporting spatial reasoning (Sec.~\ref{sec:2.1.2}). Inspired by this, AI agents internally employ an artificial cognitive map with grid and place cell layers to enable efficient spatial reasoning in complex environments.
As illustrated in Fig.~\ref{fig:cogniton_module}, $\mathcal{G}$ and $\mathcal{G}'$ denote the graph before and after contextual remapping, where the topology remains fixed but node weights change. $\hat{\mathbf{p}}_t$ represents self-motion cues, $\tilde{\mathbf{p}}_t$ external landmarks, and $\mathbf{p}_t$ the anchored position.

\textit{Grid Cells Layer.} Building on neuroscientific findings from Sec.~\ref{sec:2.A.1}, grid cells provide a grid-based metric representation of allocentric space within the cognitive map. Therefore, the grid cells layer equips the agent with an internal metric representation of space, capturing distance and direction through four components (Fig.~\ref{fig:cogniton_module}). \textbf{The Grid-based Representation} leverages a hexagonal lattice pattern, reflecting the periodic firing fields that emerge in grid cells of the medial entorhinal cortex (MEC)~\cite{hafting2005microstructure}. The hexagonal pattern demonstrates an efficient solution for space encoding, in which it can emerge naturally from energy minimization and spatial coverage objectives, making them particularly well-suited for the AI agent~\cite{sorscher2019unified}. To support the emergence of such spatial structure, \textbf{Learning-based Methods} provides an effective mechanism for inducing grid-based representations. By designing the learning frameworks consisting of neuroscience-inspired components, such as predictive loss and hexagonal regularization~\cite{Rylan2023gridneurips}, the agent can generate an internal grid-like map, allowing interaction with other modules for complex spatial reasoning tasks. Moreover, inspired by the role of grid cells in self-motion tracking, \textbf{The Path Integration} component aims to estimates spatial position from self-motion cues (i.e., velocity and direction)~\cite{banino2018vector}. It provides an essential foundation for spatial intelligence, particularly for the embodied AI agent operating in dynamic environments. However, in biological systems, grid-based representations rely not only on self-motion cues but are also anchored to external landmarks, ensuring long-term spatial inference. Therefore, to mimic this mechanism, \textbf{The Anchoring Gate} serves as the modulation interface between other modules and the internal grid-based representation. By integrating reference signals from early perceptual modules, it adjusts the accumulated drift from path integration and improves spatial consistency in reasoning.

\textit{Place Cells Layer.} Complementing grid cells, place cells encode space through sparse, context-dependent activations that form a topological map and support episodic memory via interactions with the entorhinal cortex and neocortex~\cite{keefe1971hippocampus}. The place cells layer in our model mimics the key functions of biological place cells in the HPC with four components, as described in Fig.~\ref{fig:cogniton_module}. \textbf{Topological Representation} is a core element inspired by the relational spatial coding of place cells  within cognitive map~\cite{keefe1971hippocampus},~\cite{o1978hippocampus}. It operates in a graph-based structure, where each node and edge reflect learned relations between the input (i.e., objects and locations), enabling relational spatial inference across complex tasks (i.e., navigation and object-centric)~\cite{battaglia2018relational}. While topological representation captures structural spatial relations, another key property of biological place cells is their ability to adapt representations in response to contextual changes. To reflect this, \textbf{The Contextual Remapping} mechanism enables the place cell layer to shift activation patterns based on task demands and high-level goals. By interacting with other context-aware modules, it allows the agent to maintain multiple internal spatial maps and flexibly switch between them as environmental or task contexts change~\cite{neurips2024place}. When context shifts, the active representation is rapidly transformed, preserving local spatial structure while globally reorganizing the map to reflect the new situation~\cite{neurips2024place}. This mechanism mimics hippocampal global remapping, where place-cell firing patterns reorganize across contexts while maintaining coherent internal geometry~\cite{o1978hippocampus},~\cite{leutgeb2005independent}. 

As described in Sec.~\ref{sec:2.A.3}, beyond the spatial encoding, the biological place cells are believed to function as spatial memory indices in the HPC, supporting the retrieval of episodic context tied to specific locations. Inspired by that, \textbf{The Memory Indexing} component enables the place cells layer to bind spatial representations with grounded experiences, forming a functional gateway aligned with episodic memory~\cite{graves2016hybrid}. Accordingly, it encodes key spatial activations as memory indices, serving as contextual cues for querying episodic traces stored in the downstream module. The map itself remains a lightweight, index-based representation, supporting efficient interaction with memory and reasoning modules~\cite{graves2016hybrid}. This function primarily reflects what is known in neuroscience as retrospective coding, where place cells reactivate representations of previously experienced locations. However, effective spatial reasoning also requires prospective coding, which refers to the ability of biological place cells to predict future spatial states based on contextual signals and intended goals. Therefore, \textbf{The Prospective Coding} component is designed to generate internal representations of potential future spatial states, extending beyond immediate perception. It provides learning objectives that encourage the capture of predictive regularities over time~\cite{oord2018representation}, allowing the agent to anticipate spatial configurations before they are experienced~\cite{gornet2024automated}. This component resembles preplay and replay phenomena observed in the HPC, offering a foundation for spatial inference and flexible navigation in dynamic environments.
\vspace{-0.5em}
\begin{takeaway}
    \textcolor{blue(ncs)}{\textbf{Key Insight 4.}} Cognitive Map Module mimics the human brain by empowering AI to construct an internal representation, using grid-like patterns anchored to metric features and topological networks for contextual memory indexing and relational abstraction.
\end{takeaway}

\textbf{(b) Spatial Neural Memory Module.}
\label{sec:2.2.5}
Building upon the memory-based systems of the human brain, this module is a bio-inspired architecture that endows agents with adaptive and long-term knowledge. It is structured around three interdependent subsystems: spatial–semantic encoding, episodic spatial memory, and adaptive memory updating; each responsible for constructing, retaining, and dynamically refining knowledge representations essential for real-time spatial reasoning and long-term adaptability.

\begin{figure}[t]
\centering
\vspace{-1em}
\begin{algorithm}[H]
\caption{Internal Mental Model}
\label{algo:internal-memory}
\begin{algorithmic}[1]
\Require $unified\_latent\_space, allocentric\_map$
\State $self\_motion \gets$ Current motion cues (IMU, odometry, proprioception)
\State $landmarks \gets$ Detected spatial anchors from $allocentric\_map$
\State $context\_signals \gets$ Task context, semantic cues, prior knowledge
\State $trajectory \gets$ Sequence of recent poses from working memory
\State $events \gets$ Significant interactions or observations

\Statex
\Comment{Cognitive Map}
\State $grid\_cells\_layer \gets GridCellsRepresentation(allocentric\_map, self\_motion, landmarks)$
\State $place\_cells\_layer \gets PlaceCellsRepresentation(grid\_cells\_layer, context\_signals)$
\State $cognitive\_map \gets IntegrateGridAndPlaceCells(grid\_cells\_layer, place\_cells\_layer)$

\Statex
\Comment{Spatial Neural Memory}
\State $semantic\_encoding \gets SpatialSemanticEncoding(unified\_latent\_space)$
\State $episodic\_memory \gets EpisodicMemoryUpdate(trajectory, events)$
\State $adaptive\_memory \gets AdaptiveMemoryUpdating(semantic\_encoding, episodic\_memory)$

\State $mental\_model \gets BuildInternalModel(cognitive\_map, adaptive\_memory)$
\State \Return $mental\_model$
\end{algorithmic}
\end{algorithm}
\vspace{-1.5em}
\end{figure}

\textit{Spatial–Semantic Encoding.} Spatial–Semantic Encoding provides a unified representation of both the geometry and meaning of its surroundings to agents, mirroring the function of long-term memory in humans. This is achieved through grounded visual perception, which aligns vision with semantic cues, and non-visual grounding, which incorporates tactile, auditory, motion, bio-signals, and linguistic cues to enrich or substitute visual understanding. The agent retains this information within a long-term spatial-semantic knowledge base, where experiences accumulate over time into a persistent metric map annotated with object and region-level semantics.

\textbf{Grounded Visual Perception.} 3D-VLMs and 3D-MLLMs provide a powerful approach for building unified 3D visual–semantic representations of an environment. By jointly processing multimodal 3D inputs (e.g., RGB images, depth maps, point clouds) and natural language, these large models can implicitly learn geometric structure, spatial relationships, and object semantics without relying on explicit SLAM or handcrafted mapping pipelines~\cite{ahn2022can}. This capability enables robust grounding of language instructions, such as ``place the box next to the red chair'', within volumetric scene representations~\cite{ahn2022can}. As a result, 3D foundation models support referential resolution, goal localization, serving as key components for spatial reasoning in dynamic real-world environments.

\textbf{Non-Visual Multimodal Grounding.} In the absence of vision, agents build spatial knowledge by encoding non-visual modalities into persistent multimodal embeddings that capture structural and semantic aspects of the environment~\cite{chowdhury2024meerkat}. Through this process, linguistic inputs supply spatial priors and semantic grounding while aligning with tactile, auditory, proprioceptive, and motion signals to support coherent spatial reasoning. Tactile signals encode material properties (e.g., texture, hardness, friction) into reusable object descriptors that inform manipulation strategies beyond the current scene. Auditory cues (e.g., room-scale echoes, appliance hums) can be embedded as spatial–acoustic signatures for recognizing functional zones across environments~\cite{chowdhury2024meerkat}. Proprioceptive and force feedback form embodied state memories, supporting safe and dexterous interaction patterns over time. Even bio-signals (e.g., fMRI and EEG) can be stored as neural-intent embeddings, linking latent cognitive states to likely action templates~\cite{chau2025population}. Motion trajectories, captured from inertial and kinematic sensors, can be preserved as spatio-temporal motion primitives~\cite{feng2024chatpose}, enabling agents to recall and adapt locomotion or manipulation patterns across different environments. By integrating linguistically aligned non-visual embeddings into a long-term spatial–semantic memory, foundational models can reason effectively under partial perception, leveraging accumulated multimodal experience.

\begin{figure}[t]
    \centering
    \vspace{-1em}
    \includegraphics[width=0.6\linewidth]{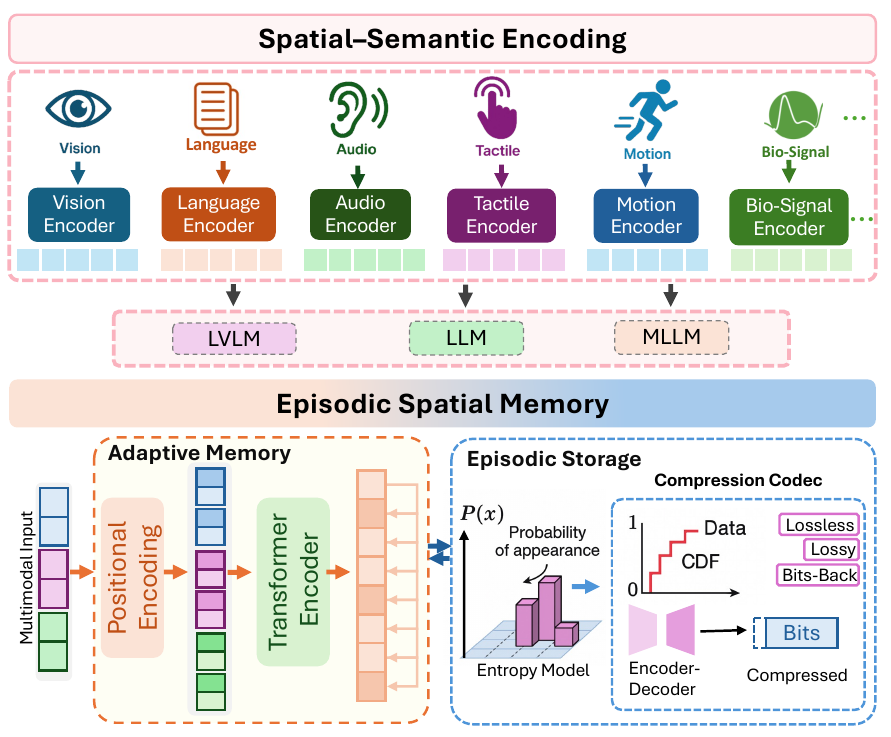}
    \caption{(A) \textbf{Spatial–Semantic Encoding.} Sensory inputs are transformed into spatial and semantic representations for unified understanding across modalities. (B) \textbf{Episodic Spatial Memory.} Multimodal inputs are encoded by a transformer, compressed via entropy modeling for efficient spatial–temporal memory.}
    \label{fig:spatial_neural_memory}
\end{figure}

\textit{Episodic Spatial Memory.} In biological systems, episodic memory, supported by the HPC-MEC, encodes spatial and temporal context and enables trajectory reasoning, causal inference, and foresight. Translating this to artificial agents, an auto-regressive transformer provides a natural computational framework for modeling sequential recall and predicting future states by unfolding stored experiences over time~\cite{cho2024spatially}. To operate in dynamic environments, agents must not only update episodes continuously for short-term reasoning, but also offload older episodes into a cold-storage layer. Hence, this episodic store can be efficiently maintained through compression, where entropy models and codec compressor (e.g., ANS) and advanced schemes, such as bits-back coding, can minimize storage cost while preserving the semantic and structural fidelity of experiences~\cite{kingma2021variational, hieu2024point}. Such a design unifies episodic recall, spatial representation, and memory compression, enabling agents to navigate, localize, and plan with both rich context and scalable memory capacity.

\textit{Adaptive Memory Updating.} 
Effective spatial reasoning in dynamic environments relies on balancing stability and plasticity, where mechanisms like systems consolidation integrate short-term episodic memories into long-term knowledge for continuous refinement.
Inspired by this, AI agents should maintain a fast-learning episodic memory that captures recent events in detail and selectively transfers relevant information into a stable long-term spatial-semantic memory. This integration allows agents to generalize from specific experiences, update their understanding of the world, and adapt to changes without erasing useful prior knowledge. Continual learning methods such as parameter isolation, synaptic consolidation, and memory replay enable incremental learning while preventing catastrophic forgetting~\cite{shin2017continual}. Online learning further supports rapid updates during real-time interactions, which is especially useful for navigation and instruction following~\cite{shin2017continual}. In settings where labeled data is limited, self-supervised learning allows agents to extract meaningful structure directly from sensory input. Adaptive memory updating acts as the bridge between experience and generalization, enabling embodied agents to learn continually and operate flexibly across changing environments.
\begin{takeaway}
    \textcolor{blue(ncs)}{\textbf{Key Insight 5.}} The Spatial Neural Memory Module emulates the human mental model by integrating spatial-semantic encoding, episodic memory, and adaptive updating, forming a dynamic internal model alongside the cognitive map for robust spatial reasoning and adaptability.
\end{takeaway}

\subsubsection{Reasoning Module} 
When humans form a cognitive map, they use it actively for complex spatial reasoning, such as imagining alternate viewpoints and manipulating spatial relationships to achieve goals. This transcends simple ``stimulus–response'' behavior, reflecting advanced intelligence. The reasoning module emulates this capability, serving as the executive hub of our framework. It transforms structured spatial knowledge from the cognitive map (Sec.~\ref{sec:2.B.4}) and spatial neural memory (Sec.~\ref{sec:2.2.5}) into task-oriented and predictive logical thinking sequences. 

\textit{Predictive World Modelling.} The core of this approach lies in using the internal cognitive map to infer the external world, forming a predictive world model. This is not just a static snapshot of the environment, but a dynamic, predictive internal simulator. In our framework, the cognitive map (Sec.~\ref{sec:2.A.4}) and spatial neural memory (Sec.~\ref{sec:2.2.5}) form an internal mental model of spatial knowledge, which is naturally enriched with intuitive physical and semantic constraints~\cite{maocom}, serving as the foundation of this world model. By leveraging generative (e.g., diffusion and variational autoencoders) and joint-predictive architecture~\cite{lecun2022path}, the agent can perform predictive inference on this mental model. For example, the model can predict what a new viewpoint will look like after moving within the environment or potential future changes in a certain area. This approach reflects the neuroscientific view that the brain continually predicts upcoming sensory input and refines its world model by comparing predictions with reality~\cite{xing2025critiques}. 

\begin{figure}[t]
    \centering
    \vspace{-1em}
    \includegraphics[width=0.7\linewidth]{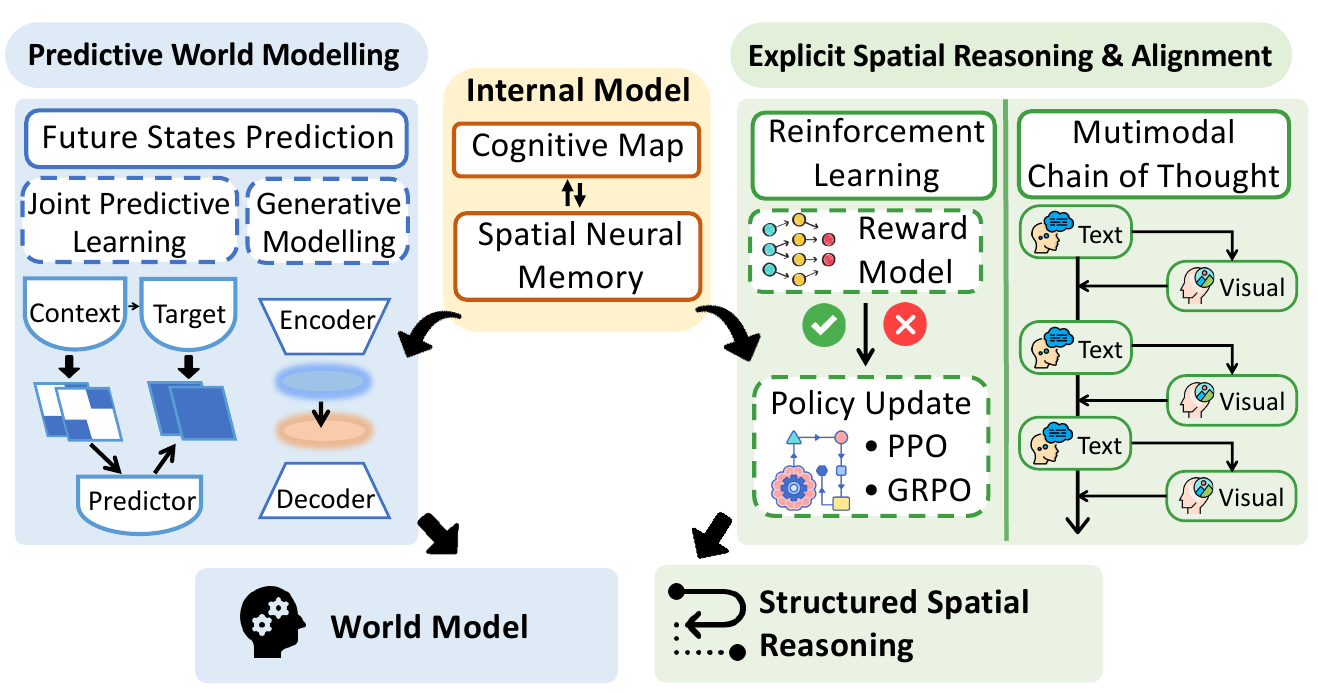}
    \vspace{-1em}
    \caption{Reasoning Module for Spatial Reasoning. Predictive world modelling refers to the predictive ability of internal model while explicit spatial reasoning using alignment methods to perform structured inference.}
    \label{fig:Reasoning module}
    \Description{reasoning}
\end{figure}

\textit{Explicit Spatial Reasoning and Alignment.} Once possessing the predictive world model, the key question becomes: how should an Artificial Spatial Intelligence agent use it to perform correct and complex multi-step spatial reasoning? Such methods aim to enable the agent to explicitly construct and apply a structured reasoning process, directly extracting structured knowledge from perceptual information. Our Reasoning Module addresses this by enabling multi-step spatial thinking. It decomposes the reasoning process, aligns it with human spatial preferences, and trains the agent to learn optimal strategies through exploration in a world-model simulation. Specifically, this can be achieved by training the model for step-by-step spatial reasoning using visual-aware Chain-of-Thought (CoT)\cite{li2025imagine} and reinforcement learning (RL)\cite{ouyang2022training}, allowing it to reason through a learned paradigm before acting. Alternatively, the model can construct intermediate spatial artifacts (e.g., auxiliary lines, visual aids) from multimodal perception and use them as the basis for inference. 
\begin{takeaway}
    \textcolor{blue(ncs)}{\textbf{Key Insight 6.}} Reasoning Module enables agents to perform complex spatial reasoning by transforming the internal model into dynamic, predictive world models, using structured multi-step reasoning and alignment with human spatial preferences to infer in complex environments. Algorithm~\ref{algo:reasoning} outlines the high-level logic of the reasoning process.
\end{takeaway}

In summary, the proposed framework integrates multimodal sensing, egocentric-allocentric conversion, mental modeling, and reasoning into a coherent pipeline for agentic spatial intelligence. Thus, Algorithm~\ref{algo:asi} summarizes the integrated process, providing a high-level pseudo-code sketch of the complete perception–cognition–action loop.

\subsubsection{Spatial Reasoning Behaviours} 
\label{sec: 2.B.7}
In the human brain, reasoning processes are distributed across specialized neural system with each tuned to particular cognitive demands. Enabling this into agents with spatial intelligence requires categorizing spatial reasoning tasks and decomposing behaviors in ways that align with corresponding modules~\cite{zinan2025nature}. As described in Sec.~\ref{sec:2.A.4}, neuroscience computational models formalize the brain’s decomposition of cognitive functions, serving as a foundation for systematically categorizing spatial reasoning behaviors in AI agents. Building on this, we adopt the \textbf{Hierarchical Active Inference} (HAI) framework, derived from foundational neuroscience models such as \textit{Bayesian Inference} and the \textit{Free Energy Principle}, and widely regarded as a minimal model for studying spatial reasoning. HAI models cognition as a layered process of perception, inference, and action, using predictive feedback across levels to manage spatial reasoning at varying levels of abstraction, thereby providing a biologically inspired framework for analyzing and decomposing such behaviors in AI agents. Following the HAI layered theory, we categorize spatial reasoning behaviours into three groups: 3D Perceptual Inference, Hidden-State Inference, and Policy Selection. This reflects the functional tiers emphasized in the HAI framework and aligns with the hierarchical nature of spatial reasoning, making it well-suited for bio-inspired AI spatial intelligence.

\textit{3D Perceptual Inference.}
In HAI, perceptual inference forms the fundamental layer of the cognitive hierarchy, processing raw sensory data to make low-level predictions about the environment. Extending this, 3D perceptual inference grounds low-level spatial intelligence by reasoning about 3D structure from multi-modal inputs (e.g., vision, tactile, auditory). Continuous processing of these modalities allows the agent to infer depth, shape, and spatial boundaries directly from raw input, without relying on higher-order cues, enabling real-time responses to dynamic environments. For instance, recent AI systems have widely explored tasks such as monocular depth estimation, 3D motion pose estimation, 3D semantic segmentation, which exemplify the low-level 3D spatial understanding from raw sensory data of AI agents. Therefore, the notion of 3D perceptual inference behaviors enables a structured foundation for anchoring higher-level spatial reasoning, ensuring that complex cognitive functions are grounded in accurate and adaptive sensory understanding.

\textit{Hidden-State Inference.} 
After raw sensory prediction, HAI advances to latent belief inference, where the brain infers hidden causes of input to build coherent, temporally extended internal representations.
Therefore, the hidden-state inference behaviors are formed to generate and manipulate internal spatial models beyond the observed scene. Operating at a higher cognitive level, this spatial reasoning behavior supports the simulation of unseen dynamics and structural patterns abstraction within the environment. Hence, we divide the hidden-state inference into three major groups: Structural Reasoning, Mental Simulation, and Abstract Reasoning, each showing complex forms of internal representation and generalization for spatial intelligence. 

\begin{figure}[t]
\centering
\vspace{-1.5em}
\begin{algorithm}[H]
\caption{Reasoning Module}
\label{algo:reasoning}
\begin{algorithmic}[1]
\Require $mental\_model$
\State $predictive\_world\_model \gets PredictFutureStates(mental\_model)$
\State $structured\_reasoning \gets ExplicitSpatialReasoning(predictive\_world\_model)$
\State $action\_policy \gets PolicyLearning(structured\_reasoning)$
\State \Return $action\_policy$
\end{algorithmic}
\end{algorithm}
\vspace{-1em}
\end{figure}

\textbf{Structural Reasoning} is the ability to infer and manipulate the spatial structure of an environment beyond immediate perception, supporting internal models that capture both physical geometry and spatial relations. 
\textit{Geometric Reasoning} infers metric properties such as distances, angles, orientations, object geometries, and relative scales, enabling accurate representations for tasks requiring fine-grained precision.
\textit{Topological Reasoning} captures non-metric relations, such as connectivity, containment, adjacency, and continuity, abstracting from exact measurements to model invariant structural relationships for generalization across environments.
\textit{Spatiotemporal Reasoning} integrates spatial structure with temporal dynamics to model how entities move, transform, or interact over time, supporting prediction, continuity tracking, and event sequence understanding.
These reasoning types form the structural backbone for hidden-state spatial cognition, enabling agents to maintain a coherent spatial understanding under uncertainty space.

\textbf{Mental Simulation} builds upon structural representations by allowing the agent to internally manipulate spatial models over time, aligning with the belief latent of the HAI framework. It enables the simulation of predictive spatial scenarios and anticipate the hidden spatial aspects of the environment. To enable that, \textit{perspective taking} is a essential spatial task which involves mentally shifting the individual viewpoint to understand how the environment appears from different positions, thereby enhancing spatial awareness. Meanwhile, \textit{visuospatial reasoning} is also another behavior that supports the mental manipulation of objects or scenes, such as rotating, resizing, or repositioning them to support spatial reasoning, with mental rotation being a well-studied example in neuroscience. These mental mechanisms jointly enable more flexible spatial reasoning beyond direct perception.  

\textbf{Abstract Reasoning} enables an agent to infer hidden rules, relational patterns, or symbolic structures not directly observable from perception. As a branch of hidden-state inference, it uncovers latent causes and abstract regularities underlying spatial or task-specific environments. Among abstract reasoning, symbolic inference enables the spatial reasoning over abstract spatial rules and relational logic beyond direct perception. Following that, lingual-based spatial reasoning uses linguistic input to guide spatial inference and decision-making in the non-perceptual environment. Additionally, schema-based inference supports generalization by applying prior abstract spatial templates to interpret new or incomplete situations. These components enable agents to reason over abstract representations, facilitating analogy, and problem-solving in complex environments.

\textit{Policy Selection.} forms the final stage of spatial reasoning in the HAI framework, where the agent transforms internal beliefs into concrete action sequences. These behaviours coordinate information from perceptual representations and inferred hidden structure, allowing the agent to reason and select actions that align with spatial goals.
Taking goal-directed navigation as an example, the agent constructs a local spatial map from sensory input, integrates inferred topological or schematic knowledge to evaluate possible routes, and selects the most efficient path toward the target. Abstract reasoning components, such as symbolic constraints or language-guided goals, may also modulate the chosen action policy (e.g., ``avoid the red zone''). By unifying information across layers, policy selection behaviours allow agents to act coherently within dynamic environments.

\begin{figure}[t]
\centering
\vspace{-1.5em}
\begin{algorithm}[H]
\caption{Agentic Spatial Intelligence Framework}
\label{algo:asi}
\begin{algorithmic}[1]
\Require $Environment$ \Comment{Raw multimodal sensory stream}

\While{true} \Comment{Continuous interaction loop}

    \State $sensory\_data \gets Environment.CollectSensors\big($ \\
    \hspace*{4em}vision: $getData(Camera, LiDAR, Depth, EventCam)$ \\
    \hspace*{4em}audio: $getData(Microphones, Ultrasonic)$ \\
    \hspace*{4em}tactile: $getData(TactileSkin, WhiskerSensors)$ \\
    \hspace*{4em}motion: $getData(IMU, VIO)$ \\
    \hspace*{4em}motor\_feedback: $getData(ForceSensors, Piezoelectric)\big)$

    \vspace{0.5em}
    \State $unified\_latent\_space \gets InformationProcessing(sensory\_data)$
    \State $allocentric\_map \gets EgocentricAllocentricConversion(unified\_latent\_space)$
    \State $mental\_model \gets InternalMemoryModel(unified\_latent\_space, allocentric\_map)$
    \State $action\_policy \gets Reasoning(mental\_model)$
    \State $action \gets ExecuteAction(action\_policy, actuators)$
    \State $feedback \gets ReceiveMotorAndEnvironmentalFeedback()$
    \State $Environment \gets Update(Environment, feedback)$

\EndWhile

\end{algorithmic}
\end{algorithm}
\vspace{-1em}
\end{figure}

\subsubsection{Implementation of Agentic Spatial Intelligence}
This section explores practical deployment pathways for agentic spatial reasoning in real-world settings, focusing on two core components: computing and communications. Building on this, we examine a progression of agents deployment paradigms, including edge AI, neuromorphic computing, and intelligent communication. Table~\ref{table:deploy} provides an overview of deployment tools.

\textit{Edge AI.}
Edge computing refers to the paradigm of processing data locally at or near the source of generation, rather than relying on centralized cloud infrastructure~\cite{lim2020federated}.
This approach enables low-latency and privacy-preserving AI implementation, making it particularly well-suited for on-device applications in mobile, wearable, and robotic systems. 

\textbf{Edge AI for Agents}. To enable on-device spatial reasoning, Edge AI relies on compact, efficient models optimized for limited compute environments. Hence, AI model compression techniques (e.g., model pruning, quantization)~\cite{cheng2024survey}, and knowledge distillation~\cite{wang2021knowledge} are employed to convert large models into Tiny AI models that maintain accurate performance while reducing memory and energy demands. 
Building upon this, recent advancements in small-scale foundation models, such as small vision-language, have enabled multimodal fusion directly on edge devices~\cite{zhu2024minigpt}, supporting spatial reasoning without relying on cloud computing. In parallel, lightweight 3D spatial representations have also emerged to support real-time processing on resource-constrained devices, employing hierarchical encoding and blob-level to enable compact scene representations with reduced memory and computational costs~\cite{zhang2024lp}.

\textbf{Hardwares/Softwares.} The deployment of agentic spatial reasoning in edge computing can be potentially supported by a diverse range of both embedded hardware and software platforms. 
Regarding hardware implementation, emerging systems such as NVIDIA Jetson Orin Nano~\cite{nvidia_jetson}, Google Coral Edge TPU~\cite{coral_tpu}, and Intel Movidius Myriad X~\cite{intel_myriad} offer a balance between low power consumption and high-performance AI inference, making them well-suited for agentic spatial reasoning.
These platforms integrate AI accelerators within system-on-chip (SoC) architectures, enabling close coupling with multimodal sensors and supporting on-device execution of lightweight 3D spatial models such as depth-based SLAM, point cloud projection, and 3D Gaussian splatting.

On the software side, a variety of optimized inference frameworks facilitate the deployment of spatial models under edge constraints.
Specifically, open-sources such as TensorRT~\cite{tensorrt}, ONNX Runtime~\cite{jin2020compiling}, Tengine~\cite{tengine}, PyTorch Mobile~\cite{zhang2022comprehensive}, and TensorFlow Lite~\cite{david2021tensorflow} provide model quantization, operator fusion, and hardware-specific acceleration, enabling efficient execution on a range of embedded processors.
These runtimes support real-time perception and reasoning pipelines, and are often integrated with lightweight system schedulers or ROS-based middleware to manage multi-sensor fusion and decision-making in spatial reasoning.

\textit{Neuromorphic Computing.}
Beyond conventional edge accelerators, neuromorphic computing offers a biologically inspired approach to spatial intelligence, emphasizing event-driven processing and low power consumption~\cite{liu2025neural}.

\textbf{Brain-inspired Computing}. Key method of this paradigm is Spiking Neural Network (SNN), which models the brain’s temporal coding mechanisms by transmitting discrete spikes across time. Specifically, SNN emulates the event-driven and temporally sparse communication patterns of biological neurons, processing information through discrete spikes rather than continuous activations~\cite{yao2023attention}.
This asynchronous and low-frequency operation enables significantly reduced energy consumption compared to conventional neural networks, particularly in real-time settings.
Due to the efficient use of compute and memory, SNN is well-suited for deploying agentic spatial reasoning on edge hardware where power and latency are critical constraints~\cite{shang2025energy}.

\begin{table}[t]
\vspace{-1em}
\caption{Deployment Paradigms for Agentic Spatial Intelligence.}
\vspace{-1em}
\label{table:deploy}
\resizebox{\textwidth}{!}{%
\begin{tabular}{l|ll|l}
\hline
Paradigms & \multicolumn{2}{l|}{Deployment Tools} & Key Feautres \\ \hline
\multirow{2}{*}{Edge AI}                & \multicolumn{1}{l|}{Hardwares}        & \begin{tabular}[c]{@{}l@{}}NVIDIA Jetson Series~\cite{nvidia_jetson}, Google Coral Edge~\cite{coral_tpu},\\ Intel Movidius Myriad~\cite{intel_myriad}\end{tabular} & \multirow{2}{*}{\begin{tabular}[c]{@{}l@{}} Low-latency multimodal fusion, model\\ compression for on-device reasoning\end{tabular}} \\ \cline{2-3} & \multicolumn{1}{l|}{Softwares}        & \begin{tabular}[c]{@{}l@{}}TensorRT~\cite{tensorrt}, ONNX Runtime~\cite{jin2020compiling}, Tengine~\cite{tengine}, \\ PyTorch Mobile~\cite{zhang2022comprehensive}, TensorFlow Lite~\cite{david2021tensorflow}\end{tabular} &  \\ 
\hline
\multirow{2}{*}{\makecell[l]{Neuromorphic\\Computing}} & \multicolumn{1}{l|}{Hardwares}        & Intel's Loihi series~\cite{davies2018loihi}, BrainChip's Akida~\cite{posey2022akida}    & \multirow{2}{*}{\begin{tabular}[c]{@{}l@{}} Event-driven, low-power SNN for temporal\\ coding in dynamic environments\end{tabular}}            \\ \cline{2-3} 
& \multicolumn{1}{l|}{Softwares}        & \begin{tabular}[c]{@{}l@{}} Brian2~\cite{stimberg2019brian}, BindsNET~\cite{hazan2018bindsnet}, SpiNNaker~\cite{furber2020spinnaker},  Norse~\cite{pehle2021norse}\end{tabular}                 &            \\ \hline
\makecell[l]{Intelligent\\Communication} & \multicolumn{1}{l|}{\makecell[l]{Simulation\\Tools}} & \begin{tabular}[c]{@{}l@{}}Aerial Omniverse Digital Twin~\cite{pegurri2025towards}, PHY-Sionna~\cite{aoudia2025sionna},\\ Aerial Commercial Testbed~\cite{nvidia_oran}, AI-RAN Sionna Kit~\cite{cammerer2025sionna}\end{tabular} & \begin{tabular}[c]{@{}l@{}} AI-native 5G/6G for low-latency,\\ adaptive multi-agent coordination\end{tabular}           \\ \hline
\end{tabular}
}
\end{table}

\textbf{Hardwares/Softwares.} Neuromorphic hardwares have been widely developed to support the efficient execution of SNN and brain-inspired models.
Notable examples include Intel's Loihi series, which features on-chip plasticity, asynchronous spike routing, and hierarchical memory structures that mimic cortical processing~\cite{davies2018loihi}; and BrainChip's Akida, which is optimized for ultra low power inference and event-based sensor integration at the edge~\cite{posey2022akida}.
By aligning hardware-level operations with the sparse, temporal nature of SNN, they allow agents to process multimodal inputs more efficiently, particularly under energy and timing constraints.

To support the deployment of the neuromorphic hardwares, various open-source SNN frameworks have been developed for both simulation and hardware execution~\cite{liu2025neural}. Simulators such as NEST~\cite{dupuy1990nest}, Brian2~\cite{stimberg2019brian}, and BindsNET~\cite{hazan2018bindsnet} offer intuitive environments for building and testing SNN architectures, with support for large-scale neural network integration. Moreover, for hardware execution development, SpiNNaker~\cite{furber2020spinnaker} enables distributed SNN execution across custom neuromorphic systems, while Norse~\cite{pehle2021norse} provides a PyTorch-based library optimized for GPU-accelerated spiking computation. Together, these tools form a critical software foundation for prototyping and deploying brain-inspired spatial reasoning agents across diverse platforms.

\textit{Intelligence Communication.} 
In addition to computing architectures, communication framework is also critical for deploying agents in dynamic environments.
In particular, advanced 5G/6G systems and AI-native communication stacks are expected to provide the low-latency, high-reliability links necessary for distributed spatial intelligence.

\textbf{5G Beyond and 6G Wireless Systems}. 
Next-generation wireless systems, such as 5G and the emerging 6G paradigm, are poised to serve as critical infrastructure for the deployment of agents' spatial reasoning in dynamic, real-world environments.
They enable ultra-reliable, low-latency transmission of spatial information, which is particularly important for embodied agents operating in motion-rich or multi-agent scenarios~\cite{zhang2025toward}.
Beyond conventional connectivity, 6G is expected to embrace AI-native wireless architectures~\cite{nvidia_oran}, including programmable networks (e.g., open-radio access networks~\cite{nvidia_oran}, rate multiple splitting access~\cite{hieu2023joint}, and integrated sensing and communication frameworks~\cite{chu2023ai}), where learning-based agents can dynamically adapt communication strategies to environmental and task-specific demands.

\textbf{Emerging Softwares.} Obtaining the essential requirements from wireless systems in the agentic era, NVIDIA has released a comprehensive Aerial Research portfolio to support the development of AI-native 6G infrastructures.
Specifically, it contains Sionna, an open-source GPU-accelerated library for simulating physical-layer communication~\cite{aoudia2025sionna}; the Aerial Omniverse Digital Twin for full-stack 6G simulation~\cite{pegurri2025towards}; and the Sionna Research Kit on Jetson for prototyping AI-RAN algorithms on embedded platforms~\cite{cammerer2025sionna}.
These platforms enable scalable and low-latency communication, which aligned with the needs of AI agents operating under mobility, partial observability, and real-time constraints.



\section{Comprehensive Analysis of Agentic Spatial Intelligence}
\label{sec:3}

\subsection{Insights From The Proposed Framework}
\label{sec:3.A}
In this section, we conduct an in-depth analysis of state-of-the-art (SOTA) research corresponding to the modules within the proposed framework. Hence, we aim to outline potential methods tailored for each module and expose critical gaps that need to be addressed for agentic systems to meet human-level spatial intelligence.

\subsubsection{Low-Level Multimodal Sensory Processing}
\label{sec:3.A.1}

\begin{table}[]
\vspace{-1em}
\caption{Representative works in low-level multimodal sensory processing for embodied agents.}
\vspace{-0.5em}
\label{table:3.A.1}
\resizebox{\textwidth}{!}{%
\begin{tabular}{>{\centering\arraybackslash}p{2.8cm}lllll}
\hline
\textbf{Category} &
  \textbf{Method} &
  \textbf{Modality} &
  \textbf{Publication} &
  \textbf{Highlights} &
  \textbf{Limitations} \\ \hline
\multirow{3}{*}{\raisebox{-0.6cm}{\rotatebox{90}{\parbox{2.2cm}{\centering \textbf{Sensor Calibration and Synchronization}}}}} &
  Li et al.~\cite{li2025accurate} &
  \begin{tabular}[c]{@{}l@{}}LiDAR, IMU, \\ Camera\end{tabular} &
  Inf. Fusion'2025 &
  \begin{tabular}[c]{@{}l@{}}Target-free continuous-time optimization \\ for multimodal calibration\end{tabular} &
  \begin{tabular}[c]{@{}l@{}}Needs precise IMU calibration; \\ high compute cost\end{tabular} \\ \cline{2-6} 
 &
  Wang et al.~\cite{wang2025observability} &
  LiDAR, IMU &
  Preprint'2025 &
  \begin{tabular}[c]{@{}l@{}}Active trajectory optimization using \\ Fisher Information\end{tabular} &
  \begin{tabular}[c]{@{}l@{}}Sensitive to sensor quality \\ and scene richness\end{tabular} \\ \cline{2-6} 
 &
  Herau et al.~\cite{herau2024soac} &
  RGB, Depth &
  CVPR'2024 &
  \begin{tabular}[c]{@{}l@{}}NeRF-based self-supervised cross-\\ sensor calibration\end{tabular} &
  \begin{tabular}[c]{@{}l@{}}Fails in linear motion; scales \\ poorly with many sensors\end{tabular} \\ \hline
\multirow{4}{*}{\raisebox{-0.6cm}{\rotatebox{90}{\parbox{3.0cm}{\centering \textbf{Noise Removal and Normalization}}}}} &
  QFormer~\cite{jiang2024qformer} &
  RGB &
  IJCAI'2024 &
  \begin{tabular}[c]{@{}l@{}}Quaternion transformer for efficient \\ image denoising\end{tabular} &
  \begin{tabular}[c]{@{}l@{}}Limiting use beyond image \\ denoising\end{tabular} \\ \cline{2-6} 
 &
  Fang et al.~\cite{fang2023partially} &
  Audio &
  ICASSP'2023 &
  \begin{tabular}[c]{@{}l@{}}Joint ego/environmental audio denoising \\ via VAE+NMF\end{tabular} &
  \begin{tabular}[c]{@{}l@{}}Requires robust ego-noise \\ pretraining; high complexity\end{tabular} \\ \cline{2-6} 
 &
  Qian et al.~\cite{qian2025sav} &
  Audio-Visual &
  JSTSP'2025 &
  \begin{tabular}[c]{@{}l@{}}Context-aware audio-visual speech \\ enhancement\end{tabular} &
  \begin{tabular}[c]{@{}l@{}}Only supports single speaker; \\ no separation\end{tabular} \\ \cline{2-6} 
 &
  RegBN~\cite{ghahremani2023regbn} &
  Multimodal &
  ANIPS'2023 &
  \begin{tabular}[c]{@{}l@{}}Batch normalization across modalities\\  with regularization\end{tabular} &
  \begin{tabular}[c]{@{}l@{}}Slow due to costly SVD and \\ L-BFGS steps.\end{tabular} \\ \hline
\multirow{5}{*}{\raisebox{-0.6cm}{\rotatebox{90}{\parbox{3.8cm}{\centering \textbf{Multisensory Attention Control}}}}} &
  AdapTac~\cite{li2025adaptive} &
  \begin{tabular}[c]{@{}l@{}}Vision, Tactile, \\ Force\end{tabular} &
  Preprint'2025 &
  \begin{tabular}[c]{@{}l@{}}Force-guided visual-tactile attention \\ for manipulation\end{tabular} &
  \begin{tabular}[c]{@{}l@{}}Lacks robustness without \\ RL fine-tuning\end{tabular} \\ \cline{2-6} 
 &
  FAM-HRI~\cite{lai2025fam} &
  \begin{tabular}[c]{@{}l@{}}Vision, Audio, \\ Language\end{tabular} &
  Preprint'2025 &
  \begin{tabular}[c]{@{}l@{}}Gaze-speech fusion for intent-driven \\ control via LLMs\end{tabular} &
  \begin{tabular}[c]{@{}l@{}}LLM/VLM inference delays \\ processing\end{tabular} \\ \cline{2-6} 
 &
  Eftekhar et al.~\cite{eftekharselective} &
  RGB &
  ICLR'2024 &
  \begin{tabular}[c]{@{}l@{}}Task-conditioned codebook for visual \\ filtering\end{tabular} &
  \begin{tabular}[c]{@{}l@{}}Codebook collapse; loss of \\ fine details\end{tabular} \\ \cline{2-6} 
 &
  UniTouch~\cite{yang2024binding} &
  \begin{tabular}[c]{@{}l@{}}Tactile, Audio, \\ Vision\end{tabular} &
  CVPR'2024 &
  \begin{tabular}[c]{@{}l@{}}Touch embeddings aligned with \\ audio-visual features\end{tabular} &
  \begin{tabular}[c]{@{}l@{}}Lacks for task-driven attention \\ modulation\end{tabular} \\ \cline{2-6} 
 &
  Gizdov et al.~\cite{gizdov2025seeing} &
  RGB &
  CVPR'2025 &
  \begin{tabular}[c]{@{}l@{}}Foveated sampling for efficient visual \\ grounding\end{tabular} &
  \begin{tabular}[c]{@{}l@{}}No dynamic gaze control; \\ fixed attention\end{tabular} \\ \hline
\multirow{4}{*}{\raisebox{-0.6cm}{\rotatebox{90}{\parbox{3.0cm}{\centering \textbf{Cross-Modal Feature Integration}}}}} &
  AVFormer~\cite{seo2023avformer} &
  Audio-Visual &
  CVPR'2023 &
  \begin{tabular}[c]{@{}l@{}}Visual adapters for audio-visual speech \\ recognition\end{tabular} &
  \begin{tabular}[c]{@{}l@{}}Not end-to-end; pretrained \\ modules required\end{tabular} \\ \cline{2-6} 
 &
  Gan et al.~\cite{gan2020looklistentact} &
  Vision, Sound &
  ICRA'2020 &
  \begin{tabular}[c]{@{}l@{}}Dual-branch model for vision-sound \\ navigation\end{tabular} &
  No spatial memory or mapping \\ \cline{2-6} 
 &
  MViTac~\cite{dave2024multimodal} &
  Vision, Tactile &
  ICRA'2024 &
  Self-supervised visual-tactile learning &
  \begin{tabular}[c]{@{}l@{}}Weak grasp prediction due to \\ sparse data\end{tabular} \\ \cline{2-6} 
 &
  Polyglot~\cite{komarichev2022polyglot} &
  \begin{tabular}[c]{@{}l@{}}Image, Audio, \\ Text\end{tabular} &
  CAGD'2022 &
  Aligns image, audio, text in shared space &
  \begin{tabular}[c]{@{}l@{}}Embeddings lack spatial/\\ topological grounding\end{tabular} \\ \hline
\end{tabular}%
}
\end{table}

This part examines recent research works on low-level perception of the agents, grounding upon the multi-sensory input (Sec.~\ref{sec:2.B.1}) and information processing (IPM) modules (Sec.~\ref{sec:2.2.2}) within our proposed framework. Specifically, we analyze the sensory processing pipeline of the IPM and examine the types of sensory modalities and sensors adopted in these works.

\textit{Sensor Calibration and Synchronization.}
Modern intelligent systems increasingly depend on multimodal sensor fusion to achieve greater levels of autonomy. Thus, precise and consistent spatiotemporal calibration across heterogeneous sensors is crucial. 
Hence, Li \textit{et al.}\cite{li2025accurate} proposed a fully automatic, target-free framework based on continuous-time optimization for the spatiotemporal calibration of multi-modal sensors, including IMU, LiDAR, camera, and Radar. By employing an IMU-centric initialization and joint batch optimization strategy, the method eliminates the need for overlapping fields of view and manual intervention. However, \textbf{the method remains limited by its reliance on precise IMU calibration and the heavy computational cost of high-dimensional factor graph optimization, which hinders real-time deployment with low-grade IMUs.} Wang \textit{et al.}\cite{wang2025observability} proposed an observability-aware active calibration framework for ground robots with multimodal sensors (e.g., microphone arrays, LiDAR, wheel odometry). By optimizing robot trajectories using B-spline curves guided by the Fisher Information Matrix, it enhances parameter observability. Although real-time online calibration with extended Kalman filtering is used, \textbf{the method remains sensitive to sensor quality, relying on DOA accuracy for microphones and scene features for LiDAR, and may degrade in complex acoustic or sparse visual environments without advanced processing}. Recently proposed method by Herau \textit{et al.} \cite{herau2024soac} uses Neural Radiance Fields (NeRF) to perform spatio-temporal calibration of multi-modal sensors without relying on physical targets or supervision. By training separate NeRFs for each camera and alternating between scene representation and cross-sensor pose optimization, it ensures alignment based only on overlapping regions. However, \textbf{the method struggles in scenes with linear motion (ambiguous time-space disentanglement), large open environments, and scales poorly with increasing sensor count due to training overhead}. These advances highlight the shift toward self-supervised, adaptable calibration, though practical deployment still requires balancing accuracy, generalizability, and efficiency in real-world conditions.

\textit{Noise Removal and Normalization.}
As traditional image denoising relied on handcrafted priors, statistical models, which often struggled with complex degradations and lacked adaptability, DL has served as a potential solution with a more adaptable and effective solution.
Building on this paradigm, recent models like QFormer \cite{jiang2024qformer} enhance efficiency by introducing a Quaternion Transformer Block that leverages quaternion algebra to model cross-channel color correlations while replacing self-attention with lightweight sequential operations. Although QFormer achieves state-of-the-art denoising performance with significantly fewer parameters and faster inference,  \textbf{its reliance on quaternion algebra and added skip-connection modifications increases implementation complexity and may limit applicability beyond color image denoising}. Fang \textit{et al.} \cite{fang2023partially} proposed a partially adaptive multichannel framework that jointly reduces ego-noise and environmental noise in human-robot interaction using a VAE-based speech model and NMF-based noise modeling. The approach pre-trains an ego-noise model to exploit its structured spatial and spectral patterns, while adapting the environmental noise model to unknown disturbances. However, \textbf{the method relies on accurate pre-training of ego-noise, which may limit its adaptability if the robot's hardware or operating conditions change}. Similarly, Qian \textit{et al.} \cite{qian2025sav} propose a novel framework that integrates Conformer and Mamba modules, enabling it to capture both global contextual dependencies and fine grained cross-modal representations. Despite its efficacy, \textbf{the method is currently limited to single speaker enhancement and does not address the more challenging problem of speaker separation in overlapping speech scenarios}. 
In 3D perception, RegBN \cite{ghahremani2023regbn} has been introduced as a regularization-based batch normalization framework tailored for multimodal data, addressing confounding effects and inter-modal dependencies without relying on learnable parameters. Demonstrated across diverse architectures and tasks, RegBN enables effective normalization of features from a broad spectrum of modalities (e.g., language, audio, image, video, depth, force, and proprioception). However, \textbf{RegBN incurs substantially higher training overhead, requiring per‐batch SVD and L-BFGS optimization when normalizing a pair of layers with dimensions 256×1024}.

\begin{figure}[t]
    \centering
    \vspace{-1em}
    \includegraphics[width=0.7\linewidth]{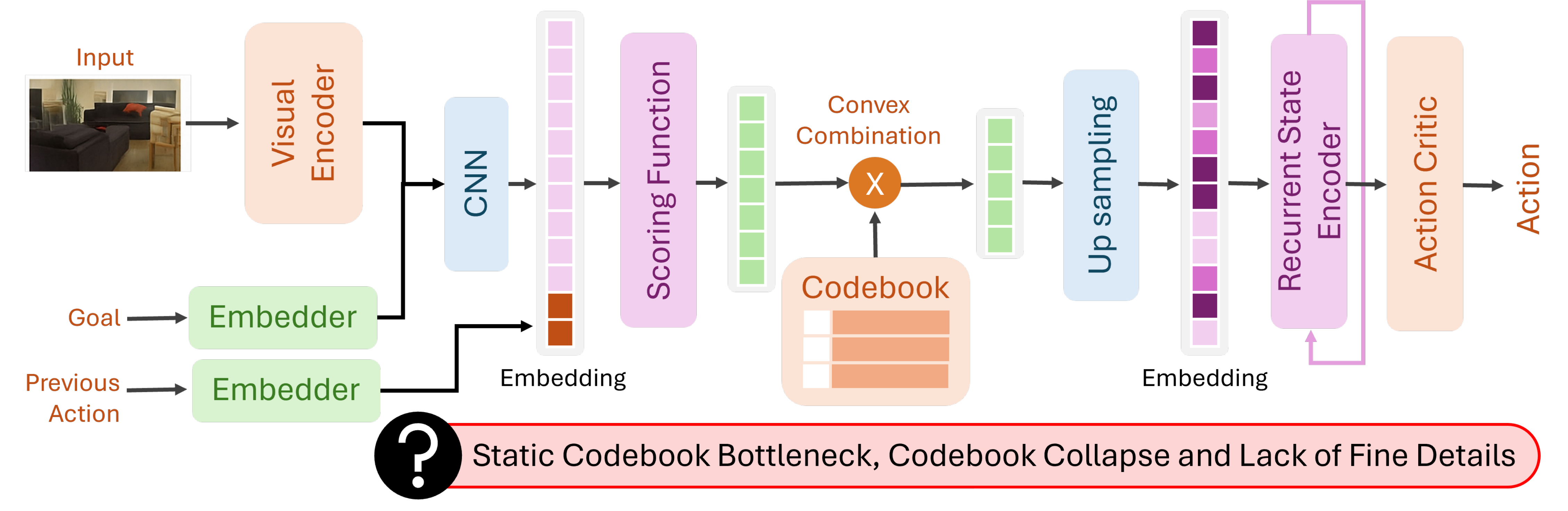}
    \caption{Architecture from \cite{eftekharselective}. Although the network achieves selective visual representation for improved generalization, the codebook bottleneck limits adaptability to novel environments and restrict the expressiveness of learned features, especially in dynamic or highly variable visual contexts.}
    \label{fig:multisensory_attention_controller}
    \Description{ipm_review}
\end{figure}

\textit{Multisensory Attention Controller.}
Recent advancements in multimodal perception and interaction have focused on adaptive fusion strategies and biologically inspired mechanisms to enhance task relevance in embodied agents. Li \textit{et al.} \cite{li2025adaptive} introduces AdapTac, an adaptive visuo-tactile fusion framework for dexterous manipulation that employs a force-guided attention mechanism to dynamically adjust visual and tactile weights across manipulation stages, reflecting modality priority shifts based on sensory relevance. While AdapTac demonstrates strong performance,\textbf{ its reliance on imitation learning without reinforcement fine-tuning limits robustness in highly dynamic or error-prone scenarios}. Complementing this, Lai \textit{et al.}\cite{lai2025fam} introduces FAM-HRI, a foundation-model-assisted framework for multimodal human-robot interaction that fuses real-time speech and gaze inputs from lightweight AR glasses. By leveraging LLM, it dynamically prioritizes relevant modalities based on user intent and context, enabling precise intent recognition and responsive robot control. However, \textbf{the reliance on heavyweight LLM/VLM inference causes processing delays, undermining real-time responsiveness}. Eftekhar \textit{et al.}~\cite{eftekharselective} introduces a lightweight, task-conditioned codebook bottleneck that filters general-purpose visual representations in embodied AI (Fig.~\ref{fig:multisensory_attention_controller}), enabling agents to focus selectively on task-relevant visual. By conditioning on goals and past actions, it prioritizes salient sensory inputs while suppressing irrelevant information. While the codebook module enhances task-relevant representation and generalization, \textbf{it suffers from codebook collapse, where only a few latent codes dominate and fine-grained visual details are lost, limiting both representational diversity and broader reasoning}. Yang \textit{et al.}~\cite{yang2024binding} proposed UniTouch, a framework for learning unified tactile representations through contrastive alignment with visual and auditory embeddings from pretrained foundation models. This multimodal alignment supports zero shot tactile question answering and grasp prediction, even in the absence of explicit touch-language pairs. By embedding tactile signals in a shared semantic space, UniTouch lays the groundwork for dynamic attention switching between modalities allowing agents to rely more heavily on touch in visually occluded or contact-rich scenarios. However, \textbf{the model lacks mechanisms for task-driven attention modulation, tactile inputs are treated uniformly regardless of context, and does not incorporate spatial anchoring or temporal memory, which limits its ability to dynamically adjust the salience of touch based on environmental uncertainty, manipulation stage, or goal relevance}. Gizdov \textit{et al.} \cite{gizdov2025seeing} introduces a foveated sampling technique that mimics human vision by allocating higher resolution to the center of fixation and lower resolution to the periphery, showing that large multimodal models can be guided to prioritize salient visual regions through foveated sampling alone, without requiring architectural modifications. The method however, uses a \textbf{single static fixation and lacks architectural adaptation or dynamic gaze mechanisms, limiting broader applicability and scalability}. Together, these works show a trend toward building flexible perception systems that are not only multimodal and adaptive, but also grounded in biological principles.

\textit{Cross-Modal Feature Integration.}
Recent advances in large language, vision, and multimodal models have greatly advanced AI agents, enabling richer reasoning and interaction across diverse sensory inputs~\cite{zhang2024vision}. Beyond language and vision, modalities such as audio and touch are increasingly integrated, supported by cross-modal architectures that align different signals within a unified representation space.
For instance, AVFormer \cite{seo2023avformer} introduces a lightweight approach to inject visual context into frozen speech recognition models by leveraging trainable adapters and visual projection layers. This design enables robust audiovisual speech recognition without requiring full model retraining. While highly data efficient and effective in zero shot settings, \textbf{AVFormer depends on separately pretrained visual and audio encoders and cannot be optimized in an end to end manner, which may constrain its adaptability for fine grained multimodal interaction}. Similarly, Gan \textit{et al.}\cite{gan2020looklistentact} validated the integration of sound and vision for navigation using a dual-branch Siamese network feeding into a GRU-based RL policy. \textbf{While effective in leveraging multimodal cues, the model lacked explicit spatial memory or mapping capabilities, limiting its robustness in acoustically challenging or ambiguous environments}. Similarly, MViTac \cite{dave2024multimodal} proposes a self supervised contrastive learning framework that integrates visual and tactile inputs to learn both intermodal and intramodal representations. It achieves strong performance on material classification and grasp prediction benchmarks, but \textbf{still lags behind supervised methods in grasp prediction tasks, largely due to the limitations of sparse and imbalanced training data}. Building on to unify representations across sensory modalities, Komarichev \textit{et al.} \cite{komarichev2022polyglot} proposed Polyglot, a contrastive learning framework that maps semantically related inputs, such as images, text, and audio, into a shared representation space. While it succeeds in aligning diverse modalities, the \textbf{resulting embeddings lack spatial interpretability, failing to capture both geometric distances and topological relationships}. Together, such approaches mark an important step toward unified learning systems that integrate perception and action, supporting the development of more adaptive grounded agents.
\begin{researchgap}
\textbf{Research Gap 1.} Recent multimodal sensory processing methods remain modality-specific, often incurring high computational costs, and still lack a unified, efficient framework capable of generalizing across heterogeneous sensors.
\end{researchgap}

\subsubsection{Egocentric-Allocentric Transformation}
\label{sec:spatial_reasoning_pipeline}

\begin{table*}[t]
\centering
\caption{Representative works in the Egocentric-Allocentric Transformation.}
\vspace{-0.5em}
\label{table:spatial_reasoning_pipeline}
\resizebox{\textwidth}{!}{%
\begin{tabular}{@{}l l l l l@{}}
\toprule
\textbf{Category} & \textbf{Method} & \textbf{Publication\&Year} & \textbf{Highlights} & \textbf{Limitations} \\
\midrule

\multirow{3}{*}{\makecell[l]{3D Egocentric\\Construction}} 
& Spatial Prediction~\cite{yang2025thinking} & ICLR'25 & Coarse layout generation using VLMs. & No geometric grounding. \\
& GS SLAM~\cite{yugay2023gaussianslam} & CVPR'24 & Dense 3D mapping from egocentric views. & No semantic understanding. \\
& Multi-View Memory~\cite{yang2024memory} & CVPR'2025 & Memory with frontier-based exploration. & No symbolic abstraction. \\

\midrule

\multirow{3}{*}{\makecell[l]{Scene\\Abstraction}} 
& Disjoint Reasoning~\cite{ravi2025outsight} & CVPR'25 & Implicit spatial consistency from VLMs. & No explicit memory. \\
& Cognitive Conversion~\cite{yin2025spatialmental} & arXiv & Symbolic egocentric-to-allocentric mapping. & No neural instantiation. \\
& 3D Prior Interaction~\cite{ma2024spatialpin} & NeurIPS'24 & Prompt-based reasoning with 3D priors. & Static priors. \\

\midrule

\multirow{3}{*}{\makecell[l]{Perspective\\Shift}} 
& Visual Perspective Taking~\cite{goral2024seeing} & NeurIPS'24 & VPT benchmark using VLM prompting. & Hallucination errors. \\
& Mental Imagery~\cite{lee2025perspective} & CVPR'25 & Perspective-aware reasoning via imagination. & Limited scene diversity testing. \\
& Pose-Aware Decoding~\cite{cartillier2021semanticmapnet} & AAAI'21 & Allocentric semantic map from egocentric. & Needs depth and supervision. \\

\bottomrule
\end{tabular}%
}
\end{table*}

As discussed in Sec.~\ref{sec:2.2.3}, the goal of this module is to mimic the act of ``imagination'', reconstructing the scene from a new point of view via intermediate geometric and symbolic abstractions. 

\textit{3D Contrusction and Scene Abstraction.} To simulate how the same scene would appear from a different viewpoint, Yang \textit{et al.}~\cite{yang2025thinking} demonstrated that VLMs can indeed form coarse spatial layouts from visual and textual cues. \textbf{However, these layouts are often imprecise and lack a firm grounding in the physical geometry of the scene, limiting an agent's ability to perform precise navigation or interaction.} The core of the problem lies in memory and continuity. Additionally, Ravi \textit{et al.}~\cite{ravi2025outsight} examined VLMs on long-horizon reasoning across frames where objects are never co-visible and found that models lag behind human performance by 28\%, with steeper declines in accuracy (60\% to 30\%) as the temporal gap widens, providing trajectories or bird's-eye-view projections to VLMs results in only marginal improvements, whereas providing oracle 3D coordinates leads to a substantial 20\% performance increase. \textbf{This highlights a core bottleneck of multi-frame VLMs in constructing and maintaining 3D scene representations over time from visual signals.} Encouraging from this challenge, Yugay \textit{et al.}~\cite{yugay2023gaussianslam} utilized 3D Gaussian Splatting to construct photorealistic maps, thereby constructing a dense, geometrically accurate, and visually rich 3D representation of the environment. This technique establishes the geometric backbone for a true allocentric map, advancing beyond the coarse spatial layouts of earlier VLM approaches. It produces a persistent, viewpoint-agnostic world model dedicated solely to geometry and rendering, without any semantic layer or reasoning capability. Complementing this geometric perspective, Yang et al.~\cite{yang2024memory} proposed a structured memory architecture that accumulates multi-view snapshots over time using frontier-based exploration. This supports long-term reasoning over spatial layouts and object positions, allowing for visually coherent and temporally grounded scene representations. \textbf{However, it lacks a symbolic interface or capacity for linguistic querying, limiting its general-purpose reasoning utility.} As can be seen, the discrete scene abstraction (object types, 3D boxes, headings) omits fine-grained details and uncertainty modeling, resulting in a non-differentiable, non-probabilistic pipeline. Ma \textit{et al.}~\cite{ma2024spatialpin} addressed that by proposing SpatialPIN, integrating 3D priors, such as NeRFs and point clouds into the prompting process. It enables VLMs to reason over spatial structure more faithfully, showing that geometry-aware prompting can complement latent mental simulation. \textbf{Though its reliance on precomputed 3D assets limits its applicability in open-world or dynamic environments.} Together, these works reveal a growing divide in the field: between systems optimized for geometric memory and those focused on symbolic spatial abstraction, with neither achieving the full spectrum of cognitive flexibility required for egocentric-to-allocentric transformation.


\textit{Visual Perspective-Taking.} Another key requirement for efficient egocentric to allocentric conversion is visual perspective-taking (VPT), the ability to query a rich 3D map to imagine a scene from another viewpoint. Accordingly, Goral\textit{ et al.}~\cite{goral2024seeing} systematically evaluated existing models on VPT tasks, in which they revealed a critical weakness: rather than performing genuine geometric transformations, most VLMs tend to ``hallucinate'' the new perspective textually. \textbf{Even with sophisticated prompts, performance fails to generalize as the model lacks multi-view fusion, mesh representation, scene graphs, or implicit tokens for simulating viewpoint shifts.} For more explicit mechanisms, Lee \textit{et al.}~\cite{lee2025perspective} proposed a novel method based on ``mental imagery simulation''. This approach aims to replace unstructured hallucination with a structured two-step pipeline of abstraction and geometric projection. However, \textbf{because each step is pretrained separately and manually integrated, errors compound across stages, in which if object detection fails (e.g., due to occlusion, clutter, or unusual objects), the entire abstraction collapses}. Moreover, Cartillier \textit{et al.}~\cite{cartillier2021semanticmapnet} proposed the SMNet, which learns to project egocentric RGB-D views into a global, allocentric semantic map using a falling-tensor memory and neural decoding module. It enables symbolic queries such as ``find the chair'' and supports basic navigation by embedding explicit object categories in space. \textbf{However, it assumes near-perfect perception and lacks mechanisms for viewpoint-conditioned simulation or tracking across time.}
This shift replaces hallucinated imagery with structured 3D priors, providing an efficient middle ground between full scene memory and latent reasoning. 
\begin{researchgap}
\textbf{Research Gap 2.} Current methods for 3D scene construction and visual perspective-taking remain fragmented between geometric and symbolic approaches, often relying on brittle or hallucinated models, and thus lack unified, persistent 3D representations that combine geometric grounding with semantic cues.
\end{researchgap}

\subsubsection{Cognitive Map in Agents}
\label{sec:3.A.4}

\begin{table*}[t]
\centering
\caption{Representative works in cognitive map for spatial reasoning.}
\vspace{-0.5em}
\label{table:3.A.4}
\resizebox{\textwidth}{!}{%
\begin{tabular}{@{}l l l l l@{}}
\toprule
\textbf{Category} & \textbf{Method} & \textbf{Publication\&Year} & \textbf{Highlights} & \textbf{Limitations} \\
\midrule

\multirow{4}{*}{\makecell[l]{Grid Cells\\ Layer}}
 & Grid-DRL~\cite{banino2018vector} & Nature’2018 & DRL and LSTM for path-integration & Fixed, supervised cues and hard-wired anchoring  \\
 & PF-Grids~\cite{sorscher2019unified} & NeurIPS’2019 & Pattern-formation for RNN/LSTM & Lacks anchoring to landmarks and drift correction \\
 & PI-ANN~\cite{schaeffer2022no} & NeurIPS’2022 & Evaluation on  DL-based grid-cell models & Grid requires specific and biologically setup \\
 & Grid-SSL~\cite{Rylan2023gridneurips} & NeurIPS’2023 & Self-supervised RNN on egocentric motion & Rely on self-motion cues; no landmark anchoring \\
 
\midrule

\multirow{4}{*}{\makecell[l]{Place Cells\\ Layer}}
 & GNN~\cite{battaglia2018relational} & DeepMind’2018 & Relational GNN for topological space & Lacks context-remapping; no spatial field \\
 & Place-RNN~\cite{neurips2024place} & NeurIPS’2024 & Similarity-based objectives for remapping & Lacks heterogeneous/multi-field coding \\
 & VPC~\cite{gornet2024automated} & NMI’2024 & Visual predictive coding with self-attention & Lacks focus on biological place-cell functions \\
 & FTM~\cite{cui2024frontier} & ECCV’2024 & Frontier topological memory with NeRF & Only targets landmark navigation \\
\bottomrule
\end{tabular}%
}
\end{table*}

As described in Sec.~\ref{sec:2.B.4}, the Cognitive Map Module integrates a grid cells layer for metric spatial representation and a place cells layer for topological, context-aware representation, together forming an internal spatial model for reasoning in dynamic environments. In this section, we review SOTA works within this module to examine emerging techniques and remaining gaps toward achieving a human-like cognitive map in AI agents. Table~\ref{table:3.A.4} presents the key highlights and gaps of the representative works.

\textit{Grid Cells in Cognitive Maps.} To enable grid-like representations of cognitive maps in deep learning models, several works~\cite{sorscher2019unified},~\cite{Rylan2023gridneurips},~\cite{banino2018vector},~\cite{schaeffer2022no} have explored the metric features within the neural networks. To that end, Sorscher \textit{et al.}~\cite{sorscher2019unified} links normative learning and mechanistic attractor models by casting position-encoding learning dynamics as a pattern-formation (PF) system, proving that representational constraints (notably nonnegativity) select hexagonal lattices, thereby offering a principled route to the grid cells layer. By analyzing in various neural networks, such as recurrent neural network (RNN) and long short-term neural network (LSTM), the authors explain the efficiency of hexagonal grids within spatial representation across different network architectures. Despite the strength of its theoretical framework, \textbf{this work does not account for several essential functions of biological grid cells in the human cognitive map, including anchoring to external landmarks, correcting drift over extended trajectories, and integrating with contextual and memory systems to support robust spatial reasoning}. To focus on more practical methods, Schaeffer \textit{et al.}~\cite{Rylan2023gridneurips} trains an RNN with self-supervised objectives (e.g., path-invariance, capacity, separation) on egocentric motion, yielding multi-modular grid codes that generalize beyond training. Thereby, this work directly instantiates the grid cells layer for metric encoding and path integration, offering a practical method for learning grid-like lattices in functioning agents. However, \textbf{this grid-based self-supervised learning (Grid-SSL) approach relies primarily on self-motion cues and lacks landmark/sensory anchoring and drift correction mechanisms and does not integrate grid codes with memory/goal signals for long-horizon, human-like spatial reasoning}. Additionally, Banino\textit{ et al.}~\cite{banino2018vector} propose a deep reinforcement learning (DRL)-based path-integration LSTM that develops entorhinal-like grid units with DRL to perform vector-based navigation. This work bridges the DRL context to our grid cells layer by forming an LSTM with a dropout-regularized linear layer to yield grid, head-direction, border-like units and a vision module to inject place/head-direction patterns, while a policy LSTM consumes the current and remembered ``goal grid code'' to guide actions. Nevertheless, \textbf{this model relies on fixed, supervised cues and hard-wired anchoring, which limits context-dependent remapping and prevents the flexible landmark integration and memory interaction seen in human grid cells.} Moreover, standing on a critics view, Schaeffer et al.~\cite{schaeffer2022no} critically evaluate DL-based grid cells models trained on path integration tasks, showing that \textbf{grid-like representations only emerge under highly specific and biologically implausible architectural, readout, and hyperparameter choices}, rather than as a natural outcome of the task. As a result, \textbf{current DL methods fail to capture key properties of biological grid cells, including stable multi-module coding, correct period ratios, robustness to sensory heterogeneity, and context-linked anchoring and drift correction.}

\begin{figure}[t]
    \centering
    \begin{subfigure}[b]{0.5\textwidth}
        \centering
        \includegraphics[width=\linewidth]{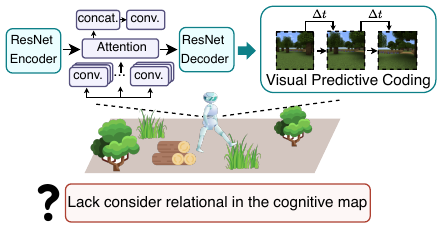}
        \caption{Visual Predictive Coding from~\cite{gornet2024automated}.}
        \label{fig:nmi}
    \end{subfigure}%
    \hfill
    \begin{subfigure}[b]{0.5\textwidth}
        \centering
        \includegraphics[width=\linewidth]{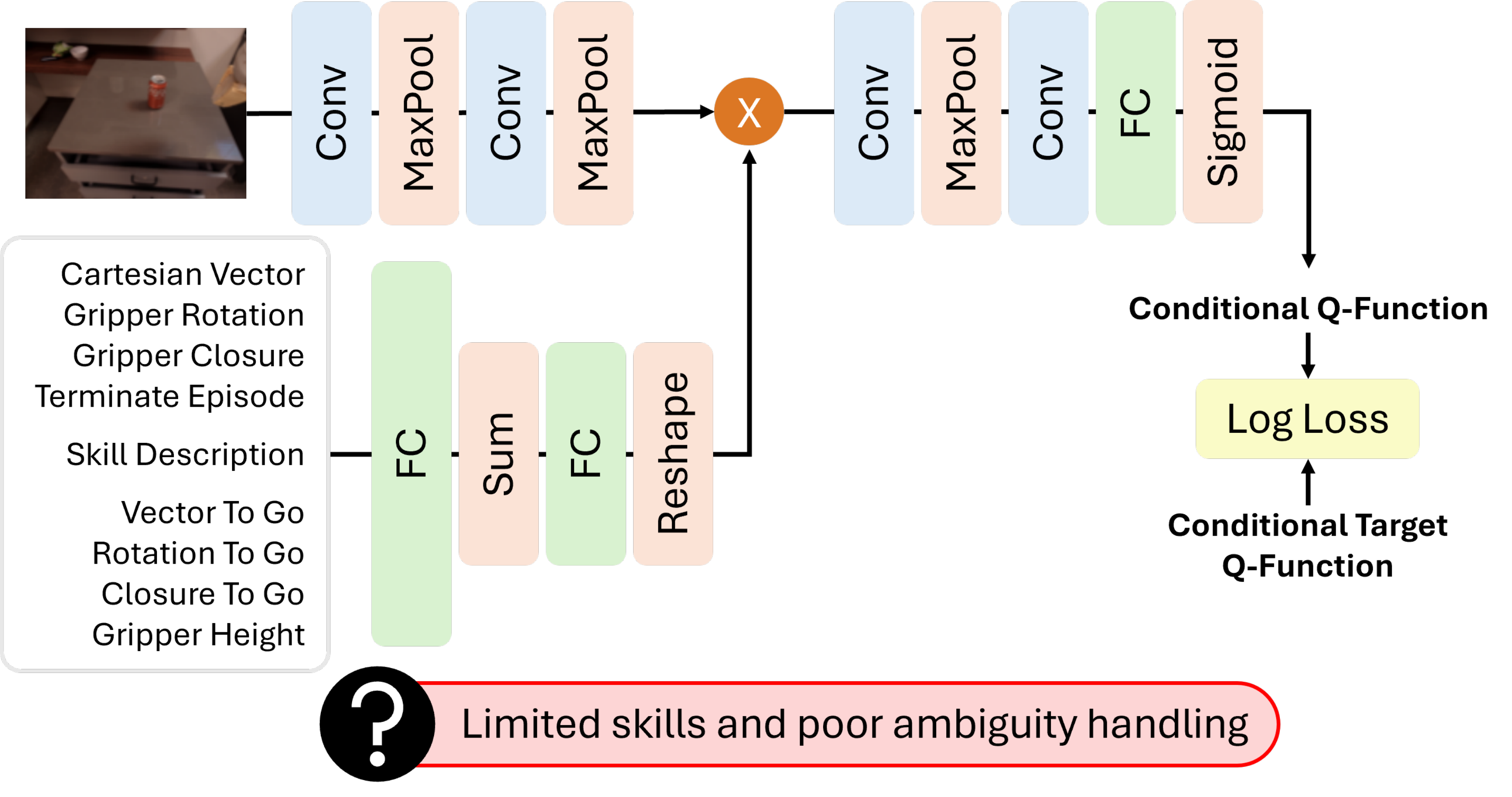}
        \caption{Architecture of Google's SayCan \cite{ahn2022can}.}
        \label{fig:saycan}
    \end{subfigure}
    \caption{The representative works tailored in (a) cognitive map module and (b) spatial neural memory module}
    \label{fig:review_fig}
    \Description{review_fig}
\end{figure}

\textit{Place Cells in Cognitive Maps.} Recent research on place cells has increasingly focused on exploring their topological nature and contextual flexibility, from graph-based representations of object–location relationships and mechanisms for contextual remapping to their role in linking with episodic memory and predicting future spatial states. An early work of Battaglia \textit{et al.}~\cite{battaglia2018relational} has introduced graph neural network (GNN) as a powerful framework for modeling relational reasoning, where nodes and edges naturally capture the topological structure of spatial environments, making it highly relevant to the topological coding of place cells in the cognitive map. However, \textbf{such graph-based models typically lack key biological characteristics of place-cell topology, such as context-dependent remapping, non-uniform spatial field distributions, and integration with landmark and reward signals observed in the human cognitive map.} To this end, Pettersen \textit{et al.}~\cite{neurips2024place} enables the contextual-remapping of place cells in neural networks by constructing a similarity-based objective that makes population codes mirror distances in space. Thereby, the proposed method yields place-like representations in feedforward networks that can flexibly reorganize their spatial maps across different contexts, achieving decorrelated activation patterns akin to global remapping. However, \textbf{the approach assumes simplified scalar context inputs and supervised position labels, and does not capture key biological properties of place cells such as heterogeneous field sizes and multi-field coding.} Additionally, Gornet \textit{et al.}~\cite{gornet2024automated} presents a visual predictive coding (VPC) method with a self-attention layer that learns latent ``place-like'' units anchored to visual landmarks, supporting landmark-based localization and vector navigation much like place cells use landmark cues. However, \textbf{it primarily targets topological/vector navigation and landmark localization from vision, without modeling key biological features of place cells.} Besides, Cui \textit{et al.}~\cite{cui2024frontier} proposes a frontier-enhanced topological memory (FTM) that augments a graph of explored places with NeRF-rendered ghost nodes for unseen frontiers, enabling landmark-guided navigation. Although this work aligns well with the topological-based memory indexing role of biological place cells, it primarily targets landmark-guided navigation and \textbf{does not address other complementary functions such as contextual remapping or integration with episodic memory.}
\begin{researchgap}
\textbf{Research Gap 3.} Current DL models of grid and place cells representations often succeed only under narrow or highly specific settings, lacking generalization and failing to capture core biological functions such as landmark anchoring, drift correction, multi-field coding, and context-dependent remapping.
\end{researchgap}

\subsubsection{Spatial Neural Memory Module}


\begin{table*}[t]
\centering
\caption{Representative works in Spatial Neural Memory Modules for embodied spatial reasoning.}
\vspace{-0.5em}
\label{table:3.A.5}
\small
\resizebox{\textwidth}{!}{%
\begin{tabular}{@{}l l l l l@{}}
\toprule
\textbf{Category} & \textbf{Method} & \textbf{Publication \& Year} & \textbf{Highlights} & \textbf{Limitations} \\
\midrule

\multirow{3}{*}{\makecell[l]{Spatial–Semantic\\ Encoding}} 
 & BEVFormer~\cite{li2024bevformer} & TPAMI'2024 & BEV map from multi-camera & Less effective than LiDAR-based methods \\
 & SayCan~\cite{ahn2022can} & PMLR'2023 & LLM + value functions & Limited skills; poor ambiguity handling \\
 & Meerkat~\cite{chowdhury2024meerkat} & ECCV'2024 & Audio-visual grounding & Struggles with occlusion, overlap, partial grounding. \\

\midrule

\multirow{2}{*}{\makecell[l]{Episodic\\ Memory}} 
 & Episodic Transformer~\cite{pashevich2021episodic} & ICCV'2021 & Full episode attention & Needs annotations; poor real-world transfer \\
 & SMT~\cite{fang2019scene} & CVPR'2019 & Attention over unstructured memory & High compute; approximation errors \\

\midrule

\multirow{2}{*}{\makecell[l]{Continual Memory\\ Updating}} 
 & Lu et al.~\cite{lu2025continual} & Preprint'2025 & Prompt tuning in CLIP & Evaluated only with CLIP; generalization untested \\
 & HiMemFormer~\cite{wang2024himemformer} & Preprint'2024 & Hierarchical memory for action & Lacks adaptive memory weighting \\

\bottomrule
\end{tabular}%
}
\end{table*}

\label{sec:3.A.5}
Spatial intelligence necessitates the ability to perceive, interpret, and adapt to dynamic environments. As discussed in Section \ref{sec:2.2.5}, effective navigation and interaction require agents to construct spatial-semantic representations, retain episodic memories, and continuously update internal memory structures in response to environmental changes. 

\textit{Spatial–Semantic Encoding}. provides agents with a unified representation that binds spatial geometry with semantic meaning. This process unfolds through grounded visual perception and non-visual multimodal grounding. Visual SLAM remains foundational in constructing egocentric metric maps via fused odometry, depth, and image-based data. These geometric scaffolds are enriched by 3D scene understanding, which leverages point clouds and learned features to differentiate object categories, surfaces, and boundaries. BEVFormer \cite{li2024bevformer} is a spatiotemporal transformer framework designed to learn bird’s-eye-view (BEV) representations directly from multi-camera inputs, enabling accurate and efficient 3D object detection and map segmentation. By introducing grid-shaped BEV queries with spatial cross-attention and temporal self-attention, it achieve state-of-the-art results on various benchmark datasets. BEVFormer's performance is still \textbf{limited by the inherent challenges of accurately inferring 3D spatial information from 2D camera inputs, making it less effective and efficient than LiDAR-based methods}. SayCan \cite{ahn2022can} grounds large language models in robotic affordances by combining language-driven task planning with reinforcement learned value functions that evaluate the feasibility of actions in real world contexts. This integration enables zero shot execution of complex, temporally extended instructions with significantly improved performance over non grounded baselines. However, SayCan is \textbf{constrained by the coverage and reliability of its pretrained skill set; if a required skill is absent or fails, the system lacks adaptability}. Moreover, \textbf{it inherits the reasoning limitations of language models, particularly in handling negation, ambiguity, and novel scenarios}. Meerkat \cite{chowdhury2024meerkat} introduces the first unified audio-visual large language model that achieves fine-grained spatial and temporal grounding by aligning image and audio features through optimal transport and cross-modal attention. Meerkat extracts CLIP-based image and CLAP-based audio features, aligns them via a weakly supervised Optimal Transport module (AVOpT), and refines region-level attention using the Audio-Visual Attention Consistency Enforcement (AVACE) module. Despite the good results, \textbf{Meerkat still struggles with occluded objects and overlapping sounds, often mislocalizing regions and capturing only partial event durations}.

\textit{Episodic Spatial Memory.} Episodic memory based architectures have emerged as a powerful paradigm for long-horizon vision and language navigation by enabling agents to reason over extended temporal contexts. Episodic Transformer \cite{pashevich2021episodic} is a multimodal transformer architecture that processes the full episode history of visual observations, actions, and language instructions to overcome the limitations of recurrent models. It employs modality-specific encoders and fuses the encoded streams through a multimodal transformer with causal attention to predict the next action. \textbf{While effective, its performance is highly dependent on high quality synthetic annotations, limiting its applicability in real world environments where such supervision is scarce}. Complementing this, Scene Memory Transformer (SMT) \cite{fang2019scene} introduces a memory-based policy that stores all past observations in an unstructured memory and uses attention mechanisms to aggregate spatio-temporal context for action selection in long-horizon tasks. Unlike RNNs or structured memory approaches, SMT employs a transformer-based encoder-decoder architecture trained with DRL to enable more flexible and expressive reasoning. However, \textbf{its attention-based design incurs quadratic computational cost with increasing memory size, which can be prohibitive in extended episodes}. While memory factorization reduces this complexity to linear time, \textbf{it introduces approximation that may degrade performance in more complex scenarios}. These architectures demonstrate the promise and trade-offs of transformer-based memory systems in enabling scalable and context-aware navigation in embodied agents.

\textit{Adaptive Memory Updating.} Dynamic environments demand not only memory retention but also continual adaptation. To operate effectively in such settings, spatial reasoning agents must update their internal models without catastrophic forgetting and flexibly accommodate changes over time and space. Lu \textit{et al.}\cite{lu2025continual} introduces TPPT-V and TPPT-VT, two lightweight continual learning strategies for CLIP that use static and learnable textual prototypes to guide visual prompt tuning. TPPT-V anchors visual representations to fixed class text embeddings to mitigate forgetting, while TPPT-VT jointly tunes visual and textual prompts with a diversity regularization to prevent representation collapse and enhance adaptation. However, \textbf{the method has only been evaluated with CLIP; its generalization, computational overhead, and efficacy on other vision–language backbones or broader continual-learning settings remain untested}. Complementing these methods, HiMemFormer \cite{wang2024himemformer} introduces a transformer-based architecture for online multi-agent action anticipation by incorporating both agent-specific and global contextual memory. It hierarchically encodes long and short-term temporal cues, employing an Agent-to-Context Encoder to merge agent-centric and contextual information, followed by a Context-to-Agent Decoder that performs coarse-to-fine action prediction using short-term memories. The model is trained with dual-stage supervision, applying cross-entropy loss to both coarse and refined predictions. But \textbf{HiMemFormer relies on fixed hierarchies for long and short-term memory without adaptive weighting or clear interpretability, which may limit its responsiveness to dynamic environmental changes}. 
\begin{researchgap}
\textbf{Research Gap 4.} Current approaches tailored with spatial neural memory struggle to achieve robust geometry–semantics fusion under occlusion and novelty, scalable and data-efficient episodic reasoning, and adaptive memory updating with generalizability and interpretability.
\end{researchgap}

\subsubsection{Spatial Reasoning in Agents}
\label{3.1.6}

\begin{table*}[t]
\centering
\caption{Representative works in reasoning module for spatial reasoning.}
\vspace{-0.5em}
\resizebox{\textwidth}{!}{%
\begin{tabular}{@{}l l l l >{\raggedright\arraybackslash}p{4.5cm}@{}}
\toprule
\textbf{Category} & \textbf{Work} & \textbf{Publication \& Year} & \textbf{Highlights} & \textbf{Limitation} \\
\midrule

\multirow{3}{*}{\makecell[l]{Predictive World Modelling}}
 & ~\cite{hafner2023mastering} & Nature’2025 & Optimizes behavior in imagined future trajectories. & \multirow{3}{*}{\makecell[l]{Lacks a multi-step, \\structured spatial reasoning}}\\
 & ~\cite{zhou2024dino} & NYU’2024 & Predicts future dynamics in the latent space. &\\
 & ~\cite{zhen20243d} & UMass’2024 & Integrates 3D-LLM with aligned diffusion models. &\\
\midrule
\multirow{3}{*}{\makecell[l]{Explicit Spatial \\Reasoning and Alignment}}
 & ~\cite{bigverdi2025perception} & CVPR’2025 & Generates and utilizes explicit spatial information. & \multirow{3}{*}{\makecell[l]{Fails to build a unified, \\dynamic world model}}\\
 & ~\cite{li2025imagine} & ICML’2025 & Generates visual thoughts as part of the reasoning. &\\
 & ~\cite{wu2025reinforcing} & Ant Group’2025 & Performs drawing operations in the visual space. &\\

\bottomrule
\end{tabular}%
}
\end{table*}

As the executive hub of our framework, the Reasoning Module transforms structured knowledge from the cognitive map into goal-oriented intelligent actions through world-model-based learning and explicit spatial reasoning. This section will analyze past research to examine how existing methods have tackled the challenge of spatial reasoning and discuss their relationship to the core components of our proposed module.

\textit{Predictive World Modelling.}
The DreamerV3 agent proposed by Hafner \textit{et al}.~\cite{hafner2023mastering} learns a predictive world model to simulate future possibilities, which directly corresponds to the world-model-based learning emphasized in our framework for planning. This method trains a Recurrent State-Space Model to predict future latent states and subsequently uses an actor-critic RL approach to derive optimal actions within imagined trajectories. However, \textbf{its reasoning process is implicit, with its policy being learned directly by maximizing rewards in simulation}. Therefore, this approach lacks an explicit, multi-step, and structured spatial reasoning paradigm, such as generating intermediate representations or following a CoT process.
Zhou et al.~\cite{zhou2024dino} presented DINO-WM, a world model that learns to predict dynamics within the latent space of pre-trained predictive visual features learned from DINOv2. The model achieves zero-shot visual planning and control by optimizing a sequence of actions at test time to reach a goal state in this latent space. However, as a JEPA-based predictive architecture, \textbf{it faces notable challenges including accurately modeling long-horizon dynamics, mitigating error accumulation during latent rollouts, and ensuring robust generalization beyond the visual distributions seen during pre-training.} Apart from traditional DL and emerging predictive learning, Zhen et al.~\cite{zhen20243d} introduced 3D-VLA, a generative-based approach that utilizes diffusion models for world modelling. By continuously predicting future multimodal scenes (e.g., images, point clouds), it  realizes the cognitive map as a predictive internal simulator to support subsequent planning. Nevertheless, \textbf{continuous generation via diffusion is highly resource-intensive and may not align with the compact, abstracted internal mental models employed by humans for efficient reasoning and planning, as it can be treated as explicit scene modeling rather than the implicit, internally maintained representations humans use.}

\textit{Explicit Spatial Reasoning and Alignment.}
Bigverdi et al.~\cite{bigverdi2025perception} proposed the ``Perception Tokens'' which align with the core idea of ``Explicit Spatial Reasoning'' in our reasoning module by enabling multimodal models to generate auxiliary information, such as depth maps or bounding boxes, as intermediate reasoning steps. This method uses a VQ-VAE to convert intermediate image representations (like depth maps) into a tokenized format and employs a multi-task training framework to have the model leverage these visual perception tokens for reasoning. However, \textbf{this work primarily focuses on distilling representations from specific visual tasks into a language model, rather than building a unified, dynamic world model that can be flexibly used for planning and prediction.}
The Multimodal Visualization-of-Thought (MVoT)~\cite{li2025imagine} framework introduces a reasoning paradigm for MLLMs that generates interleaved verbal and visual thoughts, enabling the model to visualize reasoning traces for improved performance in spatial tasks like maze navigation and environmental interactions. It corresponds with our Reasoning Module's emphasis on explicit spatial reasoning by constructing visual aids to support multi-step inferences. However, \textbf{MVoT is mainly concerned with generating static image visualizations which falls short of fully realizing our neuro-inspired module's goals.}
ViLaSR~\cite{wu2025reinforcing} proposes a ``drawing to reason in space'' framework, which enables vision-language models to solve complex spatial problems by drawing visual markers like bounding boxes and auxiliary lines on images. However, \textbf{this approach remains limited, as it emphasizes external visual input without enabling agents to simulate environmental dynamics.}
\begin{researchgap}
\textbf{Research Gap 5.} Current reasoning approaches either rely on implicit world-model learning that lacks explicit multi-step spatial reasoning, or on explicit visualization techniques that remain task-specific, static, and tied to external inputs. As a result, no existing method offers a unified reasoning framework that combines predictive world models with explicit, and neuro-inspired spatial reasoning to support robust planning and goal-directed behavior.
\end{researchgap}

\subsection{Datasets and Emerging Applications}
\label{sec:3.B}
\subsubsection{Datasets and Benchmarks}

To achieve human-level spatial intelligence, agentic systems must be evaluated on spatial reasoning tasks that align with the full spectrum of human spatial cognitive processes. While numerous 3D vision and spatial reasoning benchmarks have been proposed for AI, they are rarely organized in a way that clearly maps to specific cognitive processes observed in humans. To this end, we address this problem by providing a systematic analysis of the most representative benchmarks, structured according to a spatial reasoning behavior foundationed by HAI (mentioned in Sec.~\ref{sec: 2.B.7}) with the core levels: 3D perceptual inference, hidden-state inference, and policy selection.

\textit{3D Perceptual Inference:} At the level of low-level cognition, agents must be equipped to process diverse sensory modalities, enabling perception not just through vision, but also via auditory, motion, tactile, and bio-signal inputs.

\textbf{Visual.} In terms of low-level 3D visual inference, \textbf{NYU Depth-v2}~\cite{silberman2012indoor} delivers 1,449 densely annotated indoor RGB-D frames and remains a foundational benchmark for evaluating how well models infer depth from a single image. \textbf{ObjectNet3D}~\cite{xiang2016objectnet3d} bridges web photos and precise 3D geometry by aligning 90k images with CAD meshes across 100 categories, making it a definitive benchmark for joint category recognition and 6-Degree of Freedom (DoF) pose estimation under the clutter, occlusion, and appearance variation that synthetic data rarely captures. \textbf{MVSEC}~\cite{zhu2018multivehicle} is recorded through an event camera, delivering microsecond-level temporal resolution, challenging networks to handle high-speed motion and extreme lighting that ruin conventional frames. Its synchronized IMU and LiDAR streams enable the fusion of event data with additional sensors for tight SLAM and optical-flow benchmarks. \textbf{ARKitScenes}~\cite{baruch1arkitscenes} provides 1.6 million RGB-D frames with scene meshes and ARKit poses, showing what large-scale 3D indoor reconstruction looks like ``in the wild'' with commodity hardware. The built-in ARKit poses make it a playground for research on relocalization and neural rendering from casually scanned spaces. Finally, \textbf{Objaverse-XL}~\cite{deitke2023objaverse} aggregates 10 million richly textured 3D assets scraped from the web, dwarfing prior shape repositories and powering contemporary 3D generative models. Its sheer scale and category breadth encourage zero-shot generalization and have made it the go-to corpus for novel-view synthesis and web-scale object understanding.

\textbf{Motion.} Apart from visual, motion is also essential in low-level cognitive processing. \textbf{DIP-IMU}~\cite{DIP2018} records 64 motion-capture sequences from 10 participants equipped with 17 Xsens inertial sensors, yielding over 330,000 time-aligned frames with both IMU signals and optical motion capture ground truth. As a foundational dataset for body-pose estimation from sparse sensors, it plays a key role in equipping agentic systems with proprioceptive inference capabilities, enabling low-latency full-body pose reconstruction when visual input is limited or absent. \textbf{AMASS}~\cite{mahmood2019amass} consolidates 15 heterogeneous motion capture datasets into a unified SMPL representation, covering over 11,000 motion clips and more than 40 hours of activity from 344 subjects. Its richness in action diversity and mesh-level consistency makes it a crucial resource for training agents to generalize across complex motion patterns, supporting both motion synthesis and temporal reasoning grounded in physical embodiment. \textbf{EgoBody}~\cite{zhang2022egobody} captures 36 subjects in 125 indoor scenes using a unique hybrid setup: one actor wears a HoloLens2 headset to collect egocentric RGB-D, head pose, and gaze data, while external cameras simultaneously record ground-truth SMPL meshes. With over 200,000 paired frames linking first-person perception to third-person full-body motion, EgoBody enables agents to learn from both embodied and allocentric viewpoints, bridging perceptual input with internal motion states in dynamic spatial contexts.

\begin{table*}[t]
\centering
\vspace{-0.5em}
\caption{Representative Datasets/Benchmarks For Spatial Reasoning Evaluation.}
\label{table:datasets_spatial_reasoning}
\resizebox{\textwidth}{!}{%
\begin{tabular}{c c c c c c}
\toprule
\textbf{Category} & \textbf{Name} & \textbf{Year} & \textbf{Task} & \textbf{Type} & \textbf{Contents} \\
\midrule

\multirow{22}{*}{\rotatebox{90}{\textbf{3D Perceptual Inference}}} 
  & NYU Depth-v2~\cite{silberman2012indoor} & 2012 & Monocular Depth Estimation & Visual & 1,499 depth RGB-D frame \\
  & ObjectNet3D~\cite{xiang2016objectnet3d} & 2016 & 3D Object Recognition & Visual & 90,127 images; 100 categories \\
  & MVSEC~\cite{zhu2018multivehicle} & 2018 & Event-based Tasks & Visual & 12 event streams \\
  & ARKitScences~\cite{baruch1arkitscenes} & 2021 & 3D Indoor Reconstruction & Visual & 1.6M RGB-D frames with meshes \\
  & Objaverse-XL~\cite{deitke2023objaverse} & 2023 & Novel View Synthesis & Visual & 10M 3D objects \\

\cmidrule(lr){2-6}
  & DIP-IMU~\cite{DIP2018} & 2018 & Human Pose Estimation & Motion  & 64 sequences; 330k instants \\
  & AMASS~\cite{mahmood2019amass}  & 2019 & 3D Motion Reconstruction  & Motion  & 11k motion-SMPL pairs \\
  & EgoBody~\cite{zhang2022egobody}& 2022 & Multi-View Motion Estimation & Motion & 200k+ RGB-D-SMPL pairs \\

\cmidrule(lr){2-6}
  & ReSkin~\cite{bhirangi2022reskin}   & 2021 & Force Estimation    & Tactile-Force  & 1.2M tactile events \\
  & 9DTact~\cite{lin20239dtact}  & 2023 & 6D Force Estimation  & Tactile-Force  & 100k images-torque pairs \\
  & FeelAnyForce~\cite{shahidzadeh2024feelanyforce}  & 2024 & 3D Contact Force Estimation  & Tactile-Force  & 200k indentations \\
  & FoTa~\cite{zhao2024transferable}  & 2024 & Tactile Sensing  & Tactile  & 3M tactile images \\
  & TVL~\cite{fu2024touch}& 2024 & Language-Tactile Descriptions & Tactile & 43,741 vision-DIGIT pairs \\
  & TV-Touch~\cite{huang20253d} & 2025 & Robotics Visuo-Tactile  & Tactile  & 40k visual-tactile pairs \\

\cmidrule(lr){2-6}
  & SoundSpaces~\cite{chen2020soundspaces} & 2020 & Audio-Visual Navigation & Auditory & 20k stereo audio-RGB pairs \\
  & Ego-Exo4D~\cite{grauman2024ego} & 2024 & 3D audio-vision tasks & Auditory & \begin{tabular}[c]{@{}c@{}}1,000 hours of; audio;\\ motions; multi-view video\end{tabular} \\
  & REASSEMBLE~\cite{sliwowski2025reassemble} & 2025 & Robotic Assembly & Auditory  & \begin{tabular}[c]{@{}c@{}}4,551 demos with audio;\\ force; vision; kinematics\end{tabular} \\

\midrule

\multirow{10}{*}{\rotatebox{90}{\textbf{Hidden-State Inference}}}  
  & 3DMatch~\cite{zeng20173dmatch} & 2017 & Point Cloud Recall  & Geometrical & 70k RGB-D images \\
  & CLEVR~\cite{johnson2017clevr} & 2017 & VQA & Topological & 100k RGB-D images; 1M QA \\
  & BOP~\cite{hodan2018bop} & 2018 & 6D Pose Estimation & Geometrical & 89 objects; 339k RGB-D images\\
  & 3D Scene Graph~\cite{armeni20193d} & 2019 & 3D Graph Prediction & Topological & 572 scene graphs \\
  & SCAN2CAD~\cite{avetisyan2019scan2cad} & 2019 & 9-DoF alignment & Geometrical & 97,607 key-point pairs \\
  & AGQA~\cite{grunde2021agqa} & 2021 & VQA & \begin{tabular}[c]{@{}c@{}}Spatiotemporal/ \\ Topological\end{tabular} & 9.6k videos; 192M QA \\
  & MindCube~\cite{yin2025spatialmental} & 2025 & VQA & Mental Simulation & 3,268 images; 21,154 QA \\
  & 3DSRBench~\cite{ma20243dsrbench} & 2025 & VQA & \begin{tabular}[c]{@{}c@{}}Geometrical/ \\ Topological\end{tabular} & 2,772 QA from MS-COCO \\
  & IntPhys2~\cite{bordes2025intphys} & 2025 & Dynamic Physical Inference & Spatiotemporal & 1,416 videos \\
  & SITE~\cite{wang2025site} & 2025 & VQA & \begin{tabular}[c]{@{}c@{}}Mental Simulation/ \\ Abstract Reasoning\end{tabular} & \begin{tabular}[c]{@{}c@{}}13,172 images; 3,808 videos; \\ and 8,068 QA\end{tabular} \\
\midrule

\multirow{5}{*}{\rotatebox{90}{\textbf{Policy Selection}}} 
  & AI2-THOR~\cite{kolve2017ai2} & 2017 & Navigation/Manipulation & Decision-Making & \begin{tabular}[c]{@{}c@{}}3D egocentric environment \\ simulator\end{tabular} \\
  & TDW~\cite{gan2020threedworld} & 2020 & Navigation/Manipulation & Decision-Making & \begin{tabular}[c]{@{}c@{}}3D simulator with \\ audio and visual\end{tabular} \\
  & Habitat 2.0~\cite{szot2021habitat}& 2021 & Navigation/Manipulation  & Decision-Making & \begin{tabular}[c]{@{}c@{}}3D environment with \\ multi-sensory\end{tabular} \\
  & iGibson 2.0~\cite{li2022igibson}& 2022 & Navigation/Manipulation  & \begin{tabular}[c]{@{}c@{}}Interactive \\ Decision-Making\end{tabular} & \begin{tabular}[c]{@{}c@{}}3D environment with \\ logical states\end{tabular} \\
  & Habitat 3.0~\cite{puig2023habitat}& 2023 & Human Robot Interaction  & \begin{tabular}[c]{@{}c@{}}Interactive \\ Decision-Making\end{tabular} & \begin{tabular}[c]{@{}c@{}}3D environment with SMPL \\ and multi-agents\end{tabular} \\
\bottomrule
\end{tabular}%
}
\end{table*}

\textbf{Tactile and Force Feedback.} As tactile is a robust sensing modality for bio-inspired agentic spatial reasoning, \textbf{ReSkin}~\cite{bhirangi2022reskin} couples a magnet-particle elastomer ``skin'' with a self-calibrating board to stream dense 3-axis magnetic-field readings at kilohertz rates, resulting in 1.2 million event regime. It captures slips, taps and impacts that potentially let the agents learn to localise micro-forces with the same spatial acuity human Merkel cells provide. \textbf{9DTact}~\cite{lin20239dtact} extends optical GelSight ideas into a 32 mm cube sensor and releases 100 k image-torque pairs over 175 objects; its pixel-accurate 3-D height fields plus full 6-D force labels teach agents to infer both geometry and wrench from a single touch, a prerequisite for reasoning about object affordances in clutter. \textbf{FeelAnyForce}~\cite{shahidzadeh2024feelanyforce} gathers 200 k robot-controlled indentations on GelSight-Mini while logging RGB-D and force vectors, then shows a transformer that generalises across unseen sensors. \textbf{Fota}~\cite{zhao2024transferable} aggregates more than 3 million tactile frames from 13 different sensors and 11 tasks into one unified format, enabling the formation of foundation tactile model that supports long-horizon manipulations and mirrors how large-vision corpora seeded visual commonsense in LLM. \textbf{TVL}~\cite{fu2024touch} bridges sensory and semantic space with 44k in-the-wild vision–touch pairs, which are annotated mostly from GPT-4V, enabling open-vocabulary tactile captioning. \textbf{TV-Touch}~\cite{huang20253d} provides 40k paired DigiT images and stereo RGB frames for dexterous gripper scenes, forming a 3D visuo-tactile benchmark that unifies grasp planning and force prediction. This benchmark can potentially become a high-level testbed for policy-centred agents.

\textbf{Auditory.} Regarding auditory sensing, \textbf{SoundSpaces}~\cite{chen2020soundspaces} augments the photorealistic Matterport3D and Replica worlds with geometrical-acoustic simulations, providing more than 20k stereo room-impulse-response/RGB pairs that let agents ``hear'' a target and learn audio-visual way-finding under realistic reverberation and occlusion. \textbf{Ego-Exo4D}~\cite{grauman2024ego} scales auditory perception into the social realm, curating 1,000 hours of multichannel audio synchronised with egocentric head-mounted video, exocentric multi-view cameras, 3-D point clouds, IMU, gaze and rich language commentary of skilled activities; by aligning what the actor hears with external observations, it trains agents to reason jointly from ``inside'' and ``outside'' perspectives during complex human motion. \textbf{REASSEMBLE}~\cite{sliwowski2025reassemble} turns sound into a manipulation cue: its 4,551 contact-rich assembly demonstrations record microphone waveforms alongside force–torque, event-camera, and multi-view RGB streams on the NIST Task Board, giving agents the multimodal evidence needed to predict subtle contact events, verify insertions, and plan corrective actions when visual data alone is ambiguous. 


\textit{Hidden-State Inference:} Hidden-state inference demands that an agents reconstruct what it cannot directly observe, such as latent geometry, topology, dynamics, and causal relations, and the current benchmark landscape targets these facets in complementary ways. \textbf{3DMatch}~\cite{zeng20173dmatch} learns local 3-D patch descriptors from 70k RGB-D fragments captured in noisy indoor scans, challenging models to recover geometric-based point-cloud correspondences despite occlusion and partial views. \textbf{CLEVR}~\cite{johnson2017clevr} focuses on Visual Question Answering (VQA) with 100k rendered scenes and 1 M template-controlled questions, forcing agents to topologically track object attributes and spatial relations that are implicit rather than visible at first glance. \textbf{BOP}~\cite{hodan2018bop} aggregates 339k RGB-D images of 89 real objects across multiple setups, providing pixel-accurate meshes and ground-truth 6-DoF poses so that hidden geometric features must be inferred under heavy clutter. \textbf{3D Scene Graph}~\cite{armeni20193d} converts whole ScanNet reconstructions into 572 hierarchical graphs (e.g., building-room-object-camera), evaluating whether a model can induce the latent relational topology of an environment rather than mere geometry. \textbf{SCAN2CAD}~\cite{avetisyan2019scan2cad} aligns CAD models to 1k Matterport scans via 97k manually annotated key-point correspondences, demanding 9-DoF alignment that links unseen canonical shape priors to noisy real scans. Moreover, \textbf{AGQA}~\cite{grunde2021agqa} provides a VQA baseline for spatiotemporal reasoning by pairing 9.6k synthetic videos with 192 M questions whose logic chains span multiple events. Therefore, agents must integrate hidden causal states across long temporal windows. \textbf{MindCube}~\cite{yin2025spatialmental} distills spatial-mental-simulation into 3,268 cube-stack images and 21,154 VQA-based questions that require imagining unseen faces and inside-out rotations, directly probing the ability to run internal mental models for spatial reasoning. \textbf{3DSRBench}~\cite{ma20243dsrbench} introduces 2,762 open-vocabulary 3D questions over COCO scenes, intertwining both geometric and topological cues so the model must infer implicit spatial relations beyond labeled objects. \textbf{IntPhys 2}~\cite{bordes2025intphys} delivers 1,416 photorealistic videos of object violations (permanence, solidity, continuity, immutability), testing whether systems can foresee physically plausible trajectories and flag hidden violations as infants do. Finally, \textbf{SITE}~\cite{wang2025site} also design a VQA that unifies 13,172 images, 3,808 videos, and 8,068 multi-choice prompts across scales from figural to environmental. It adds novel “view-taking” and dynamic-scene tasks so that spatial reasoning models must reconstruct intrinsic vs. extrinsic frames and unseen motion paths to answer correctly. Together these benchmarks trace a continuum from latent geometry to abstract mental simulation, providing the scaffolding needed to train and evaluate agents that reason beneath the surface.

\textit{Policy Selection:} At this level, spatial reasoning research leans on photorealistic simulators that let agents learn end-to-end decision-making under realistic physics and multisensory feedback. \textbf{AI2-THOR}~\cite{kolve2017ai2} set the stage with 120 household scenes, whose physics-driven arm enables agents to co-optimise where to navigate and how to manipulate objects. \textbf{ThreeDWorld (TDW)}~\cite{gan2020threedworld} raises the ceiling with near-photoreal visuals, high-fidelity spatial audio, cloth-and-liquid physics, a generative scene engine, and multi-agent support—exposing policies to cross-modal cues and complex material dynamics unavailable in earlier platforms. \textbf{Habitat 2.0}~\cite{szot2021habitat} pivots from static navigation to interactive rearrangement, combining GPU-accelerated rigid-body physics with thousands of scanned apartments so an agent must decide both where to move and how to reconfigure its surroundings. \textbf{iGibson 2.0}~\cite{li2022igibson} complements this with a logic-state sampler that spawns endless task variations (cook, slice, soak, freeze) and a VR teleoperation interface, supporting curriculum design and imitation learning for long-horizon household activities.  Pushing toward social intelligence, \textbf{Habitat 3.0}~\cite{puig2023habitat} introduces controllable humanoid avatars, accurate SMPL physics, and human-in-the-loop collaboration benchmarks, creating a sandbox where agents must negotiate shared goals with simulated people before venturing into real homes.

\subsubsection{Emerging Applications of AI Spatial Intelligence}
The spatial reasoning in agents is reflected in its wide-ranging applications across both virtual and physical domains, from immersive digital interfaces to real-world robotic systems. 

\textit{Virtual Applications:}
Virtual applications are increasingly demanding AI systems with the capability of interpreting and responding to spatial context. 

\textbf{Virtual Assistance Systems.} Agentic spatial reasoning is enabling assistance systems to move beyond basic dialogue toward context-aware, physically grounded interaction. Equipped with tools such as Meta’s Project Aria~\cite{engel2023project}, agents can interpret spatial language (e.g., ``behind the cabinet''), localize the user, and map relevant objects in real time. This capability supports emerging applications from everyday retail assistance to technical maintenance, where AR devices can highlight precise components directly within the user’s view.

\textbf{Immersive Virtual and Extended Reality.}
Agentic spatial reasoning is poised to become a cornerstone of immersive virtual and extended reality (VR/XR), empowering agents to perceive, interpret, and act seamlessly within persistent 3D environments. As industries invest heavily in the metaverse vision, through platforms like Meta Quest~\cite{meta_quest} and photorealistic systems such as Codec Avatars~\cite{metaImmersiveTelepresence}, agents with spatial thinking can bring these worlds to life. They enable NPCs that navigate virtual spaces with human-like awareness, adapt to user positions, and maintain natural, context-rich interactions. This means VR agents can guide human avatars through social hubs and virtual games by leveraging terrain for strategic advantage within a shared workspace.

\begin{figure}[t]
    \centering
    \vspace{-1em}
    \begin{subfigure}[b]{0.5\textwidth}
        \centering
        \includegraphics[width=\linewidth]{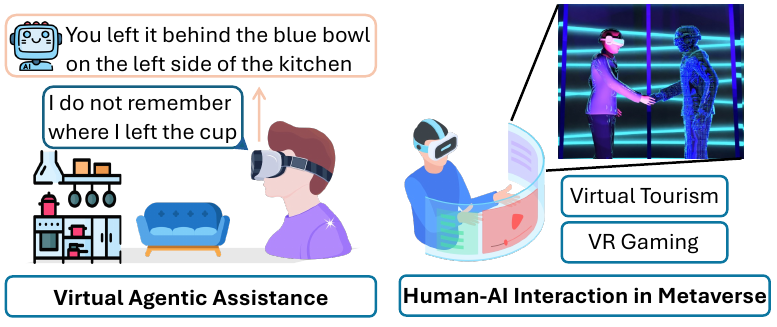}
        \caption{Virtual Applications.}
        \label{fig:sumcol}
    \end{subfigure}%
    \hfill
    \begin{subfigure}[b]{0.5\textwidth}
        \centering
        \includegraphics[width=\linewidth]{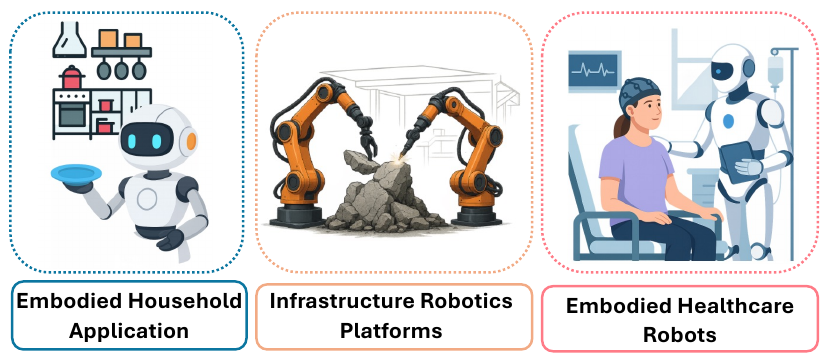}
        \caption{Embodied Physical Applications. }
        \label{fig:sumrow}
    \end{subfigure}
    \vspace{-2em}
    \caption{Applications from Agentic Spatial Intelligence, including (a) Virtual and (b) Physical Applications.}
    \label{fig:colrow}
    \Description{app}
\end{figure}

\textit{Embodied Physical Applications:}
Agentic spatial intelligence holds significant potential for robotics applications, enabling embodied systems to perceive, reason, and act autonomously in complex, unstructured environments. Many modern robots (e.g., Tesla Optimus~\cite{tesla_optimus}) have already integrated large foundation models for perception and control, and equipping them with spatial reasoning capabilities would further enhance their ability to plan in dynamic spaces, and adapt behaviors to complex tasks. This advancement would accelerate the development of versatile robotic platforms capable of operating effectively across service, industrial, and assistive domains.

\textbf{Embodied Household Assistance.} Spatial reasoning enables household robots to understand room layouts, track object locations, and efficiently plan for tasks such as fetching items, rearranging furniture, or assisting with meal preparation. For instance, a robot equipped with LiDAR mapping and semantic perception could navigate to the kitchen, locate a specific mug in a cupboard, and bring it to the user in the living room while avoiding pets and clutter along the way.

\textbf{Industrial and Manufacture Robot Platform.} Spatial intelligence allows industrial robots to map factory floors, coordinate with other machines, and adapt their paths in real time to avoid collisions or bottlenecks. For example, autonomous mobile robots in Amazon warehouses use LiDAR, vision, and predictive path planning to deliver shelves to pickers efficiently, while robotic arms adjust their motions dynamically when working alongside human operators.

\textbf{Healthcare Assistant Robots.} In medical settings, spatial reasoning enables robots to interpret complex room geometries, maintain precise localization, and coordinate movement around patients, staff, and equipment while executing context-specific tasks. For instance, robots can autonomously navigate crowded hospital corridors to deliver medication, reposition equipment in surgical theaters, or assist patients at home by locating and fetching essential items while avoiding obstacles.

\section{Future Direction}
\label{sec:4}
Based on our proposed computational framework (Sec.~\ref{sec:2.B}) and recent research gap (RG) of SOTA works (discussed in Sec.~\ref{sec:3.A}), this section will examine key future research directions for achieving human-like spatial reasoning capability in future agentic systems. 
\vspace{-12pt}

\subsection{Bio-inspired Multimodal Sensing} 
Contemporary AI systems primarily rely on static, pre-defined modality pipelines, lacking the dynamic, context-aware integration found in biological perception (\textbf{RG-1} in Sec.~\ref{sec:3.A.1}). To approach human-like spatial intelligence in agents, future research must incorporate both active and passive sensing mechanisms that adaptively attend to task-relevant multimodal inputs.

\textit{Biological Spatial Intelligence.} Beyond the human-inspired sensory, bio-signals enhanced by Brain–Computer Interfaces (BCI) offer a promising channel for AI spatial intelligence by capturing physiological states for human–agent communication~\cite{hieu2024comst}, thereby guiding agent spatial behavior (e.g., navigation)~\cite{guo2025neuro}. Non-invasive BCI methods utilize brain signals, such as \textbf{Electroencephalography (EEG)},  \textbf{Magnetoencephalography (MEG)}, \textbf{Functional MRI (fMRI)}, each of which offers high temporal/spatial resolution neural recording for AI-driven brain decoding (e.g., visual imagery)~\cite{chau2025population}, leading to immersive human-agent interaction~\cite{hieu2024bci}. Parallely, \textbf{Electromyography (EMG)}, particularly \textbf{ Wrist-based surface EMG (sEMG)}, records skeletal muscle activity to decode gestures, effort, and coordination~\cite{sivakumar2024emg2qwerty}, enabling the translation of muscle activity into real-time interfaces for gesture recognition, silent typing, and direct muscular communication~\cite{sivakumar2024emg2qwerty, salter2024emg2pose}.

\textit{Sensing Framework with Bottom-Up and Top-Down Attention.} Developing sensing methods that integrate bottom-up perception driven by salient environmental cues and top-down attention guided by internal goals and task relevance. Such combination allows agents to dynamically prioritize and adapt their sensory intake based on changing contexts, rather than relying solely on static, pre-trained modality pipelines of recent AI methods. Emerging backbone lies in transformer-based attention mechanisms, whose predictive nature and contextual flexibility make them well-suited for modeling both top-down and bottom-up perceptual processes~\cite{jain2022bottom}.  

\textit{Brain-inspired Sensing with Spiking-based Encoding}. In biological systems, perception is not only multimodal but also inherently time-sensitive, with neural representations shaped by the precise timing of spikes across modalities. Therefore, integrating spiking neural networks (SNNs) into the modality-specific encoders presents a compelling opportunity. With the discrete event-driven spikes feature, combining SNNs with traditional neural networks, such as convolution-based~\cite{shi2024spikingresformer} and transformer-based~\cite{guo2023transformer}, can enhance temporal resolution and contextual reasoning, forming a hybrid processing pipeline that more faithfully captures the richness of human-like perception.
\begin{researchgap}
\textbf{Research Direction 1.} Future bio-inspired multimodal sensing systems should integrate dynamic, context-aware attention and event-sensitive neural encoding to achieve human-like spatial perception across varied environments.
\end{researchgap}

\subsection{Spatial Perspective Taking} 
In terms of perspective taking, current AI systems struggle with reliably converting between egocentric and allocentric spatial frames, often treating this transformation as an implicit byproduct rather than modeling it explicitly (\textbf{RG-2} in Sec.~\ref{sec:spatial_reasoning_pipeline}). In contrast, biological systems perform flexible perspective shifts through specialized brain regions, highlighting the need for future spatial reasoning research to develop dedicated mechanisms for such transformations.

\textit{Cross-View Learning and Alignment.} A key challenge in egocentric–allocentric transformation is the ability to align observations across different viewpoints, particularly when scenes are only partially observed. Cross-view learning offers a pathway for agents to learn viewpoint-invariant spatial representations by contrasting or associating multiple perspectives of the same environment~\cite{debnath2025rasp}. Emerging backbone techniques include contrastive learning and self-supervised alignment frameworks with contrastive loss of single and multiple viewpoints~\cite{frey2023probing} and DINO-based self-supervised objectives~\cite{oquab2023dinov2} for map-view matching between egocentric and allocentric viewpoints.

\textit{Bidirectional Frame Conversion.} To support human-like spatial reasoning, agents must be able to flexibly convert between egocentric and allocentric frames in both directions, depending on the task context. This requires architectures that can learn structured spatial transformations and selectively recall or update representations anchored to different reference frames. Promising backbone techniques include transformer-based sequence-to-structure models (e.g., ViT) for structural mapping and attention routing~\cite{chen2023trans4map}, enabling modular and context-aware internal frame switching across perceptual and memory systems.
\begin{researchgap}
\textbf{Research Direction 2.} Future spatial perspective-taking systems should develop explicit mechanisms for bidirectional egocentric–allocentric transformations to enable flexible, task-adaptive spatial reasoning across diverse viewpoints.
\end{researchgap}

\subsection{Internal Cognitive Map Modelling}
Although spatial mapping techniques have advanced, current AI agents still lack an internal model that coherently captures structural and semantic information for general spatial reasoning. In contrast, biological cognitive maps can flexibly illustrate the world via topology, geometry, and episodic context (\textbf{RG-3} in Sec.~\ref{sec:2.B}), highlighting key directions for developing such capabilities in AI agents. Therefore, future research should focus on developing hybrid cognitive map architectures that unify topological, metric, and semantic representations into a flexible and context-aware spatial memory system.

\textit{Hybrid Metric-Topological Representation.} To navigate effectively, agents must combine the strengths of both metric precision and topological generalisation, just as humans leverage Euclidean distances for fine control and graph-like structures for flexible route planning. However, most current systems isolate these two representations, resulting in either overly rigid or overly abstract behaviours when faced with spatial uncertainty. Unifying these two forms via GNN, enhanced with successor representations or local coordinate anchors, offers a path toward cognitively grounded spatial models that balance generalisation with geometric fidelity~\cite{whittington2022build}.

\textit{Hierarchical Spatial Abstraction.} Human cognition structures space hierarchically, allowing flexible reasoning across scales—from objects in a room to navigating across cities. In contrast, existing AI systems lack such layered abstraction, making their spatial reasoning either too granular or too coarse for complex, goal-directed tasks. As a future direction,
Hierarchical graph learning with knowledge graph~\cite{jiang2024kg} or graph-based contrastive framework~\cite{li2024hierarchical} provide a feasible foundation to model this abstraction dynamically, thereby potentially enabling the formation of internal world representation with their robust in-context inference and topological structure.

\textit{Context Mapping.} To act intelligently, an agent must go beyond geometry and understand what entities are, what they afford, and how they relate to goals—mirroring the semantic grounding found in human spatial cognition. Yet most AI maps are purely geometric, lacking the capacity to bind visual inputs to meaningful, actionable concepts.
Integrating the retrieval of spatial memory with auto-regressive architectures such as VLMs and spatially-aware transformers~\cite{cho2024spatially}, with a decision-making layer using DRL methods can enable dynamic memory mapping with efficient contextual retrieval.
\begin{researchgap}
\textbf{Research Direction 3.} Future internal cognitive map systems should develop hybrid architectures that integrate topological, metric, and semantic representations to enable flexible, context-aware spatial reasoning across diverse scales and tasks.
\end{researchgap}

\subsection{Adaptive Spatial Memory Systems}
As noted in \textbf{RG-4} of Sec.~\ref{sec:3.A.5}, current AI agents rely on shallow or static memory mechanisms that lack semantic binding, episodic recall, or long-term consolidation. In contrast, biological systems leverage a hierarchical memory architecture for flexible recall and continual integration, suggesting future spatial reasoning should move toward dynamic, multi-scale memory models inspired by hippocampal–cortical interactions.

\textit{Semantic-Spatial Memory Integration.} Robust action planning requires an agent to know not just where things are but also what they are and how they can be used. Therefore, binding semantics to spatial nodes creates long-term knowledge that supports affordance reasoning, and multi-step goal execution. To this end, lingual-based modality serves as the backbone for semantic binding, in which the generalization from recent language models can support the spatial processing via supervised fine-tuning (SFT)~\cite{ouyang2022training} or post-training methods (e.g., RLHF, DPO)~\cite{ouyang2022training}. 

\textit{Dual-Tier Episodic Architecture.} In dynamic space, agents need a fast, high-resolution memory buffer for recent experiences while still retaining condensed traces of older episodes for days or weeks. Therefore, designing a memory architecture that emphasizes both active and static episodic memory is essential for agents, as processing 3D space requires a large amount of resources. Without this split, either the memory overflows or the agent forgets events vital for future reasoning. A hot/cold arrangement mirrors hippocampal replay and enables one-shot learning plus durable recall. A promising future direction for efficient spatial memory in agents is to integrate flexible neural codec compression, leveraging bits-back coding~\cite{hieu2024point} and deep-learning-based entropy models~\cite{li2024nerfcodec} for cold storage of 3D space, with transformer-based memory architectures that support long-context episodic interactions~\cite{chen2024melodi}. 

\textit{Continual Memory Update.} Deploying agents in the real world requires integrating unfamiliar layouts and objects without degrading existing knowledge. Naïve fine-tuning often leads to catastrophic forgetting, where updated weights overwrite core representations and destabilize navigation. To address this, lifelong learning mechanisms should preserve stable spatial representations while selectively incorporating novel scene information, allowing the map to expand coherently over time. Promising directions include modular memory architectures that localize updates to relevant submodules, preventing global interference, and dynamic capacity allocation via growable or sparsely activated networks to maintain scalability. Predictive coding frameworks can further stabilize learning by guiding updates to minimize sensory prediction errors, ensuring that spatial knowledge is refined consistently as new experiences are acquired.

\textit{Consolidation Replay.} Humans replay salient experiences during rest to strengthen useful associations and prune noise, a process that agents should emulate. Without replay, semantic–spatial links drift and the map fragments as environments change. Future work should explore saliency-based replay prioritization using reward cues, and generative replay to simulate diverse past experiences~\cite{shin2017continual}. Cross-modal replay can strengthen multimodal associations. Also, integrating replay with temporal credit assignment will enhances reasoning over long-horizon tasks~\cite{wayne2018unsupervised}.
\begin{researchgap}
\textbf{Research Direction 4.} Future adaptive spatial memory systems should develop dynamic, multi-scale architectures inspired by biological processes to enable robust, scalable, and continually updating spatial reasoning across diverse environments.
\end{researchgap}

\subsection{Spatial Inference with Predictive Nature}
Although strong spatial representations have been developed, a major challenge lies in how agentic systems can actively leverage these representations for efficient and context-aware reasoning (\textbf{RG-5} in Sec.~\ref{sec:3.A.5}). To address this, future research should prioritize enabling the predictive nature of spatial representations, forming an interactive world model that simulates potential outcomes and guides decision-making. Moreover, such a world model must be tightly aligned with not only real-world signals but also the human preference to prevent hallucinations and maintain both consistent spatial awareness and trustworthiness throughout spatial reasoning.

\textit{Predictive Learning Frameworks.} Reasoning improves when the world model can accurately forecast how the cognitive map will evolve after hypothetical moves, supporting shortcut discovery, detour evaluation, and risk-aware planning. By training latent-dynamics models to perform multi-step look-ahead and to expose prediction uncertainty, the agent gains a ``mental simulator'' that guides decisions before acting. Promising backbones lies on 3D generaive modeling (e.g., Genie 3~\cite{genie3} and visual predictive learning~\cite{lecun2022path}, in which the former can model the 3D world, providing the agents capability of understanding the 3D environment, while the latter enable the visual-centric predictive manner for agentics' spatial reasoning, mimic the predictive nature of human brain.

\textit{Spatial-aware Explicit Alignment.} Even a well-trained world model can falter when its internal representations drift from the underlying geometry and semantics established by earlier mapping stages. To ensure consistency across perception, memory, and reasoning, explicit alignment mechanisms are essential. \textbf{Incorporating spatial-equivariant architectures or post-training alignment objectives that tether reasoning embeddings to ground-truth topology and viewpoint transformations enhances multi-step inference and mitigates hallucination.} Future agents can further benefit from spatial-centric backbones such as Visual-of-Thought (VoT)~\cite{li2025imagine} and preference optimization with spatial task-centric~\cite{yan2025task}, where structured alignment is learned through reasoning-guided preference feedback over spatial tasks.

\textit{Affective Reasoning.} In addition to physical and causal knowledge, humans use emotional and social commonsense to anticipate intentions and adapt behavior. Affective reasoning allows agents to interpret ``why'' actions are taken, not just ``what'' is possible. For example, people tend to avoid collisions, follow social norms, and move toward goals they value~\cite{liukno}. By integrating affective priors with the predictive world model, the agent can infer motivations, model social dynamics, and adapt reasoning strategies accordingly~\cite{howint}. \textbf{Neurosymbolic affective knowledge} bases such as SenticNet~\cite{senticnet} provide the structured representation needed to encode these priors, enabling richer, human-aligned spatial reasoning in complex environments. 

\begin{researchgap}
\textbf{Research Direction 5.} Future spatial thinking systems should develop predictive, interactive world models aligned with real-world and human preferences to enable trustworthy, context-aware reasoning and planning across dynamic environments.
\end{researchgap}

\section{Conclusion}
\label{sec:5}
In this work, we proposed a conceptual computational framework through the lens of neuroscience framework for Agentic Spatial Intelligence, paving the way to achieve the human spatial intelligence in agentic systems. Following the neuroscience foundations, we defined the category of spatial reasoning behaviours through the basis of neuroscience models. Building upon these components, we conducted a framework-oriented analysis on recent representative works, identifying gaps and limitations in current approaches from a neurocognitive perspective. In addition, we systematically reviewed and categorized recent benchmarks and datasets based on our defined spatial reasoning behaviours, offering a cognitively grounded lens for evaluating agentic spatial reasoning. Finally, we outlined future research directions for each module, highlighting promising techniques that align with both biological principles and emerging machine learning paradigms. Altogether, this work offers a general perspective that providing a landscape to structure the design of spatial reasoning, paving a principled path toward building agents capable of flexible, human-aligned spatial intelligence.

\bibliographystyle{ACM-Reference-Format}
\bibliography{References/ref} 
\end{document}